\documentclass[journal]{IEEEtran}

\usepackage[pdftex]{graphicx}
\usepackage{amsmath}
\usepackage{multirow}
\usepackage{array}
\usepackage{subfig}
\usepackage{booktabs}
\usepackage[hyphens]{url}
\usepackage{hyperref}
\usepackage{xcolor}
\usepackage{float}
\usepackage{amsfonts}
\usepackage{balance}
\usepackage{cite}

\usepackage{amssymb}
\usepackage{algorithm}
\usepackage{algorithmic}

\newcommand\cincludegraphics[2][]{\raisebox{-0.1\height}{\includegraphics[#1]{#2}}}

\begin{document}
	\title{Dynamic Multi-Context Segmentation of Remote Sensing Images based on Convolutional Networks}
	
	\author{Keiller~Nogueira,
		Mauro~Dalla~Mura,
		Jocelyn~Chanussot,
		William~Robson~Schwartz,
		Jefersson~A.~dos~Santos
		\thanks{K.~Nogueira, W.~R.~Schwartz, and J.~A.~dos~Santos are with the Department of Computer Science, Universidade Federal de Minas Gerais, Brazil email: \{keiller.nogueira, william, jefersson\}@dcc.ufmg.br}%
		\thanks{M.~Dalla~Mura, and J.~Chanussot are with Univ. Grenoble Alpes, CNRS, Grenoble INP, GIPSA-lab, 38000 Grenoble, France email: \{mauro.dalla-mura, jocelyn.chanussot\}@gipsa-lab.grenoble-inp.fr}%
	}
	
	\markboth{IEEE}
	{Nogueira \MakeLowercase{\textit{et al.}}}
	
	\maketitle
	
	\begin{abstract}	
		Semantic segmentation requires methods capable of learning high-level features while dealing with large volume of data. Towards such goal, Convolutional Networks can learn specific and adaptable features based on the data. However, these networks are not capable of processing a whole remote sensing image, given its huge size. To overcome such limitation, the image is processed using fixed size patches. The definition of the input patch size is usually performed empirically (evaluating several sizes) or imposed (by network constraint). Both strategies suffer from drawbacks and could not lead to the best patch size. To alleviate this problem, several works exploited multi-context information by combining networks or layers. This process increases the number of parameters resulting in a more difficult model to train. In this work, we propose a novel technique to perform semantic segmentation of remote sensing images that exploits a multi-context paradigm without increasing the number of parameters while defining, in training time, the best patch size. The main idea is to train a dilated network with distinct patch sizes, allowing it to capture multi-context characteristics from heterogeneous contexts. While processing these varying patches, the network provides a score for each patch size, helping in the definition of the best size for the current scenario. A systematic evaluation of the proposed algorithm is conducted using four high-resolution remote sensing datasets with very distinct properties. Our results show that the proposed algorithm provides improvements in pixelwise classification accuracy when compared to state-of-the-art methods.
	\end{abstract}
	
	
	\IEEEpeerreviewmaketitle
	
	\section{Introduction} \label{sec:intro}

	The increased accessibility to high spatial resolution data provided by new sensor technologies has opened new horizons to the remote sensing community~\cite{toth2016remote}, allowing a better understanding 
	of the Earth's surface~\cite{yifang2015global}.
	Towards such understanding, one of the most important task is semantic labeling (or segmentation)~\cite{thoma2016survey}, which may be stated as a task of assigning a semantic category to every pixel in an image.
	Semantic segmentation allows the creation of thematic maps aiming to help in the comprehension of a scene~\cite{thoma2016survey}.
	In fact, semantic labeling has been an essential task for the remote sensing community~\cite{richards1999remote} given that its outcome, the thematic map, generates essential and useful information capable of assisting in the decision making of a wide range of fields, including environmental monitoring, intelligent agriculture~\cite{nogueira2015coffee}, disaster relief~\cite{fustes2014cloud,nogueira2018exploiting}, urban planning~\cite{volpi2017dense}.

	Given the importance of such task, several methods~\cite{dey2010review,deeplearningreview} have been proposed for the semantic segmentation of remote sensing images.
	The current state-of-the-art method for semantic segmentation is based on a resurgent approach, called deep learning~\cite{deeplearningbook}, that can learn specific and adaptable spatial features directly from the images.
	Specifically, deep learning aims at designing end-to-end trainable neural networks, i.e., systems that map raw input into an output space depending on the task.
	These systems are capable of learning features and classifiers (in distinct layers) and adjust the parameters, at running time, based on accuracy, giving more importance to one layer than another depending on the problem.
	This end-to-end feature learning (e.g., from image pixels to semantic labels) is the great advantage of deep learning when compared to previous state-of-the-art methods~\cite{lecun2015deep}, such as low-level~\cite{penatti2012comparative,penatti2015deep,nogueira2017towards} and mid-level (e.g. Bag of Visual Words~\cite{sivic2003video}) descriptors.
	
	\begin{figure}[t]
		\centering
		\includegraphics[width=\columnwidth]{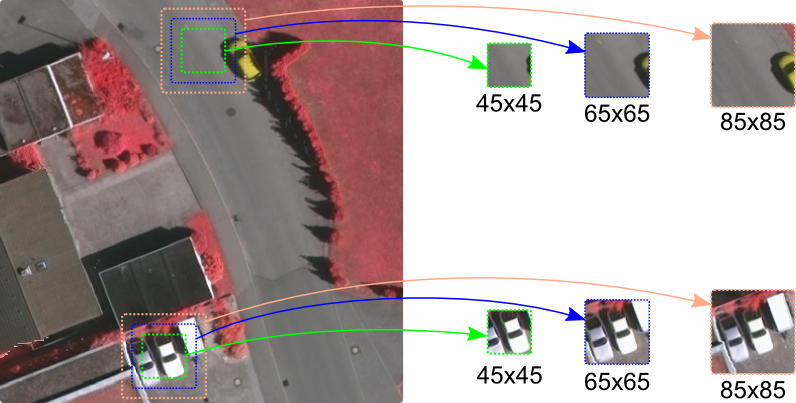}
		\caption{Example showing the importance of multi-context information.
			In the top case, while smaller contexts may not provide enough information for the understanding of the scene, a large context brings more information that may help the model to identify that it is a road with a car on it.
			In the bottom scenario, smaller contexts bring enough information for the identification of cars, while a large context may confuse the network and lead it to misclassify a different object as a car.}
		\label{fig:multi_context_importance}
	\end{figure}
	
	Among all networks, a specific type, called Convolutional (Neural) Networks, ConvNets or CNNs~\cite{deeplearningbook}, is the most traditional one for learning visual features in computer vision applications, as well as remote sensing. 
	This type of network relies on the natural stationary property of an image, i.e., the information learned in one part of the image can be used to describe any other region of the image.
	Furthermore, ConvNets usually obtain different levels of abstraction for the data, ranging from local low-level information in the initial layers (e.g., corners and edges), to more semantic descriptors, mid-level information (e.g., object parts) in intermediate layers, and high-level information (e.g., whole objects) in the final layers.
	
	Although originally proposed for image classification, to become more suitable for the semantic labeling task, these ConvNets were adapted to output a dense prediction, i.e., to produce another image (usually with the same resolution of the input) that has each pixel associated to a semantic class.
	Based on this idea, several networks~\cite{long2015fully,badrinarayanan2015segnet,noh2015learning} achieved state-of-the-art for the labeling task in the computer vision domain.
	Because of their success, these approaches were naturally introduced into the remote sensing scenario.
	Although somehow successful in this domain, these approaches could be improved if some differences for aerial images were taken into account.
	Specifically, the main difference concerns the definition of spatial context.
	In classical computer vision applications, the spatial context is restricted by the scene.
	In the case of remote sensing images, the context is typically delimited by an input patch (mainly because of memory constraints, given the huge size of remote sensing images).
	Therefore, the definition of the best input patch size is of vital importance for the network, given that patches of small size could not bring enough information to allow the network to capture the patterns while, larger patches could lead to semantically mixed information, which could affect the performance of the ConvNet.
	In the literature, the definition of this patch size is usually performed using two strategies: 
	(i) empirically~\cite{sherrah2016fully,volpi2017dense}, by evaluating several sizes and selecting the best one, which is a very expensive process, given that, for each size, a new network must be trained (without any guarantee for the best patch configuration), and
	(ii) imposed~\cite{audebert2016semantic,marmanis2016classification}, in which the patch size is defined by network constraints (i.e., changing the patch size implies modifying the architecture).
	This could be a potentially serious limitation given that the patch size required by the network could be not even close to the optimal one.
	Hence, it is clear that both current strategies suffer from drawbacks and could not lead to the best patch size.
	
	An attempt to alleviate such dependence of the patch size is to aggregate multi-context information\footnote{In this work, multi-context (sometimes called multi-scale, according to deep learning recent literature) refers to spatial context difference and, therefore, any method that exploits (direct or indirectly) images with distinct scales is aggregating multi-context information.}.
	Multi-context paradigm has been proven to be essential for segmentation methods~\cite{santos2012multiscale,sherrah2016fully}, given that it allows the model to extract and capture patterns of varying granularities, helping the method to aggregate more useful information.
	Precisely, as presented and explained in (caption of) Figure~\ref{fig:multi_context_importance}, smaller contexts may be preferable in some situations while larger ones can be useful in other scenarios.
	Therefore, several works~\cite{marcu2016dual,audebert2016semantic,maggiori2016high,marmanis2016classification,paisitkriangkrai2016semantic,wang2017gated} incorporate the benefits of the multi-context paradigm in their architectures using different approaches.
	Some of them~\cite{marcu2016dual,paisitkriangkrai2016semantic,audebert2016semantic} train several distinct layers or networks, one for each context, and combine them for the final prediction.
	Others~\cite{marmanis2016classification,maggiori2016high,wang2017gated} extract and merge features from distinct layers in order to aggregate multi-context information.
	Independently of the approach, to aggregate multi-context information, more parameters are included in the final model, resulting in a more complex learning process~\cite{deeplearningbook}.
	
	In this work, we propose a novel technique to perform semantic segmentation of remote sensing images that exploits the multi-context paradigm without increasing the number of parameters while defining adaptively the best patch size for the inference stage.
	Specifically, this technique is based upon an architecture composed exclusively on dilated convolutions~\cite{YuKoltun2016}, which are capable of processing input patch of varying sizes without distinction, given that they learn the patterns without downsampling the input.
	In fact, the multi-context information is aggregated to the model by allowing it to be trained using patches of varying sizes (and contexts), a process that increases scale-invariance and reduces over-fitting~\cite{he2014spatial}.
	This procedure allows the extraction of multi-context information without any combination of distinct networks or layers (a common process of deep learning-based multi-context approaches), resulting in a method with fewer parameters and easier to train.
	Moreover, during the training stage, the network gives a score (based on accuracy or loss) for each patch size.
	Then, in the prediction phase, the process selects the patch size with the highest score to perform the segmentation.
	Therefore, differently from empirically selecting the best patch size which requires a new network trained for each evaluated patch (increasing the computational complexity and training time), the proposed technique evaluates several patches during the training stage and selects the best one for the inference phase doing only a unique training procedure.
	Aside from the aforementioned advantages, the proposed networks can be fine-tuned for any semantic segmentation application, since they do not depend on the patch size to process the data.
	This allows other applications to benefit from the patterns extracted by our models, a very interesting feature specially when working with small amounts of labeled data~\cite{nogueira2017towards}.
	
	In practice, these are the contributions of this work:
	\begin{itemize}
		\item Our main contribution is a novel approach that performs remote sensing semantic segmentation by doing a unique training procedure that aggregates multi-context information while determining the best input patch size for the inference stage,
		\item Network architectures capable of performing semantic segmentation of remote sensing datasets with distinct properties, and that can be trained or fine-tuned for any semantic segmentation application.
	\end{itemize}
	
	The paper is structured as follows.
	Related works are presented in Section~\ref{sec:relwork} while the concept of dilated convolutions is introduced in Section~\ref{sec:background}.
	We explain the proposed technique in Section~\ref{sec:methodology}.
	Section~\ref{sec:experiments} presents the experimental protocol and Section~\ref{sec:results} reports and discusses the obtained results.
	Finally, in Section~\ref{sec:conclusions} we conclude the paper and point at promising directions for future work.

	\section{Related Work} \label{sec:relwork}
	
	As introduced, deep learning has made its way into the remote sensing community, mainly due to its success in several computer vision tasks~\cite{krizhevsky2012imagenet,long2015fully,zhang2015saliency,zhang2016scene,du2017stacked}.
	Towards a better understanding of the Earth's surface, a myriad of techniques~\cite{marcu2016dual,audebert2016semantic,maggiori2016high,sherrah2016fully,marmanis2016classification,paisitkriangkrai2016semantic,wang2017gated} have been proposed to perform semantic segmentation in remote sensing images.
	Based on previous successful models~\cite{kokkinos2015pushing,long2015fully}, several of the proposed works exploit the benefits of the multi-context paradigm.

	In~\cite{audebert2016semantic}, the authors fine-tuned a deconvolutional network (based on SegNet~\cite{badrinarayanan2015segnet}) using $256\times256$ fixed size patches.
	To incorporate multi-context knowledge into the learning process, they proposed a multi-kernel technique at the last convolutional layer.
	Specifically, the last layer is decomposed into three branches.
	Each branch processes the same feature maps but using distinct filter sizes generating different outputs which are combined into the final dense prediction.
	They argue that these different scales smooth the final predictions due to the combination of distinct fields of view and spatial context.
	
	Sherrah~\cite{sherrah2016fully} proposed methods based on fully convolutional networks~\cite{long2015fully}.
	The first architecture was purely based on the fully convolutional paradigm, i.e., the network has several downsampling layers (generating a coarse map) and a final bilinear interpolation layer, which is responsible to restore the coarse map into a dense prediction.
	In the second strategy, the previous network was adapted by replacing the downsampling layers with dilated convolutions, allowing the network to maintain the full resolution of the image.
	Finally, the last strategy evaluated by the authors was to fine-tune~\cite{nogueira2017towards} pre-trained networks over the remote sensing datasets.
	None of the aforementioned strategies exploit the benefits of the multi-context paradigm.
	Furthermore, these techniques were evaluated using several input patch sizes with final architectures processing patches with $128\times128$ or $256\times256$ pixels depending on the dataset.
	
	Marcu et al.~\cite{marcu2016dual} combined the outputs of a dual-stream network in order to aggregate multi-context information for semantic segmentation.
	Specifically, each network processes the image using patches of distinct size, i.e., one network process $256\times256$ patches (in which the global context is considered) while the other processes $64\times64$ patches (where local context is taken into account).
	The outputs of these architectures are combined in a later stage using another network.
	Although they can train the network jointly, in an end-to-end process, the number of parameters is really huge allowing them to use only small values of batch size (10 patches per batch).
	In~\cite{paisitkriangkrai2016semantic}, the authors proposed a multi-context semantic segmentation by combining ConvNets, hand-crafted descriptors, and Conditional Random Fields~\cite{lafferty2001conditional}.
	Specifically, they trained three ConvNets, each one with a different patch size ($16\times16$, $32\times32$ and $64\times64$ pixels).
	Features extracted from these networks are combined with hand-crafted ones and classified using random forest classifier.
	Finally, Conditional Random Fields~\cite{lafferty2001conditional} are used as a post-processing method in an attempt to improve the final results.

	In~\cite{marmanis2016classification}, the authors proposed multi-context methods that combine boundary detection with deconvolution networks (specifically, based on SegNet~\cite{badrinarayanan2015segnet}).
	The main contribution of this work is the Class-Boundary (CB) network, which is responsible to help the proposed methods to give more attention to the boundaries.
	Based on this CB network, they proposed several methods.
	The first uses three networks that receive as input the same image but with different resolutions (as well as the output of the corresponding CB network) and output the label predictions, which are aggregated, in a subsequent fusion stage, generating the final label.
	They also experimented fully convolutional architectures~\cite{long2015fully} (with several skip layers in order to aggregate multi-context information) and an ensemble of several architectures.
	All aforementioned networks initially receive $256\times256$ fixed size patches.
	Maggiori et al.~\cite{maggiori2016high} proposed a multi-context method that performs labeling segmentation based on upsampled and concatenated features extracted from distinct layers of a fully convolutional network~\cite{long2015fully}.
	Specifically, the network, that receives as input patches of $256\times256$ or $512\times512$ pixels (depending on the dataset), is composed of several convolutional and pooling layers, which downsample the input image.
	Downsampled feature maps extracted from several layers are, then, upsampled, concatenated and finally classified by another convolutional layer.
	This proposed strategy resembles somehow the DenseNets~\cite{huang2016densely}, with the final layer having connections to the previous ones.
	Wang et al.~\cite{wang2017gated} proposed to extract features from distinct layers of the network to capture multi-context low- and high-level information.
	They fine-tuned a ResNet-101~\cite{he2016deep} to extract salient information from $600\times600$ patches.
	Feature maps are then extracted from intermediate layers, combined with entropy maps, and upsampled to generate the final dense prediction.

	In this work, we perform semantic segmentation by exploiting a multi-context network composed uniquely of dilated convolutions.
	Three main differences between the proposed approach and the aforementioned works may be pointed out:
	(i) the proposed technique exploits a fully convolutional network that does not downsample the input data (a common process performed in most works~\cite{long2015fully,YuKoltun2016,marmanis2016classification}),
	(ii) the multi-context strategy is exploited during the training process without any modification of the network or combination of several architectures (or layers), and
	(iii) instead of evaluating possible patch sizes (to find the best one) or to use a patch size determined by network constraints (which could not be the best one), the proposed algorithm determines the best patch size adaptively in training time.

	\section{Dilated ConvNets} \label{sec:background}
	
	Dilated convolutions were originally proposed for the computation of wavelet transform~\cite{holschneider1990real} and employed in the deep learning context (as an alternative to deconvolution layers) mainly for semantic segmentation~\cite{YuKoltun2016,chen2016deeplab,sherrah2016fully}.
	In dilated convolutional layers, filter weights are employed differently when compared to standard convolutions.
	Specifically, filters of this layer may have gaps (or ``holes'') between their parameters.
	These gaps, inserted according to the dilation rate $r$, enlarge the convolutional kernel but preserve the number of trainable parameters since the inserted holes are not considered in the convolution process.
	Therefore, this dilation rate $r$ can be seen as a parameter responsible to define the final alignment of the kernel weights.
	
	Formally, for each location $i$, the output $y$ of an one-dimension dilated convolution given as input a signal $x$ and filter $w$ of length $K$ is calculated as:
	
	\begin{eqnarray} \label{eq:dilated_conv}
	y[i] = \sum_{k=1}^{K} x[i + rk] w[k]
	\end{eqnarray}
	
	The dilation rate parameter $r \in \mathbb{N}$ corresponds to the stride with which the input signal is sampled.
	As illustrated by Figure~\ref{fig:dilated_example}, smaller rates result in a more clustered filter (in fact, rate 1 generates a kernel identical to the standard convolution) while larger rates make an expansion of the filter, producing a larger kernel with several gaps.
	Since this whole process is independent of the input data, changing the dilation rate does not impact in the resolution of the outcome, i.e., in a dilated convolution, independent of the rate, input and output have the same resolution (considering appropriate stride and padding).
	
	\newcommand{\exDilatedFig}{0.15}
	
	\begin{figure}[h]
		\centering
		\subfloat[Rate 1]{
			\includegraphics[width=\exDilatedFig\textwidth]{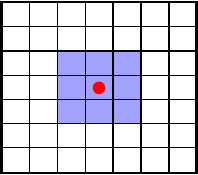}
		}
		\subfloat[Rate 2]{
			\includegraphics[width=\exDilatedFig\textwidth]{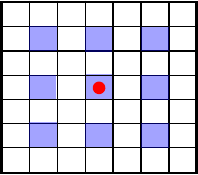}
		}
		\subfloat[Rate 3]{
			\includegraphics[width=\exDilatedFig\textwidth]{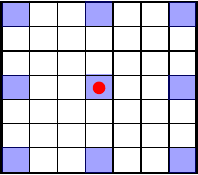}
		}
		\caption{Example of dilated convolutions.
			Dilation supports expansion of the receptive field without loss of resolution or coverage of the input.
		}
		\label{fig:dilated_example}
	\end{figure}
	
	By enlarging the filter (with such gaps), the network expands its receptive field (since the weights will be arranged in a more sparse shape) but preserves the resolution and no downsampling in the data is performed.
	Hence, this process has several advantages, such as:
	(i) supports the expansion of the receptive field without increasing the number of trainable parameters per layer~\cite{YuKoltun2016}, which reduces the computational burden, and
	(ii) preserves the feature map resolution, which may help the network to extract even more useful information from the data, mainly of small objects.
	
\begin{figure}[h]
		\centering
		\includegraphics[width=0.48\textwidth]{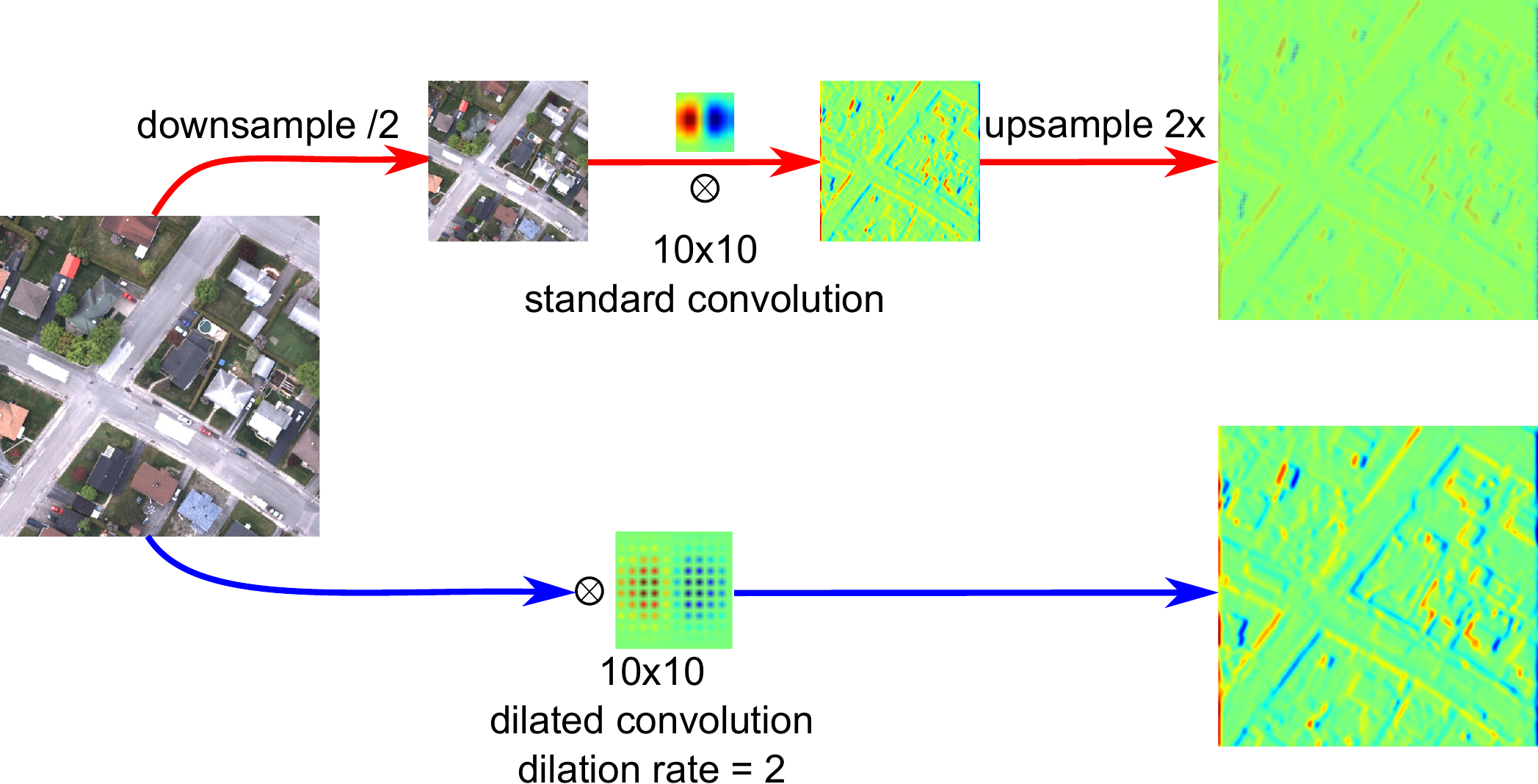}
		\caption{Comparison between dilated and standard convolutions.
			Top (red) row presents the feature extraction process using a standard convolution over a downsampled image and then an upsample in order to recover the input resolution (a common procedure performed in ConvNets).
			Bottom (blue) row presents the feature extraction process using dilated convolution with rate $r = 2$ applied directly to the input (without downsample).
			The outcomes clearly show the benefits of dilated convolutions over standard ones.
		}
		\label{fig:comparison_dilated}
	\end{figure}

	To better understand the aforementioned advantage, a comparison between dilated and standard convolution is presented in Figure~\ref{fig:comparison_dilated}.
	Given an image, the first network (in red) performs a downsampling operation (that reduces the resolution by a factor of 2) and a convolution, using horizontal Gaussian derivative as the kernel.
	The obtained low-resolution feature map is then enlarged by an upsampling operation (with a factor of 2) that restores the original resolution but not the information lost during the downsampling process.
	The second network (blue) computes the response of a dilated convolution on the original image.
	In this case, the same kernel was used but rearranged with dilation rate $r = 2$, making both networks have the same receptive field.
	Although the filter size increases, only non-zero values are taken into account when performing the convolution.
	Therefore, the number of filter parameters and of operations per position stay constant.
	Furthermore, it is possible to observe that salient features are better represented by the dilated model since no downsampling is performed over the input data.

	\section{Dynamic Multi-Context Dilated Convolution} \label{sec:methodology}
	
	In this section, we present the proposed method for dynamic multi-context semantic segmentation of remote sensing images.
	The proposed methodology is presented in Section~\ref{subsec:proposed} while the network architectures are described in Section~\ref{subsec:arch}.
	
	\subsection{Dynamic Multi-Context Algorithm} \label{subsec:proposed}
	
	We propose a novel method to perform semantic segmentation of remote sensing images that:
	(i) exploits the multi-context paradigm without increasing the number of trainable parameters of the network, and
	(ii) defines, in training time, the best patch size that should be exploited by the network in the \textbf{test phase}.
	
	As presented in Algorithm~\ref{algo:training}, the \textbf{training process} receives as input:
	(i) the data $\mathcal{D}$, where the images and their reference labels come from, 
	(ii) a patch size distribution $\mathcal{P}$, that represents the probability function (kept the same during all the training procedure) from which the patch sizes will come from,
	(iii) the patch scores $\mathcal{S}$ (initialized with zeros), which will be used during the training procedure to accumulate the score of the patch sizes produced by the network,
	(iv) the network $\mathcal{N}$, which, in this work, can be seen as a function that processes the input batch $(\mathcal{X}, \mathcal{Y}) \in \mathcal{D}$ (a tuple of patches and reference semantic labels) with respect to the current weights $\mathcal{W}$, updating them, and outputting a score for the batch $v$, that can be seen, somehow, as a quality assessment of the patch size relative to the current network,
	(v) the number of iterations or epochs $n$.
	
	The first step of the training procedure is to randomly select a patch size $\lambda_k$ from the distribution $\mathcal{P}$, which may be any valid distribution, such as uniform or multinomial.
	Then, this patch size $\lambda_k$ is used to create a new batch $(\mathcal{X}_{\lambda_k \times \lambda_k}, \mathcal{Y}_{\lambda_k \times \lambda_k}) \in \mathcal{D}$.
	Observe that, at each iteration of the algorithm, a new patch size is selected and a new random batch (using different sites) is sampled based on this size.
	This batch is then employed to train the network $\mathcal{N}$, i.e., to update its weights $\mathcal{W}$.
	It is important to emphasize that this training process (performed by the sampled batch) represents only a single step (iteration) of the mini-batch optimization strategy~\cite{deeplearningbook} (and not the full train) that processes one whole batch to then update the network weights $\mathcal{W}$.
	As aforementioned, for each step of the mini-batch training algorithm, the network $\mathcal{N}$ outputs a score for the current batch $v$, which can be any metric (such as a loss or accuracy) that estimates the performance of the network based on the current batch.
	This generated score $v$ is used to update the patch scores $\mathcal{S}$, which accumulate, throughout the training procedure, the scores of the patch sizes and are employed in the selection of the best patch size during the inference stage.
	Note the difference between the patch distribution $\mathcal{P}$ and the scores $\mathcal{S}$, i.e., while the former is a distribution employed during the whole training procedure the latter accumulates the scores of the patch sizes given by the network to be employed in the prediction phase.
	Hence, there is no connection between patch distribution $\mathcal{P}$ and the scores $\mathcal{S}$, and an update in $\mathcal{S}$ has no impact on $\mathcal{P}$, which is kept fixed throughout the training process.
	
	All the aforementioned steps are repeated during the training process until the number of iterations $n$ is reached.
	As it can be noticed, the multi-context information is aggregated to the model by allowing the network to be trained using batches composed of patches of multiple sizes.
	This process allows the network to capture and extract features by considering distinct context regions, a very important process as presented and explained in (caption of) Figure~\ref{fig:multi_context_importance}.
	
	When the training phase is finalized, the algorithm outputs the updated network $\mathcal{N}$ (i.e., its weights $\mathcal{W}$) and the updated patch scores $\mathcal{S}$.
	The second benefit of the proposed method is almost a direct application of the patch scores $\mathcal{S}$ created during the training phase.
	Precisely, in the \textbf{prediction phase}, scores $\mathcal{S}$ over the patch sizes are averaged and analyzed.
	The \textbf{best patch} size $\lambda^{*}$ (which corresponds to the highest or lowest score, for accuracy and loss, respectively) is then selected and used to create patches.
	The network processes these patches (of $\lambda^{*}\times\lambda^{*}$ pixels) outputting the prediction maps, but no updates in the patch scores $\mathcal{S}$ are performed.
	It is important to highlight that the proposed technique can only choose the best patch size within all possible sizes determined by the patch distribution $\mathcal{P}$, since only the patches within $\mathcal{P}$ are evaluated by the algorithm.
	
	\begin{algorithm}\captionsetup{labelfont={sc,bf}, labelsep=newline}
		\caption{Process of dynamic training a Convolutional Networks.}
		\label{algo:training}
		\begin{algorithmic}
			\REQUIRE{data $\mathcal{D}$, network $\mathcal{N}$ with its weights $\mathcal{W}$, number of iterations $n$, patch distribution $\mathcal{P}$, and patch scores $\mathcal{S}$ (initialized with zeros).}
			\ENSURE{updated of the network weights $\mathcal{W}$, and patch scores $\mathcal{S}$.}
			
			\FOR{t=1 to n}
			\STATE{$\lambda_k  = \mathcal{P}(k)$} \COMMENT{Randomly select current size}
			\STATE{$(\mathcal{X}_{\lambda_k \times \lambda_k}, \mathcal{Y}_{\lambda_k \times \lambda_k}) \in \mathcal{D}$} \COMMENT{Create new batch}
			\STATE{$v_{\lambda_k} = N(\mathcal{X}_{\lambda_k \times \lambda_k}, \mathcal{Y}_{\lambda_k \times \lambda_k}; \mathcal{W})$} \COMMENT{Continue training}
			\STATE{$\mathcal{S}_{\lambda_k} = \mathcal{S}_{\lambda_k} + v_{\lambda_k}$} \COMMENT{Update scores}
			\ENDFOR
		\end{algorithmic}
	\end{algorithm}
	
	\subsection{Architectures} \label{subsec:arch}
	
	As presented in Section~\ref{sec:background}, the properties of the dilated convolutions~\cite{YuKoltun2016} make them fit perfectly into the proposed multi-context methodology, given that a network composed of such layers is capable of processing an input of any size without downsampling it.
	This creates the possibility of processing patches of any size without constraints.
	Although these layers have the advantage of computing feature responses at the original image resolution, a network composed uniquely of dilated convolutions would be costly to train especially when processing entire (large) scenes.
	However, as previously mentioned, processing an entire remote sensing image is not possible (because of its huge size) and, therefore splitting the image into small patches is already necessary, which naturally alleviates the training process.
	
	Though, in this work, we explore networks composed of dilated convolutions, other types of ConvNets could be used, such as fully convolutions~\cite{long2015fully} and deconvolutions~\cite{noh2015learning,badrinarayanan2015segnet}.
	These networks can also process patches of varying size, but they have restrictions related to a high variation of the patch size.
	Specifically, these networks need to receive a patch larger enough to allow the generation of a coarse map, that is upsampled to the original size.
	If the input patch is too small, the network could reach a situation where it is not possible to create the coarse map and, consequently, the final upsampled map.
	Such problem is overcome by dilated convolutions~\cite{YuKoltun2016}, which are allowed to process patches of any size, without distinction, always outputting results with the same resolution of the input data (given proper configurations, such as stride and padding).
	Such concept is essential to allow the variance of patch sizes, from very small values (such as $7\times7$) to larger ones (for instance, $256\times256$).

	\begin{figure*}[t]
		\centering
		\subfloat[Dilated ConvNet with max rate 6]{
			\includegraphics[width=\textwidth]{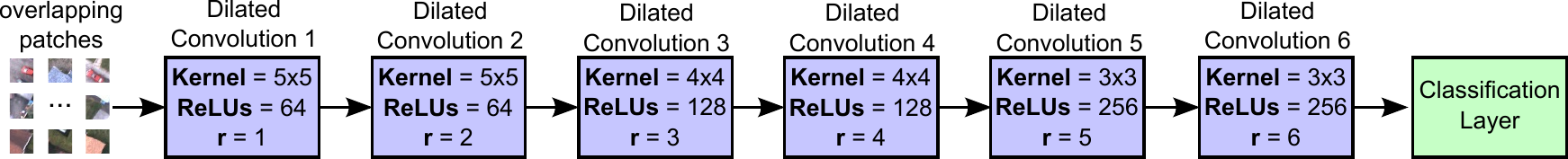}
			\label{dilated1}
		}
		\hspace{1mm}
		\subfloat[Densely Dilated ConvNet]{
			\includegraphics[width=\textwidth]{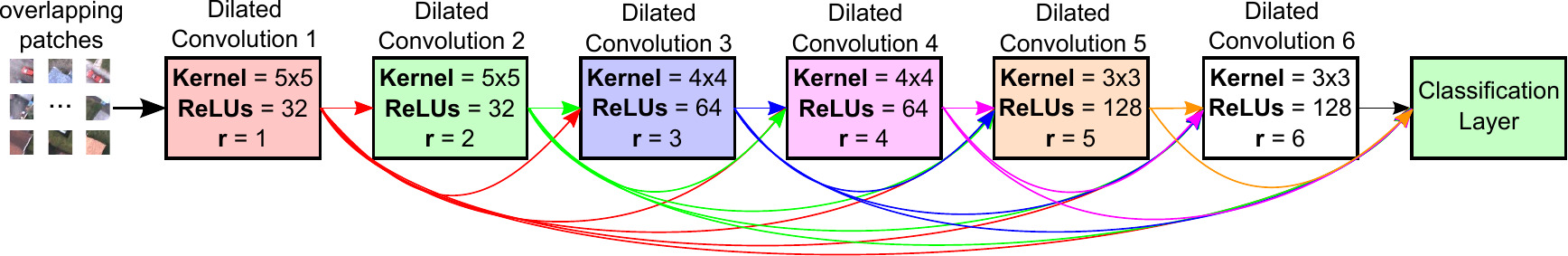}
			\label{dilated2}
		}
		\hspace{1mm}
		\subfloat[Dilated ConvNet with max rate 6 and pooling]{
			\includegraphics[width=\textwidth]{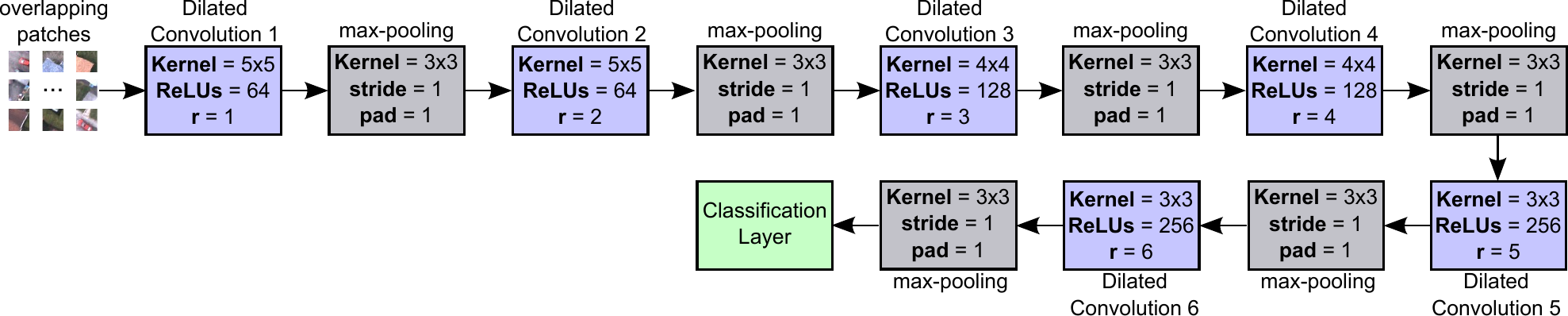}
			\label{dilated3}
		}
		\hspace{1mm}
		\subfloat[Dilated ConvNet with max rate 8 and pooling]{
			\includegraphics[width=\textwidth]{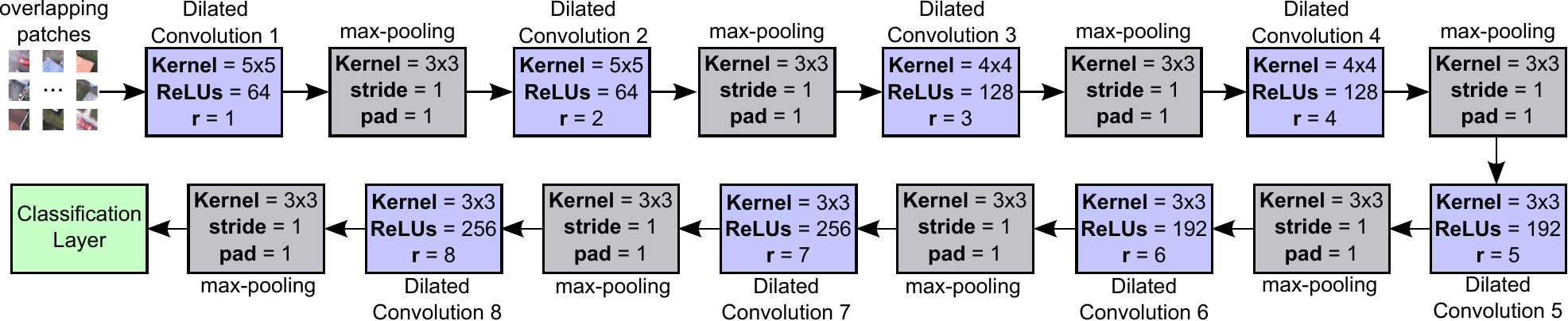}
			\label{dilated4}
		}
		\caption{Dilated Convolutional Network architectures.}
		\label{fig:cnn_arch}
	\end{figure*}
	
	Considering this, a full set of experiments (guided by~\cite{bengio2012practical}) was performed in order to define the best architectures.
	After the experiments, four networks, illustrated in Figure~\ref{fig:cnn_arch}, have been selected (based on the accuracy) and extensively evaluated in this work.
	The first network, presented in Figure~\ref{dilated1}, is composed of seven layers:
	six dilated convolutions (that are responsible to capture the patterns of the input images) and a final $1\times1$ convolution layer, which is responsible to generate the dense predictions.
	There is no pooling or normalization in this network, and all layers have stride~1.
	Specifically, the first two convolutions have $5\times5$ filters with dilation rate $r$ 1 and 2, respectively.
	The following two convolutions have $4\times4$ filters but rate 3 and 4 while the last two convolutions have smaller filters ($3\times3$) but 5 and 6 as dilation rate.
	Because this network has 6 layers responsible for the feature extraction, it will be referenced as \textbf{Dilated6}.
	The second network (Figure~\ref{dilated2}) is based on densely connected networks~\cite{huang2016densely}, which recently achieved outstanding results on the image classification task.
	This network is very similar to the first one having the same number of layers and configuration.
	The main difference between these networks is that a layer receives as input feature maps of all preceding layers.
	Hence, the last layer has access to all feature maps generated by all other layers of the network.
	This process allows the network to combine different feature maps with distinct level of abstraction, supporting the capture and learning of a wide range of feature combination.
	Because this network has 6 layers responsible for the feature extraction and is densely connected, it will be referenced in this work as \textbf{DenseDilated6}.
	The third network, presented in Figure~\ref{dilated3}, has the same configuration of the Dilated6, but with pooling layers between each convolutional one.
	Given a specific combination of stride and padding, no downsampling is performed over the inputs in these pooling layers.
	Because of the number of layers and the pooling layers, this network will be referenced hereafter as \textbf{Dilated6Pooling}.
	The last network (Figure~\ref{dilated4}) is an extension of the previous one, having 8 dilated convolutions instead of only 6.
	The last two convolutional layers have smaller filters ($3\times3$) but 7 and 8 as dilation rate.
	There are pooling layers between all convolutional ones.
	Given that this network has 8 dilated convolutional and pooling layers, it will be referenced hereafter as \textbf{Dilated8Pooling}.
	Although only this network with 8 layers is explored in this work, other variant networks (such as Dilated8 and DenseDilated8) were initially considered but not retained for further experiments due to the similar initial performance and longer training time when compared to the Dilated6 variant networks.
	
	\section{Experimental Setup} \label{sec:experiments}
	
	In this section, we present the experimental setup.
	Specifically, Section~\ref{subsec:datasets} presents the datasets employed in this work.
	Baselines are described in Section~\ref{subsec:baselines} while the experimental protocol is introduced in Section~\ref{subsec:protocol}.
	
	\subsection{Datasets} \label{subsec:datasets}
	
	To better evaluate the effectiveness of the proposed method, we carried out experiments on four high-resolution remote sensing datasets with very distinct properties.
	The first one is an agricultural dataset composed of multispectral high-resolution scenes of coffee crops and non-coffee areas.
	The others are urban datasets which have the objective of mapping targets such as roads, buildings, and cars.
	The first one is the GRSS Data Fusion contest dataset (consisting of very high-resolution images), while the others are the Vaihingen and Potsdam datasets, provided in the framework of the 2D semantic labeling contest organized by the ISPRS Commission III\footnote{\url{http://www2.isprs.org/commissions/comm3/wg4/semantic-labeling.html}} and composed of multispectral high-resolution images.
	
	\subsubsection{Coffee Dataset} \label{subsubsec:coffee_dataset}
	
	This dataset~\cite{keiller2016icpr} is composed of 5 images taken by the SPOT sensor in 2005 over a famous coffee grower county (Monte Santo) in the State of Minas Gerais, Brazil.
	Each image has $500\times500$ pixels with green, red, and near-infrared bands (in this order), which are the most useful and representative ones for discriminating vegetation areas~\cite{santos2012multiscale}.
	More specifically, the dataset consists of 1,250,000 pixels classified into two classes: coffee (637,544 pixels or 51\%) and non-coffee (612,456 pixels or 49\%).
	Figure~\ref{fig:coffee_results} presents the images and ground-truths of this dataset.

	This dataset is very challenging for several different reasons, including:
	(i) high intraclass variance, caused by different crop management techniques,
	(ii) scenes with distinct plant ages, since coffee is an evergreen culture and,
	(iii) images with spectral distortions caused by shadows, since the South of Minas Gerais is a mountainous region.
	
	\subsubsection{GRSS Data Fusion Dataset} \label{subsubsec:grss_dataset}
	
	Proposed for the 2014 IEEE GRSS Data Fusion Contest, this dataset~\cite{liao2015processing} is composed of two (training and testing) fine-resolution visible (RGB) images that cover an urban area near Thetford Mines in Quebec, Canada.
	Both training and testing images have $0.2$ meter of spatial resolution, with the former having $2830\times3989$ and the latter $3769\times4386$ pixels of resolution. 
	Training and testing images, as well as the respective ground-truths, are presented in Figure~\ref{fig:grss_results}.
	
	Pixels are categorized into seven classes: trees, vegetation, road, bare soil, red roof, gray roof, and concrete roof.
	The dataset is not balanced, as can be seen in Table~\ref{tab:grss_stat}.
	It is important to highlight that not all pixels are classified into one of these categories, with some pixels considered as uncategorized or unclassified.
	
	\begin{table}[]
		\centering
		\caption{Number of pixels per class for the GRSS Data Fusion dataset.}
		\label{tab:grss_stat}
		\begin{tabular}{@{}l|rr|rr@{}}
			\toprule
			\textbf{}              & \multicolumn{2}{c}{\textbf{Train}}                                      & \multicolumn{2}{c}{\textbf{Test}}                                       \\ \midrule
			\textbf{Classes}       & \multicolumn{1}{c}{\textbf{\#Pixels}} & \multicolumn{1}{c}{\textbf{\%}} & \multicolumn{1}{c}{\textbf{\#Pixels}} & \multicolumn{1}{c}{\textbf{\%}} \\
			\textbf{Road}          & 112,457                               & 19.83                           & 808,490                               & 55.77                           \\
			\textbf{Trees}         & 27,700                                & 4.89                            & 100,528                               & 6.93                            \\
			\textbf{Red roof}      & 45,739                                & 8.05                            & 136,323                               & 9.40                            \\
			\textbf{Grey roof}     & 53,520                                & 9.44                            & 142,710                               & 9.84                            \\
			\textbf{Concrete roof} & 97,821                                & 17.25                           & 109,423                               & 7.55                            \\
			\textbf{Vegetation}    & 185,242                               & 32.65                           & 102,948                               & 7.10                            \\
			\textbf{Bare soil}     & 44,738                                & 7.89                            & 49,212                                & 3.41                            \\
			\textbf{Total}         & 567,217                               & 100.00                           & 1,449,634                             & 100.00                           \\ \bottomrule
		\end{tabular}
	\end{table}
	
	\subsubsection{Vaihingen Dataset} \label{subsec:vaihingen}
	
	As introduced, this dataset~\cite{vaihingen} was released for the 2D semantic labeling contest of the International Society for Photogrammetry and Remote Sensing (ISPRS).
	It is composed by a total of 33 image tiles (with an average size of $2494\times2064$ pixels), that are densely classified into six possible labels: impervious surfaces, building, low vegetation, tree, car, clutter/background.
	Sixteen of these images have ground-truth available while the remaining ones, considered the test set, do not have available annotation, requiring submission of the predictions in order to be evaluated.
	The pixel distribution for the labeled images can be seen in Table~\ref{tab:isprs_stat}.
	
	Each image of this dataset is composed of near-infrared, red and green channels (in this order) and has a spatial resolution of $0.9$ meter.
	A Digital Surface Model (DSM) coregistered to the image data was also provided, allowing the creation of a normalized DSM (nDSM) by~\cite{gerke2015use}.
	In this work, we use the spectral information (NIR-R-G) and the nDSM, i.e., the input data for the method has 4 dimensions: NIR-R-G and nDSM.
	Examples of the Vaihingen Dataset can be seen in Figure~\ref{fig:vaihingein_validation_results}.
	
	\begin{table}[]
		\centering
		\caption{Number of pixels per class for ISPRS dataset, i.e., Vaihingen and Potsdam.}
		\label{tab:isprs_stat}
		\begin{tabular}{@{}l|rr|rr@{}}
			\toprule
			\textbf{}              & \multicolumn{2}{c}{\textbf{Vaihingen}}                                      & \multicolumn{2}{c}{\textbf{Potsdam}}                                       \\ \midrule
			\textbf{Classes}       & \multicolumn{1}{c}{\textbf{\#Pixels}} & \multicolumn{1}{c}{\textbf{\%}} & \multicolumn{1}{c}{\textbf{\#Pixels}} & \multicolumn{1}{c}{\textbf{\%}} \\
			\textbf{Impervious Surfaces} & 21,815,349                              & 27.94                           & 245,930,445                             & 28.46                           \\
			\textbf{Building}            & 20,417,332                              & 26.15                           & 230,875,852                             & 26.72                           \\
			\textbf{Low Vegetation}      & 16,272,917                              & 20.84                           & 203,358,663                             & 23.54                           \\
			\textbf{Tree}                & 18,110,438                              & 23.19                           & 126,352,970                             & 14.62                           \\
			\textbf{Car}                 & 945,687                                & 1.21                            & 14,597,667                              & 1.69                            \\
			\textbf{Clutter/Background}  & 526,083                                & 0.67                            & 42,884,403                              & 4.96                            \\
			\textbf{Total}               & 78,087,806                              & 100.00                          & 864,000,000                             & 100.00                         \\ \bottomrule
		\end{tabular}
	\end{table}
	
	\subsubsection{Potsdam Dataset} \label{subsec:potsdam}
	
	Also proposed for the 2D semantic labeling contest, this dataset~\cite{potsdam} has 38 tiles of the same size ($6000\times6000$ pixels), with a spatial resolution of $0.5$ meter.
	From the available patches, 24 are densely annotated (with same classes as for the Vaihingen dataset), in which the pixel distribution is presented in Table~\ref{tab:isprs_stat}.
	Analogously to the Vaihingen dataset, the remaining images are considered the test set and do not have available annotation, requiring submission of the predictions in order to be evaluated.
	This dataset consists of 4-channel images (near-infrared, red, green and blue), Digital Surface Model (DSM), and normalized DSM (nDSM).
	In this work, all spectral channels plus the nDSM are used as input for the ConvNet, resulting in a 5-dimensional input data.
	Some samples of these images are presented in Figure~\ref{fig:postdam_validation_results}.

	\subsection{Baselines} \label{subsec:baselines}
	
	For the coffee dataset, we employed the Cascaded Convolutional Neural Network (CCNN)~\cite{nogueira2015coffee} as baseline.
	This method employs a multi-context strategy by aggregating several ConvNets in order to perform the classification of fixed size tiles towards the final segmentation of the image.
	For the GRSS Data Fusion Dataset, we employed, as baseline, the work of Santana et al.~\cite{santanadeep}.
	Their algorithm extracts features with many levels of context by exploiting different layers of a pre-trained convolutional network, which are then combined in order to aggregate multi-context information.
	
	Aside this, for both aforementioned datasets, we also considered as baseline the method conceived by~\cite{keiller2016icpr}, in which specific networks are used to perform labeling segmentation using the pixelwise paradigm, i.e., each pixel is classified independently by the classifier.
	Also, for these two datasets, we considered as baselines:
	(i) Fully Convolutional Networks (FCN)~\cite{long2015fully}.
	In this case, the pixelwise architectures proposed by~\cite{keiller2016icpr} were converted into fully convolutional network and exploited as baseline.
	(ii) Deconvolutional networks~\cite{badrinarayanan2015segnet,noh2015learning}.
	Again, the pixelwise architectures proposed by~\cite{keiller2016icpr} were converted into deconvolutional network (based on the well-known SegNet~\cite{badrinarayanan2015segnet} architecture) and exploited as a baseline in this work.
	(iii) dilated network~\cite{YuKoltun2016}, which is, in this case, the Dilated6Pooling (Figure~\ref{dilated3}).
	All these networks were trained traditionally using patches of constant size defined according to a set of experiments of~\cite{keiller2016icpr}, i.e., patches of $7\times7$ and $25\times25$ for Coffee and GRSS Data Fusion datasets, respectively.
	
	For the remaining datasets (Vaihingen and Potsdam), 
	we refer to the official results published on the challenge website\footnote{\url{http://www2.isprs.org/commissions/comm2/wg4/vaihingen-2d-semantic-labeling-contest.html} and \url{http://www2.isprs.org/commissions/comm2/wg4/potsdam-2d-semantic-labeling.html}.} as baselines for the proposed work.
	
	\subsection{Experimental Protocol} \label{subsec:protocol}
	
	For the Coffee~\cite{keiller2016icpr} and the GRSS Data Fusion~\cite{liao2015processing} datasets, we employed the same protocol of~\cite{keiller2016icpr}.
	Specifically, for the former dataset, we conducted a five-fold cross-validation to assess the performance of the proposed algorithm.
	In this strategy, five runs are executed, where, at each run, three coffee scenes are used as training while, one is used as validation, and the remaining one is used as test.
	The reported results are the average metric of the five runs followed by its corresponding standard deviation.
	For the GRSS Data Fusion dataset, an image was used for training while the other was used for test, since this dataset has a clear definition of training/testing.
	
	For Vaihingen~\cite{vaihingen} and Potsdam~\cite{potsdam} datasets, we followed the protocol proposed by~\cite{volpi2017dense}.
	For the Vaihingen dataset, 11 out of the 16 annotated images were used to train the network.
	The 5 remaining images (with IDs 11, 15, 28, 30, 34) were employed to validate and evaluate the segmentation generalization accuracy.
	For the Potsdam dataset, 18 (out of 24) images were used for training the proposed technique.
	The remaining 6 images (with IDs 02\_12, 03\_12, 04\_12, 05\_12, 06\_12, 07\_12) were employed for validation of the method.
	
	Four metrics~\cite{congalton2008assessing} were considered to assess the performance of the proposed algorithm: overall and average accuracy, kappa index and F1 score.
	Overall accuracy is a metric that considers the global aspects of the classification, i.e., it takes into account all correctly classified pixels indistinctly.
	On the other hand, average accuracy reports the average (per-class) ratio of correctly classified samples, i.e., it outputs an average of the accuracy of each class.
	Kappa index measures the agreement between the reference and the prediction map.
	Finally, F1 score is defined as the harmonic mean of precision and recall.
	These metrics were selected based on their diversity: overall accuracy and kappa are biased toward large classes (relevance of classes with small amount of samples are canceled out by those with large amount) while average accuracy and F1 are calculated specifically for each class and, therefore, are independent of class size.
	Hence, the presented results are always some combination of such metrics in order to provide enough information about the effectiveness of the proposed method.
	
	The proposed method and network\footnote{The code has been made publicly available at~\url{https://github.com/keillernogueira/dynamic-rs-segmentation/}.} were implemented using TensorFlow~\cite{tensorflow2015}, a framework conceived to allow efficient exploitation of deep learning with Graphics Processing Units (GPUs).
	All experiments were performed on a 64 bits Intel i7 4960X machine with 3.6GHz of clock and 64GB of RAM memory. 
	Four GeForce GTX Titan X with 12GB of memory, under an 8.0 CUDA version, were employed in this work.
	Note, however, that each GPU was used independently and that all networks proposed here can be trained using only one GPU.
	Ubuntu version 16.04.3 LTS was used as operating system.
	
	As previously stated, a set of experiments (guided by~\cite{bengio2012practical}) was executed to define the hyperparameters.
	After all the setup experiments, the best values for hyperparameters, presented in Table~\ref{tab:parameters}, were determined for each dataset.
	The number of iterations increases with the complexity of the dataset in order to ensure convergence.
	In the proposed models, the learning rate, responsible to determine how much an updating step influences the current value of the network weights, starts with a high value and is reduced during the training phase using the exponential decay~\cite{tensorflow2015} with parameters defined according to the last column of Table~\ref{tab:parameters}.
	
	\begin{table}[]
		\centering
		\caption{Hyperparameters employed in each dataset.}
		\label{tab:parameters}
		\resizebox{\columnwidth}{!}{
			\begin{tabular}{@{}lrrrr@{}}
				\toprule
				\multicolumn{1}{c}{\textbf{Datasets}} & \multicolumn{1}{c}{\textbf{\begin{tabular}[c]{@{}c@{}}Learning\\Rate\end{tabular}}} &
				\multicolumn{1}{c}{\textbf{\begin{tabular}[c]{@{}c@{}}Weight\\Decay\end{tabular}}} & \multicolumn{1}{l}{\textbf{Iterations}} & \multicolumn{1}{c}{\textbf{\begin{tabular}[c]{@{}c@{}}Exponential Decay\\(decay/steps)\end{tabular}}} \\ \midrule
				\textbf{Coffee Dataset}           & 0.01                                       & 0.001                                     & 150,000                                 & 0.5/50,000                                       \\
				\textbf{GRSS Data Fusion Dataset} & 0.01                                       & 0.005                                     & 200,000                                 & 0.5/50,000                                       \\
				\textbf{Vaihingen Dataset}        & 0.01                                       & 0.01                                      & 500,000                                 & 0.5/50,000                                       \\
				\textbf{Potsdam Dataset}          & 0.01                                       & 0.01                                      & 500,000                                 & 0.5/50,000                                       \\ \bottomrule
			\end{tabular}
		}
	\end{table}
	
	\section{Results and Discussion} \label{sec:results}
	
	In this section, we present and discuss the obtained results.
	Specifically, we first analyze the parameters of the proposed technique:
	Section~\ref{subsec:distri} presents the results achieved using different patch distributions,
	Section~\ref{subsec:score} analyzes distinct functions to update the patch size score, and
	Section~\ref{subsec:range} evaluates different ranges for the patch size.
	Then, a comparison between the dilated and standard convolution is presented in Section~\ref{subsec:conv_op}.
	A convergence analysis of the proposed technique is performed in Section~\ref{subsec:convergence} while a comparison between networks trained with the proposed and standard training techniques is presented in Section~\ref{subsec:dilated_comparison}.
	Finally, a comparison with the state-of-the-art is reported in Section~\ref{subsec:state}.
	
	\subsection{Patch Distribution Analysis} \label{subsec:distri}
	
	As explained at the beginning of Section~\ref{subsec:proposed}, the algorithm receives as input a list of possible patch sizes and a correspondent distribution.
	In fact, any distribution could be used, including uniform or multinomial.
	Given the influence of this distribution over the proposed algorithm, experiments have been conducted to determine the most appropriate distribution.
	Towards this, we selected and compared three distinct distributions.
	First is the \textbf{uniform} distribution over a range of values, i.e., given two extreme points, all intermediate values (extremes included) inside this range should have the same probability of being selected.
	Second is the uniform distribution but over selected values (and not a range).
	In this case, referenced as \textbf{uniform fixed}, the probability distribution is equally divided into the given values (the remaining intermediate points have no probability of being selected).
	The last distribution evaluated is the \textbf{multinomial}.
	In this case, ordinary values inside a range have the same probability but several given points have twice the chance of being selected.
	
	The main difference between the evaluated distributions is related to the prior knowledge of the application.
	In the uniform distribution, no prior knowledge is assumed, and all patch sizes from the input range have the same probability, taking more time to converge the model.
	The uniform fixed distribution assumes a good knowledge of the application and only pre-defined patch sizes can be (equally) selected and evaluated, taking less time to converge the model.
	The multinomial distribution tries to blend previous ideas.
	Assuming a certain prior knowledge of the application, the multinomial distribution weighs the probabilities allowing the network to give more attention to specific pre-defined patch sizes but without discarding the others.
	If prior intuition is confirmed, these pre-defined patch sizes are randomly selected more often and the network should converge faster.
	Otherwise, the proposed process is still able to use other (non-pre-defined) patch sizes and converge the network anyway.
	
	Results of this analysis can be seen in Table~\ref{tab:distribution_analysis}.
	Note that all experiments were performed using the Coffee dataset~\cite{keiller2016icpr}, Dilated6 network (Figure~\ref{dilated1}), accuracy as score function, and hyperparameters presented in Table~\ref{tab:parameters}.
	In these experiments, patches size varied from $25\times25$ to $50\times50$.
	Specifically, for the uniform distribution, any value between 25 and 50 has the same probability of being selected, while for the multinomial distribution, all values have some chance to be selected, but these two points have twice the probability.
	For the uniform fixed, these two patch sizes split the total probability and each one has 50\% of being selected.
	Overall, the variation of the distribution has no serious impact on the final outcome, since results are all very similar.
	However, given its simplicity and faster convergence, for the remaining of this work, results will be reported using the \textbf{uniform fixed} distribution.
	
	\begin{table}[]
		\centering
		\caption{Results over different distributions.}
		\label{tab:distribution_analysis}
		\resizebox{\columnwidth}{!}{
			\begin{tabular}{@{}lrrrr@{}}
				\toprule
				& \multicolumn{1}{c}{\textbf{Overall Accuracy}} & \multicolumn{1}{c}{\textbf{Kappa}} & \multicolumn{1}{c}{\textbf{Average Accuracy}} & \multicolumn{1}{c}{\textbf{F1 Score}} \\ \midrule
				\textbf{Uniform}     & 86.13$\pm$2.39                                & 69.39$\pm$3.48                     & 84.81$\pm$1.65                                & 84.58$\pm$1.90                        \\
				\textbf{Uniform Fixed}  & 86.27$\pm$1.44                                & 69.41$\pm$2.01                     & 84.85$\pm$1.66                                & 84.62$\pm$1.06                        \\ 
				\textbf{Multinomial} & 86.06$\pm$1.68                                & 68.94$\pm$2.94                     & 84.56$\pm$2.00                                & 84.39$\pm$1.51                        \\ \bottomrule
			\end{tabular}
		}
	\end{table}
	
	\subsection{Score Function Analysis} \label{subsec:score}
	
	As introduced in Section~\ref{sec:methodology}, at each training iteration an update is performed in the score of patch sizes, which are used in the selection of the best patch size during the testing stage.
	In this work, we evaluated two possible score functions that could be employed in this step: the loss and the accuracy.
	In the first case, the loss is a measure (obtained using cross entropy~\cite{deeplearningbook}, in this case) that represents the error generated in terms of the ground-truths and the network predictions.
	In the second case, the score is represented by the pixelwise classification accuracy~\cite{congalton2008assessing} of the images.
	
	To analyze the most appropriate score function, experiments were performed varying only this particular parameter and maintaining the remaining ones.
	Specifically, these experiments were conducted using: the Coffee dataset~\cite{keiller2016icpr}, Dilated6 network (Figure~\ref{dilated1}), uniform fixed distribution (over $25\times25$ and $50\times50$), and same hyperparameters presented in Table~\ref{tab:parameters}.
	Results can be seen in Table~\ref{tab:score_analysis}.
	Through the table, it is possible to see that both score functions achieved similar results.  
	However, since \textbf{accuracy score} is more intuitive, for the remaining of this work, results will be reported using this function.
	
	\begin{table}[]
		\centering
		\caption{Results over different score functions.}
		\label{tab:score_analysis}
		\resizebox{\columnwidth}{!}{
			\begin{tabular}{@{}lrrrr@{}}
				\toprule
				& \multicolumn{1}{c}{\textbf{Overall Accuracy}} & \multicolumn{1}{c}{\textbf{Kappa}} & \multicolumn{1}{c}{\textbf{Average Accuracy}} & \multicolumn{1}{c}{\textbf{F1 Score}} \\ \midrule
				\textbf{Accuracy} & 86.27$\pm$1.44                                & 69.41$\pm$2.01                     & 84.85$\pm$1.66                                & 84.62$\pm$1.06                        \\
				\textbf{Loss}     & 86.15$\pm$1.96                                & 69.16$\pm$3.41                     & 84.68$\pm$2.02                                & 84.49$\pm$1.76                        \\ \bottomrule
			\end{tabular}
		}
	\end{table}
	
	\subsection{Range Analysis} \label{subsec:range}
	
	Although the presented approach is proposed to select automatically the best patch size, in training time, avoiding lots of experiments to adjust such size (as done in several works~\cite{keiller2016icpr,paisitkriangkrai2016semantic,volpi2017dense}), in this section, the patch size range is analyzed in order to examine the robustness of the method.
	
	This range is evaluated on all datasets, except Potsdam.
	Such dataset is very similar to Vaihingen one and, therefore, analysis and decisions made over the latter dataset are also applicable to the Potsdam one.
	Furthermore, in order to evaluate such dataset, a validation set, created according to~\cite{volpi2017dense}, was employed.
	Experiments were conducted varying only the patch size range but maintaining the remaining configurations.
	Particularly, the experiments employed the same hyperparameters (presented in Table~\ref{tab:parameters}), Dilated6 network (Figure~\ref{dilated1}), and uniform fixed distribution.
	
	Table~\ref{tab:range} presents the obtained results.
	Each dataset was evaluated over several ranges, selected based on previous works~\cite{keiller2016icpr,volpi2017dense}.
	Specifically, each dataset was evaluated in a large range (comprising from small to large sizes) and subsets of such range.
	Table~\ref{tab:range} also presents the most selected patch size (for the testing phase) for each experiment, giving some insights about how the proposed method behaves during such step.

	For the Coffee dataset~\cite{keiller2016icpr}, obtained results are all very similar making it difficult to define a better or worse range. 
	Hence, any patch size range could be selected for further experiments, showing the robustness of the proposed algorithm which yielded similar results independently of the patch size range.
	Because of processing time (smaller patches are processed faster), in this case, patch size range $25,50$ was selected and used in all further experiments.
	
	For remaining datasets, a specific range achieved the best result.
	For the GRSS Data Fusion dataset~\cite{liao2015processing}, the best result was obtained when considering the largest range ($7,14,21,28,35,42,49,56,63,70$), i.e., the range varying from small to large patch sizes.
	For Vaihingen~\cite{vaihingen}, the intermediate range ($45,55,65,75,85$) achieved the best result.
	Therefore, in these cases, such ranges were selected and used in the remaining experiments of this work.
	However, as can be seen through Table~\ref{tab:range}, other ranges also produce competitive results and could be selected and used without significant loss of performance, which confirms the robustness of the proposed method in relation to the patch size range, allowing it to process images without the need of experimentally searching for the best patch size configuration.
	
	In terms of patch size selection (during the inference phase), the algorithm really varies depending on the experiment.
	For the Coffee dataset, the most selected patch sizes were 50 and 75, showing a trend towards such interval.
	For the remaining datasets, larger patches were favored in our experiments.
	This may be justified by the fact that urban areas have complex interactions and larger patches allow the network to capture more information about the context.
	Though the best patch size is really dependent on the experiment, current results showed that the proposed approach is able to learn and select the best patch size in processing time producing interesting outcomes when compared to state-of-the-art works, a fact reconfirmed in Section~\ref{subsec:state}.
	
	\begin{table}[h]
		\centering
		\caption{Results of the proposed approach when varying the input range of patch sizes.
			For Vaihingen, a validation set (created according~\cite{volpi2017dense}) is employed.
			Bold patch size ranges were selected for all further experiments.}
		\label{tab:range}
		\resizebox{\columnwidth}{!}{
			\begin{tabular}{@{}ccccc@{}}
				\toprule
				\multicolumn{1}{c}{\textbf{Datasets}} & \multicolumn{1}{c}{\textbf{\begin{tabular}[c]{@{}cc@{}}Patch\\Size\\Range\end{tabular}}} & 
				\multicolumn{1}{c}{\textbf{\begin{tabular}[c]{@{}cc@{}}Most\\Selected\\Size\end{tabular}}} & \multicolumn{1}{c}{\textbf{\begin{tabular}[c]{@{}c@{}}Overall\\Accuracy\end{tabular}}} &  \multicolumn{1}{c}{\textbf{\begin{tabular}[c]{@{}c@{}}Average\\Accuracy\end{tabular}}} \\ \midrule
				\multirow{4}{*}{\textbf{Coffee}}    & \textbf{25,50}                        & 50                                              & 86.27$\pm$1.44                       & 84.85$\pm$1.66       \\
				& 50,75                                 & 50                                              & 87.32$\pm$1.82                              & 85.59$\pm$1.59                        \\
				& 75,100                                & 75                                              & 86.07$\pm$1.95                                 & 85.91$\pm$1.68                                \\
				& 25,50,75,100,125                     & 75                                              & 87.11$\pm$1.74                          & 85.17$\pm$1.52                              \\ \midrule
				\multirow{4}{*}{\textbf{GRSS}}      & 7,14,21,28,35                         & 35                                              & 87.93                                    & 85.87                                      \\
				& 28,35,42,49,56                        & 49                                              & 87.71                                  & 85.26                                      \\
				& 42,49,56,63,70                        & 70                                              & 88.33                                        & 88.04                                     \\
				& \textbf{7,14,21,28,35,42,49,56,63,70} & 70                                              & 90.10                                         & 90.13                                        \\ \midrule
				\multirow{3}{*}{\textbf{Vaihingen}} & 25,45,55,65                           & 65                                              & 86.60                                            & 71.03                                      \\
				& \textbf{45,55,65,75,85}               & 85                                              & 88.66                                      & 71.96                                     \\
				& 25,45,55,65,85,95,100                 & 95                                              & 87.44                               & 71.30                                    \\ \bottomrule
			\end{tabular}
		}
	\end{table}
	
	\subsection{Convolution Operation Analysis} \label{subsec:conv_op}
	
	Although the proposed networks use dilated convolutions, it is possible to recreate such architectures using standard convolution operations.
	As introduced in Section~\ref{sec:background}, the only difference between these convolution operations is the possibility to have gaps between the filter weights, a special characteristic of the dilated convolutions~\cite{YuKoltun2016}.
	Such aspect makes all the difference since dilated convolution can expand the exploited context (by enlarging the filter weights) without increasing the number of parameters, while standard convolutions are not able to do this since the filters are always grouped (without gaps).
	This is a great advantage since a deeper network composed of standard convolution operations (without any downsample or upsample operation) would require more layers in order to aggregate a large context, while a network composed of dilated convolutions can expand the context without increasing the number of parameters, requiring fewer layers.
	
	In order to demonstrate this advantage of dilated convolutions over standard ones, we performed experiments comparing two networks that have exactly the same architecture (Dilated6 -- Figure~\ref{dilated1}) but differ in the convolution operation: while one network uses dilated convolutions, the second architecture employs the standard operation.
	Since the Dilated6 network does not have pooling layers, the comparison between these networks is totally fair, given that the only difference is the convolution operation type.
	All datasets were used in this experiment, except Potsdam.
	This is because the Vaihingen and Potsdam datasets are very similar and analysis performed over one can also be extended to the other.
	A validation set, created according to~\cite{volpi2017dense}, was used to evaluate the Vaihingen dataset.
	Experiments were executed preserving all configurations and varying only the convolution type.
	Particularly, the configuration was defined taken into account previous experiments, i.e., it uses uniform fixed distribution, patch ranging according to Section~\ref{subsec:range}, accuracy as score function, and hyperparameters presented in Table~\ref{tab:parameters}.
	
	Results can be seen in Table~\ref{tab:convolution_type}.
	Overall, architectures based on dilated convolution outperformed the networks that employ the standard operation.
	Since the only difference between the networks is the convolution (and, consequently, the exploited context), these results show the advantage of the dilated operations over the standard one.
	
	\begin{table}[]
		\centering
		\caption{Results of the Dilated6 network trained using distinct convolution types.}
		\label{tab:convolution_type}
		\begin{tabular}{@{}ccrr@{}}
			\toprule
			\textbf{Datasets}                   & \textbf{\begin{tabular}[c]{@{}c@{}}Convolution\\ Type\end{tabular}} & \multicolumn{1}{c}{\textbf{\begin{tabular}[c]{@{}c@{}}Overall\\ Accuracy\end{tabular}}} & \multicolumn{1}{c}{\textbf{\begin{tabular}[c]{@{}c@{}}Average\\ Accuracy\end{tabular}}} \\ \midrule
			\multirow{2}{*}{\textbf{Coffee}}    & \textbf{Standard}                                                   & 84.13$\pm$1.28	                                                                                    & 82.97$\pm$0.48                                                                                   \\
			& \textbf{Dilated}                                                    & 86.27$\pm$1.44                                                                          & 84.85$\pm$1.66                                                                          \\ \midrule
			\multirow{2}{*}{\textbf{GRSS}}      & \textbf{Standard}                                                   & 85.70	                                                                                    & 85.31                                                                                  \\
			& \textbf{Dilated}                                                    & 90.10                                                                                   & 90.13                                                                                   \\ \midrule
			\multirow{2}{*}{\textbf{Vaihingen}} & \textbf{Standard}                                                   & 86.13	                                                                                    & 69.65                                                                                    \\
			& \textbf{Dilated}                                                    & 88.66                                                                                   & 71.96                                                                                   \\ \bottomrule
		\end{tabular}
	\end{table}
	
	\subsection{Convergence Analysis} \label{subsec:convergence}
	
	In this section, we analyze the convergence of the proposed technique. 
	Figure~\ref{fig:convergence} presents the convergence of the datasets using the Dilated6 network, accuracy as score function, uniform fixed distribution, and hyperparameters presented in Table~\ref{tab:parameters}.
	According to the figure, the loss and accuracy vary significantly at the beginning of the process but, with the reduction of the learning rate, the networks converge independently of the use of distinct patch sizes.
	Moreover, the test/validation accuracy (green line) converges and stabilizes showing that the networks can learn to extract features from patches of multiple sizes while selecting the best patch size for testing.
	
	\newcommand{\convergenceFigSize}{0.24}
	\begin{figure*}[h]
		\centering
		\subfloat[Coffee dataset -- fold 1]{
			\includegraphics[width=\convergenceFigSize\textwidth]{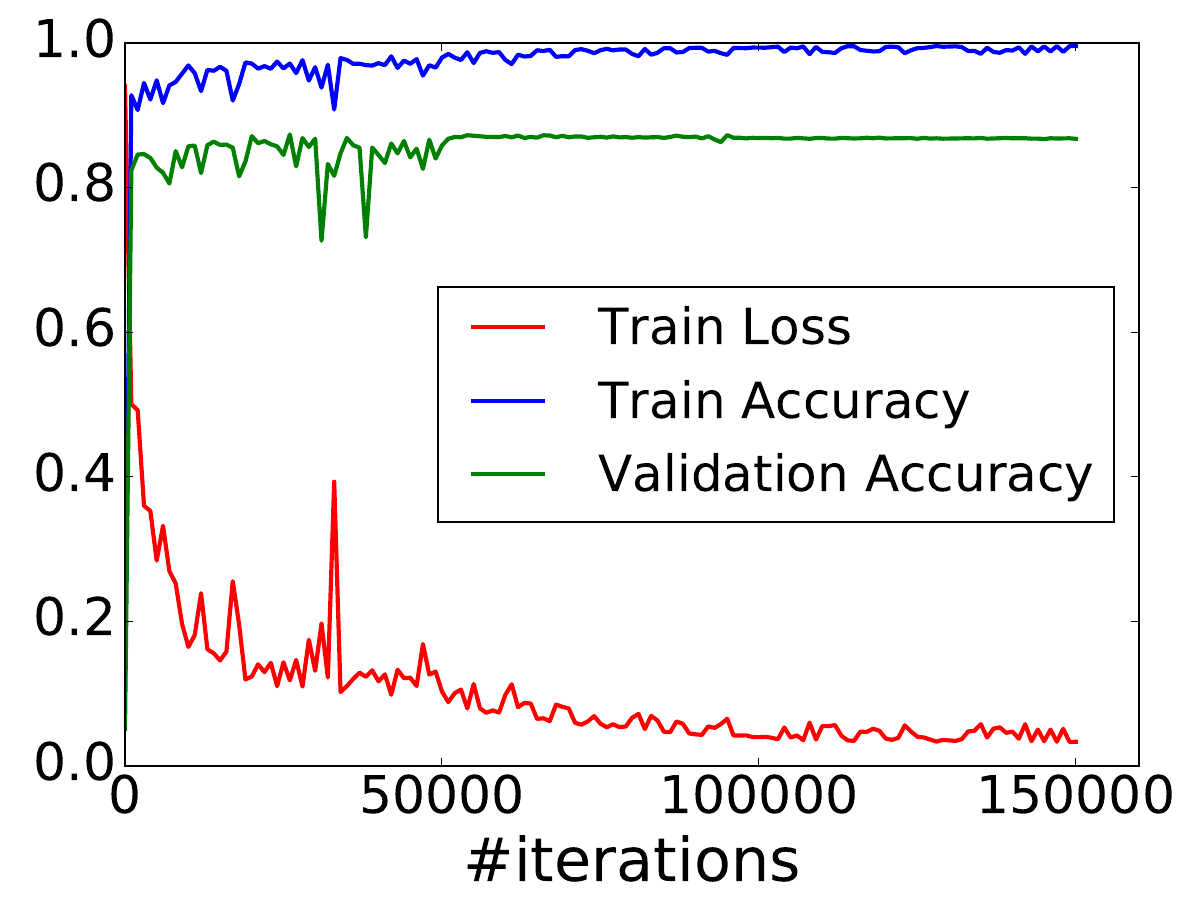}
			\label{coffee_convergence}
		}
		\subfloat[GRSS Data Fusion dataset]{
			\includegraphics[width=\convergenceFigSize\textwidth]{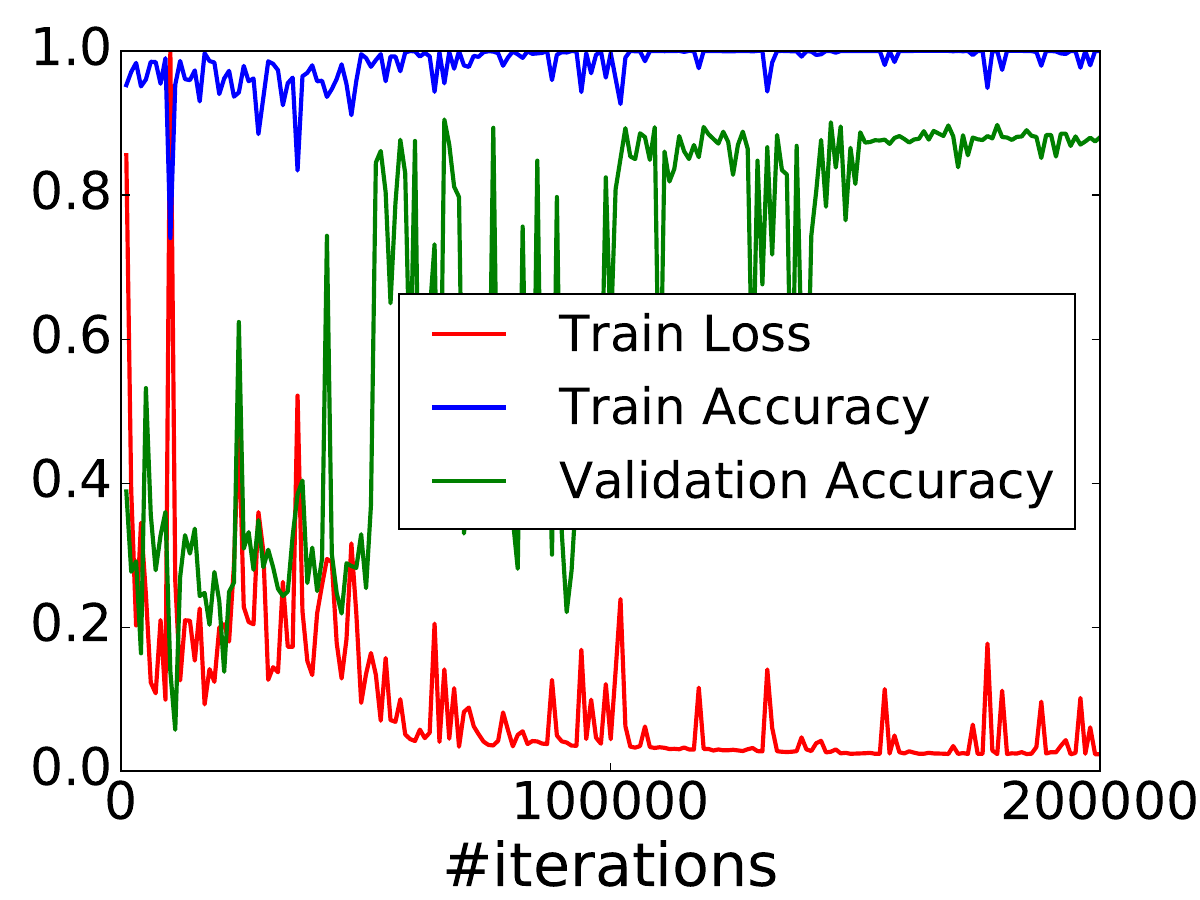}
			\label{contest_convergence}
		}
		\subfloat[Vaihingen dataset]{
			\includegraphics[width=\convergenceFigSize\textwidth]{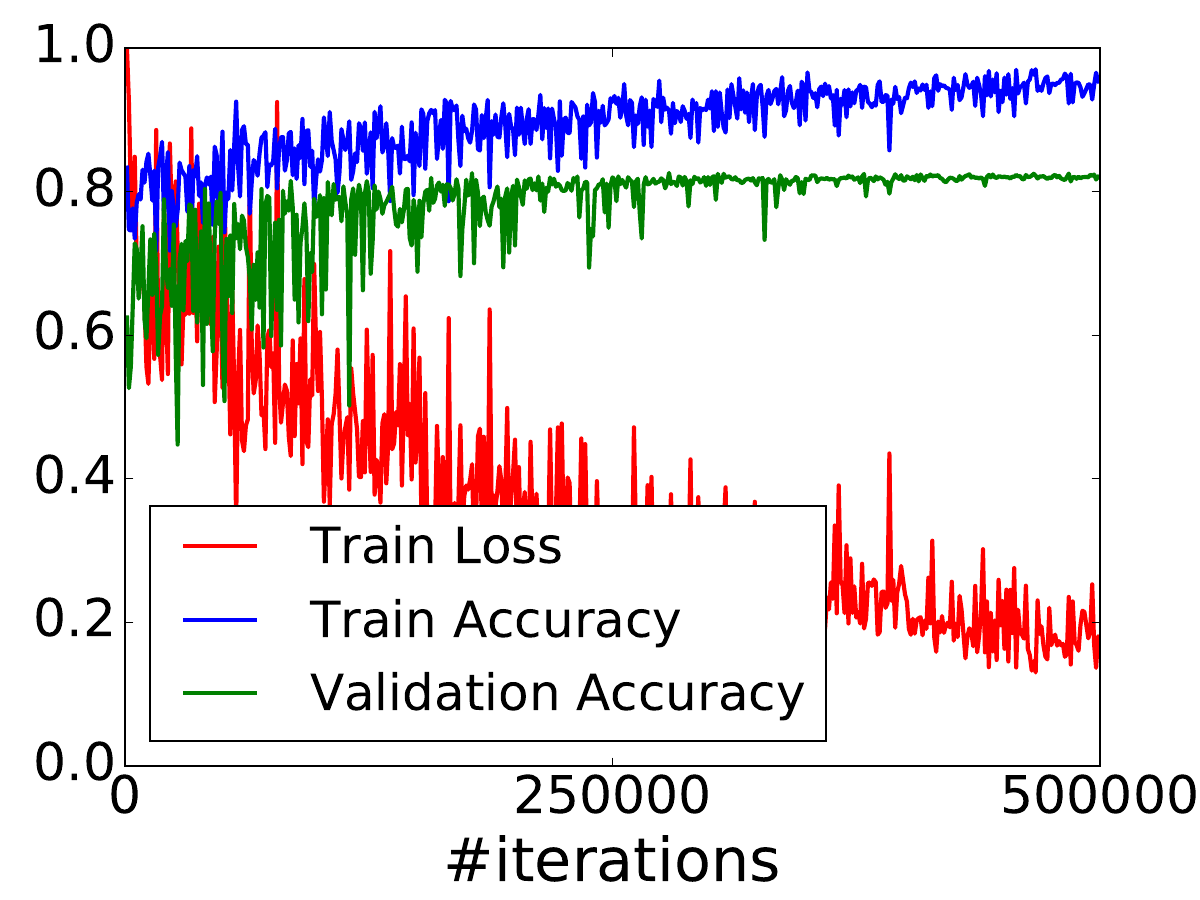}
			\label{vaihingen_convergence}
		}
		\subfloat[Potsdam dataset]{
			\includegraphics[width=\convergenceFigSize\textwidth]{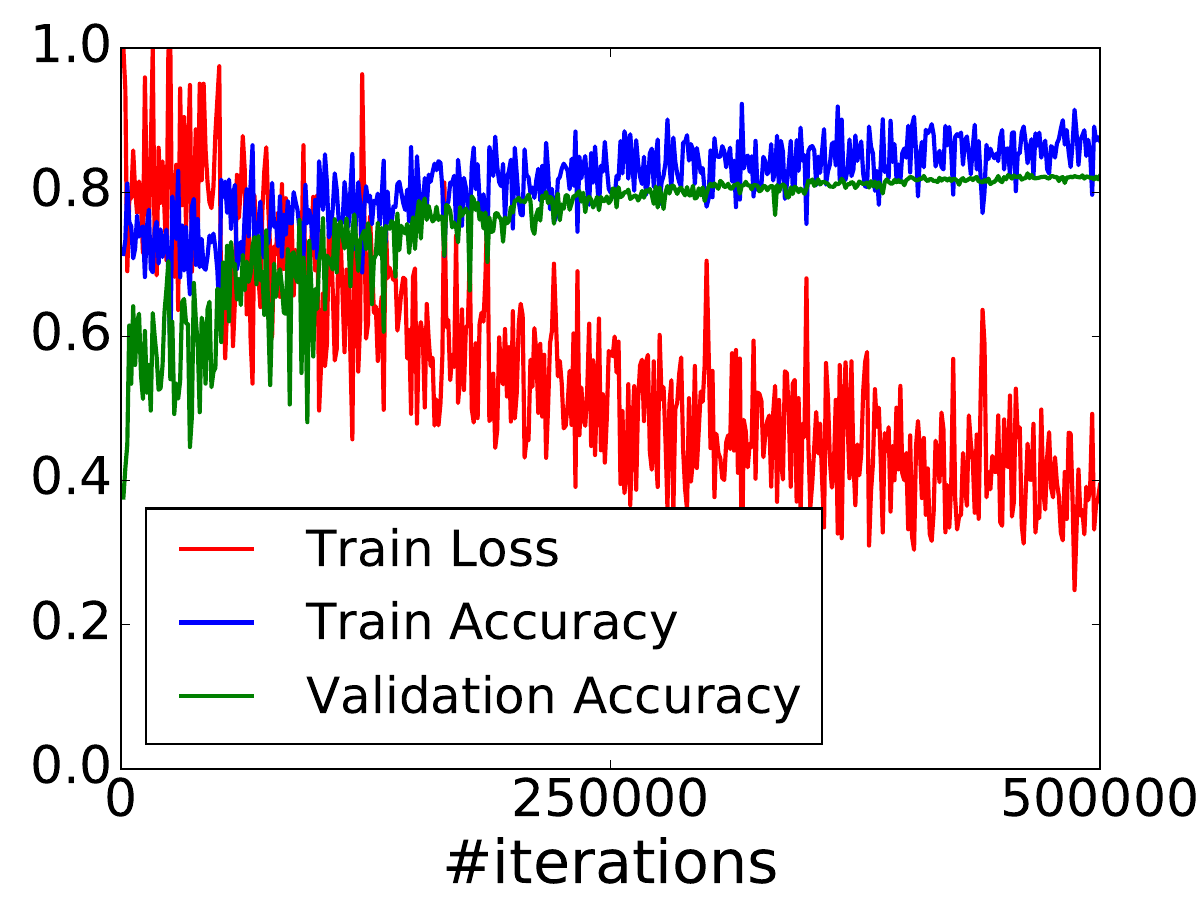}
			\label{potsdam_convergence}
		}
		\caption{Convergence of Dilated6 network for all datasets.
			For the Coffee dataset, only the fold 1 is reported.
			For Vaihingen and Potsdam datasets, the validation set (created according~\cite{volpi2017dense}) is reported.}
		\label{fig:convergence}
	\end{figure*}
	
	\subsection{Performance Analysis} \label{subsec:dilated_comparison}
	
	To analyze the efficiency, in terms of performance and processing time, of the proposed algorithm, several experiments were conducted comparing the same network trained using two distinct methods:
	(i) the \textbf{traditional} training process~\cite{deeplearningbook}, in which the network is trained using patches of constant size, without any variation.
	This method is the standard one when it comes to neural networks and is the most exploited in the literature for training deep learning-based techniques.
	Also, this is the approach that is used to empirically selects the best patch size, which is traditionally done by training several networks, one for each considered patch.
	(ii) the proposed \textbf{dynamic} training process, in which the network is trained with patches of varying size.
	
	Two datasets were selected to be evaluated using these training strategies:
	(i) the GRSS Data Fusion dataset, which has the largest patch size range (according to Section~\ref{subsec:range}) allowing a better comparison between the training strategies, and 
	(ii) Vaihingen dataset, which is very similar to Potsdam one and, therefore, allows the conclusions to be applied to this one.
	To evaluate this dataset, a validation set, created according~\cite{volpi2017dense}, was employed.
	
	Specifically, in these experiments, Dilated6 network (Figure~\ref{dilated1}) is trained using both strategies.
	For the proposed dynamic training process, previous experiments were taken into account, i.e., it uses uniform fixed distribution, patch ranging according to Section~\ref{subsec:range}, accuracy as score function, and hyperparameters presented in Table~\ref{tab:parameters}.
	Concerning the traditional training process, several networks (with same architecture) were trained using each of the possible patch sizes.
	
	Results of these experiments are presented in Table~\ref{tab:training_time}.
	For both datasets, networks trained with the proposed approach outperform the models trained with the traditional training process (independently of the patch size), showing the ability of the proposed method to capture multi-context information from patches of distinct size which improve the performance of the final model. 
	Also, on average, the processing time of the proposed method is lower than the traditional training process, in which the computational time increases with the increase of the patch size, an expected behavior given that the convolution process using large inputs takes more time than using smaller ones.
	
	Specifically, for the GRSS Data Fusion dataset, considering only models trained with the traditional method, the best result is achieved by the network using patches of $70\times70$ pixels.
	This ConvNet took around 160 hours to train using 200,000 iterations and achieved 86.93\% of average accuracy.
	However, the model trained with the proposed dynamic process outperforms this result while taking less time to train.
	Particularly, Dilated6 network trained using the dynamic process produced 90.13\% of average accuracy while taking around 81 hours to train.
	This improvement in the performance is due to the exploitation of distinct contexts provided by different patch sizes during the training procedure.
	This process of using distinct patch sizes also speeds up the training, given that small patches (which are processed faster) are also used together with large ones.
	
	The same conclusions hold for the Vaihingen dataset.
	Precisely, the best result using the traditional method is achieved by the network trained with patches of $85\times85$ pixels.
	This ConvNet took around 325 hours to train using 500,000 iterations and achieved 66.96\% of average accuracy.
	However, this result was outperformed by the network trained using the proposed dynamic strategy, while taking less training time.
	Such model produced 71.96\% of average accuracy while taking around 220 hours to train.
	
	Moreover, the proposed dynamic strategy has another advantage: while the empirical method would require training several networks in order to select the best patch size, resulting in a greater computational time, the proposed strategy combines all patch sizes during the training stage while selecting the best size for the inference phase, requiring only one full procedure to achieve its final result.
	Hence, overall, the proposed method requires less training time than the empirical approach, while achieving better results.
	
	\begin{table}[]
		\centering
		\caption{Comparison between the dilated network trained using the proposed and the traditional method.}
		\label{tab:training_time}
		\resizebox{\columnwidth}{!}{
			\begin{tabular}{@{}cccrr@{}}
				\toprule
				\textbf{Dataset}                           & \textbf{\begin{tabular}[c]{@{}c@{}}Training\\Process\end{tabular}} & \multicolumn{1}{c}{\textbf{Patch Size}} & \multicolumn{1}{c}{\textbf{\begin{tabular}[c]{@{}c@{}}Training\\Time (hours)\end{tabular}}} & 
				\textbf{\begin{tabular}[c]{@{}c@{}}Average\\Accuracy\end{tabular}} \\
				\midrule
				\multirow{7}{*}{\textbf{GRSS Data Fusion}} & \multirow{4}{*}{\textbf{Traditional}}                               & 7                                       & 35 & 83.33  \\
				&                                                                     & 28                                      & 59 & 85.48 \\
				&                                                                     & 49                                      & 86 & 85.94 \\
				&                                                                     & 70                                      & 160 & 86.93 \\
				\cmidrule{2-5}
				& \textbf{Dynamic}                                                    & \begin{tabular}[c]{@{}c@{}}7,14,21,\\ 28,35,42,49,\\ 56,63,70\end{tabular}            & 81       & 90.13                                               \\
				\midrule
				\multirow{6}{*}{\textbf{Vaihingen}}        & \multirow{5}{*}{\textbf{Traditional}}                               & 45                                      & 125        & 66.29                                               \\
				&                                                                     & 55                                      & 160        & 66.77                                                \\
				&                                                                     & 65                                      & 200       & 66.84                                               \\
				&                                                                     & 75                                      & 260       & 66.65                                               \\
				&                                                                     & 85                                      & 325       & 66.96                                                \\
				\cmidrule{2-5}
				& \multicolumn{1}{l}{\textbf{Dynamic}}                                & 45,55,65,75,85                          & 220       & 71.96                                               \\        
				\bottomrule
			\end{tabular}
		}
	\end{table}
	
	\subsection{State-of-the-art Comparison} \label{subsec:state}
	
	\subsubsection{Coffee dataset} \label{subsec:coffee_state_results}
	
	Using analysis performed on previous sections, we have conducted several experiments over the Coffee dataset~\cite{keiller2016icpr}.
	Results of these experiments, as well as the state-of-the-art baselines, are presented in Table~\ref{tab:coffe_results}.
	In order to allow a visual comparison, prediction maps for the Coffee dataset using different networks trained with the proposed method are presented in Figure~\ref{fig:coffee_results}.

	Overall, all baselines produced similar results.
	While the pixelwise network~\cite{keiller2016icpr} yielded a slightly worse result with a higher standard deviation, all other baselines reached basically the same level of performance, with a smaller standard deviation.
	This may be justified by the fact that the pixelwise network does not learn much information about the pixel interaction (since each pixel is processed independently), while the other methods process and classify a set of pixels simultaneously.
	Because of the similar results, all baselines are comparable.
	
	This same behavior may be seen among the networks trained with the proposed methodology.
	Although these networks achieved comparable results, such models outperformed the baselines.
	Furthermore, the Dilated6Pooling trained with the proposed dynamic method produced better results than the same network trained with traditional training process (mainly in the Kappa Index).
	These results show the effectiveness of the proposed technique that produces state-of-the-art outcomes by capturing multi-context information while selecting the best patch size, two great advantages when compared to the traditional training process.

	\begin{table}[h]
		\centering
		\caption{Results for the Coffee dataset.}
		\label{tab:coffe_results}
		\resizebox{\columnwidth}{!}{
			\begin{tabular}{@{}clrr@{}}
				\toprule
				\multicolumn{1}{c}{\textbf{\begin{tabular}[c]{@{}c@{}}Training\\Process\end{tabular}}} & \multicolumn{1}{c}{\textbf{Network}} &
				\multicolumn{1}{c}{\textbf{\begin{tabular}[c]{@{}c@{}}Average\\Accuracy\end{tabular}}} & \multicolumn{1}{c}{\textbf{Kappa}} \\
				\midrule
				\multirow{5}{*}{\textbf{Traditional}} & Pixelwise~\cite{keiller2016icpr}  						& 81.72$\pm$2.38 	& 	62.75$\pm$7.42	\\
				& CCNN~\cite{nogueira2015coffee}  						& 82.80$\pm$2.30 	& 	64.60$\pm$4.34	\\
				& FCN~\cite{long2015fully}  		  						& 83.25$\pm$2.47 	& 	66.00$\pm$3.55	\\
				& Deconvolution Network~\cite{badrinarayanan2015segnet} 	& 82.61$\pm$2.05 	& 	65.56$\pm$3.47	\\
				& Dilated network (Dilated6Pooling)						& 82.52$\pm$1.14 	& 	66.14$\pm$2.27	\\
				\midrule
				\multirow{4}{*}{\textbf{Dynamic}}  & Dilated6	  											& 84.79$\pm$1.66	&	69.41$\pm$2.01	\\
				& DenseDilated6			  								& 85.88$\pm$2.34	&	71.51$\pm$2.74	\\
				& Dilated6Pooling								    	& 85.77$\pm$1.74	&	72.27$\pm$1.38	\\
				& Dilated8Pooling								   		& 86.67$\pm$1.39	&	73.78$\pm$1.87	\\
				\bottomrule
			\end{tabular}
		}
	\end{table}
	
	\newcommand{\exCoffee}{0.15}
	\begin{table*}[t]
		\begin{center}
			\begin{tabular}{>{\centering\arraybackslash}m{2.5cm} >{\centering\arraybackslash}m{2.5cm} >{\centering\arraybackslash}m{2.5cm} >{\centering\arraybackslash}m{2.5cm} >{\centering\arraybackslash}m{2.5cm} >{\centering\arraybackslash}m{2.5cm} @{}m{0pt}@{} } 
				\textbf{Original Image} & \textbf{Ground-Truth} & \textbf{Dilated6} & \textbf{DenseDilated6} & \textbf{Dilated6Pooling} & \textbf{Dilated8Pooling} & \\
				\cincludegraphics[width=\exCoffee\textwidth]{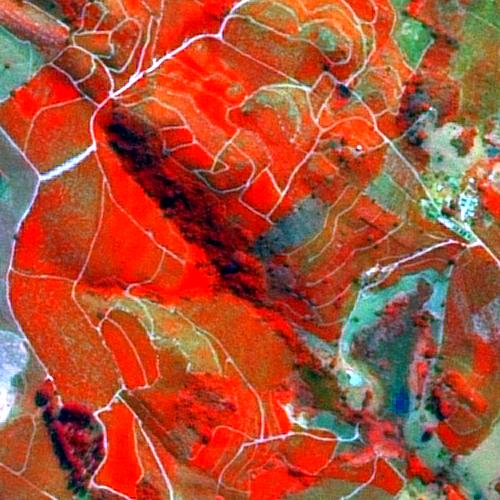} &
				\cincludegraphics[width=\exCoffee\textwidth]{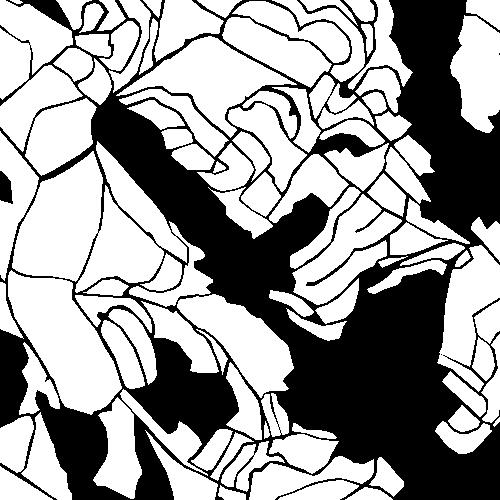} &
				\cincludegraphics[width=\exCoffee\textwidth]{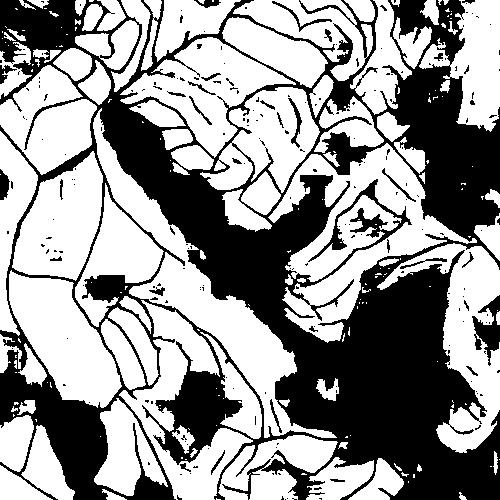} &
				\cincludegraphics[width=\exCoffee\textwidth]{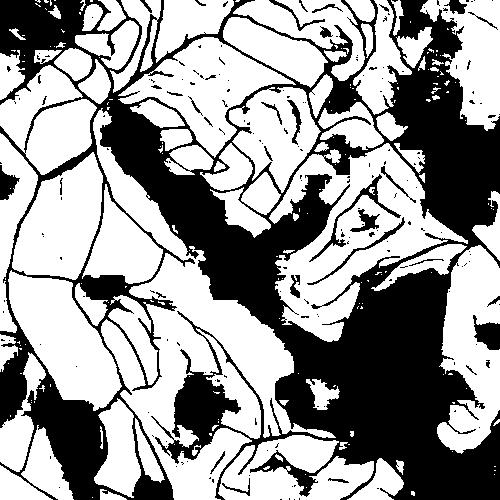} &
				\cincludegraphics[width=\exCoffee\textwidth]{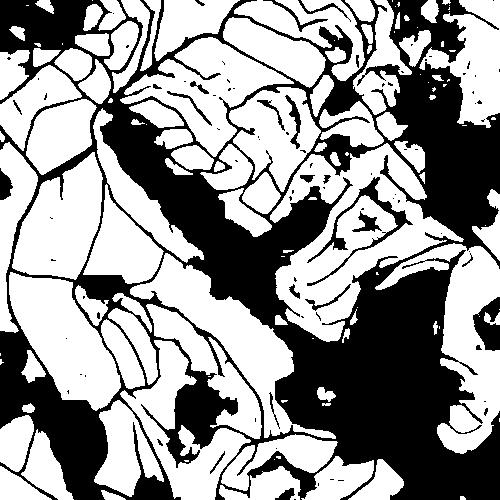} &
				\cincludegraphics[width=\exCoffee\textwidth]{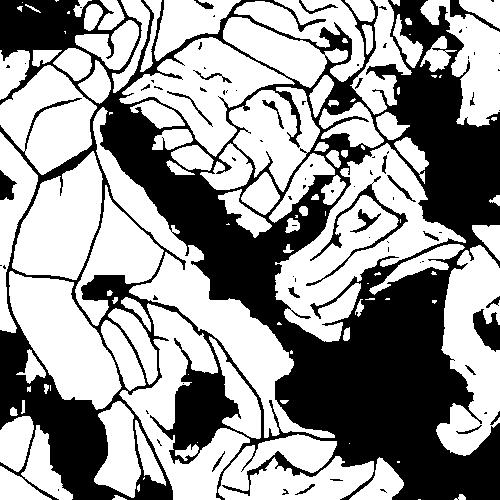} & \\[1.5cm]
				\cincludegraphics[width=\exCoffee\textwidth]{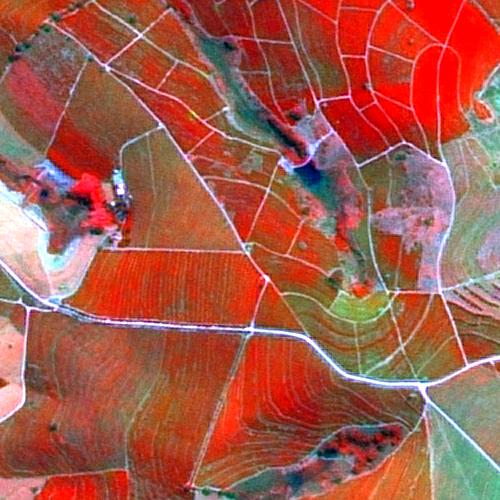} &
				\cincludegraphics[width=\exCoffee\textwidth]{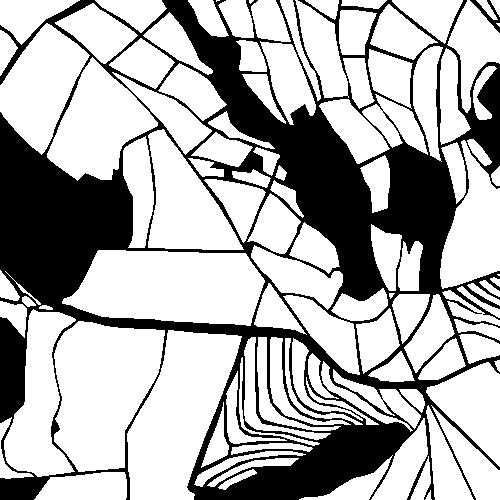} &
				\cincludegraphics[width=\exCoffee\textwidth]{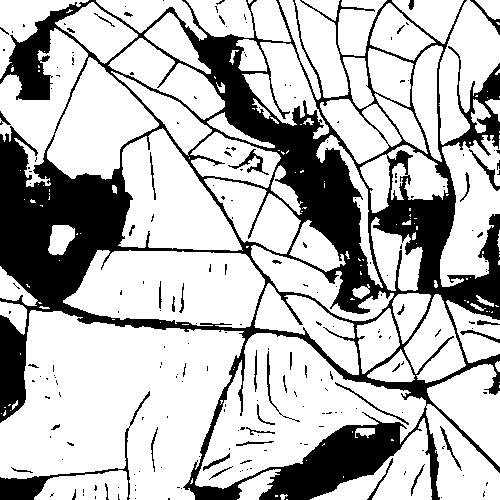} & 
				\cincludegraphics[width=\exCoffee\textwidth]{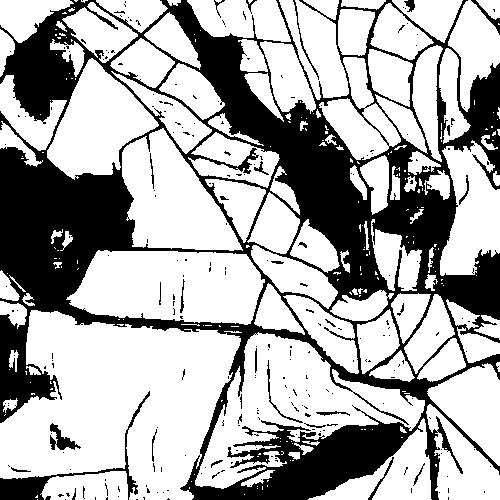} &
				\cincludegraphics[width=\exCoffee\textwidth]{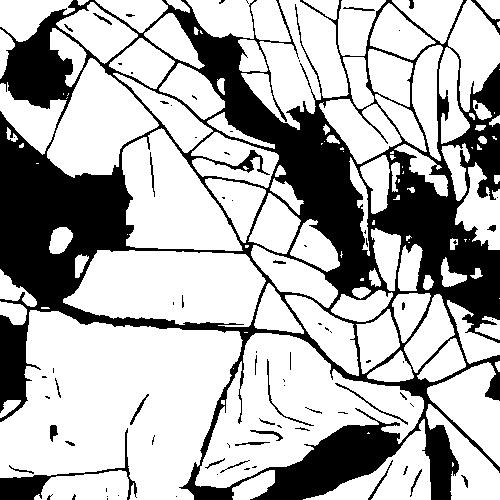} &
				\cincludegraphics[width=\exCoffee\textwidth]{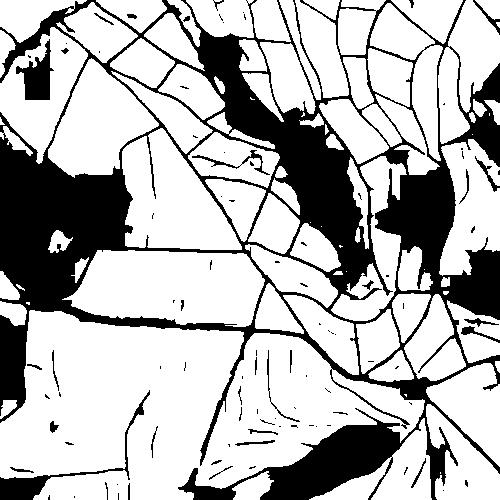} & \\[1.5cm]
			\end{tabular}
		\end{center}
		\captionof{figure}{Two images of the Coffee Dataset, their respective ground-truths and the prediction maps generated by the proposed algorithm. Legend -- White: Coffee areas. Black: Non Coffee areas.}
		\label{fig:coffee_results}
	\end{table*}
	
	\subsubsection{GRSS Data Fusion Dataset} \label{subsec:grsss_state_results}
	
\begin{figure*}[h!]
		\centering
		\subfloat[Training]{
			\includegraphics[height=0.26\textwidth]{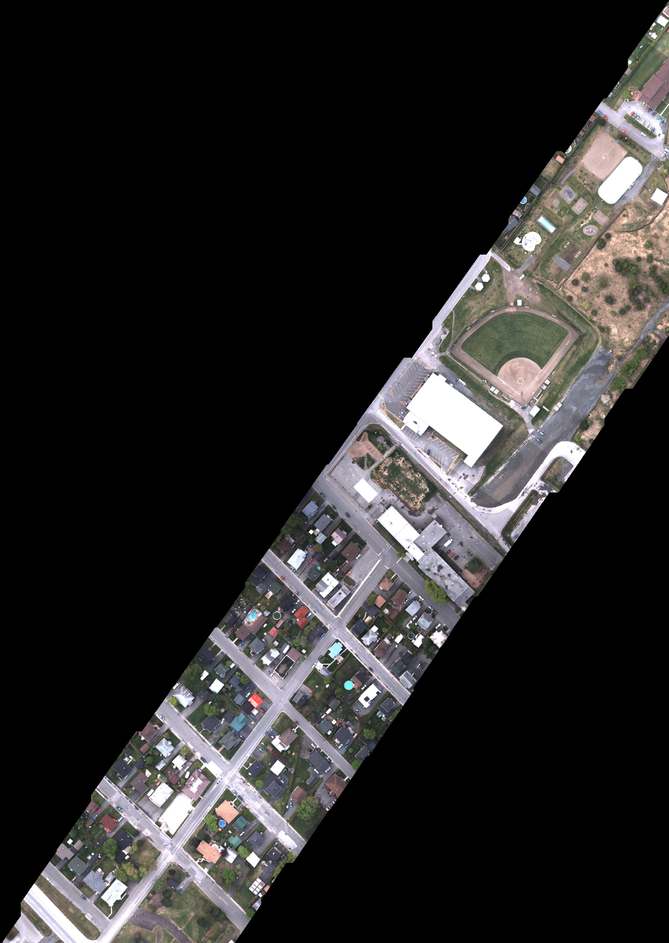}
			\includegraphics[height=0.26\textwidth]{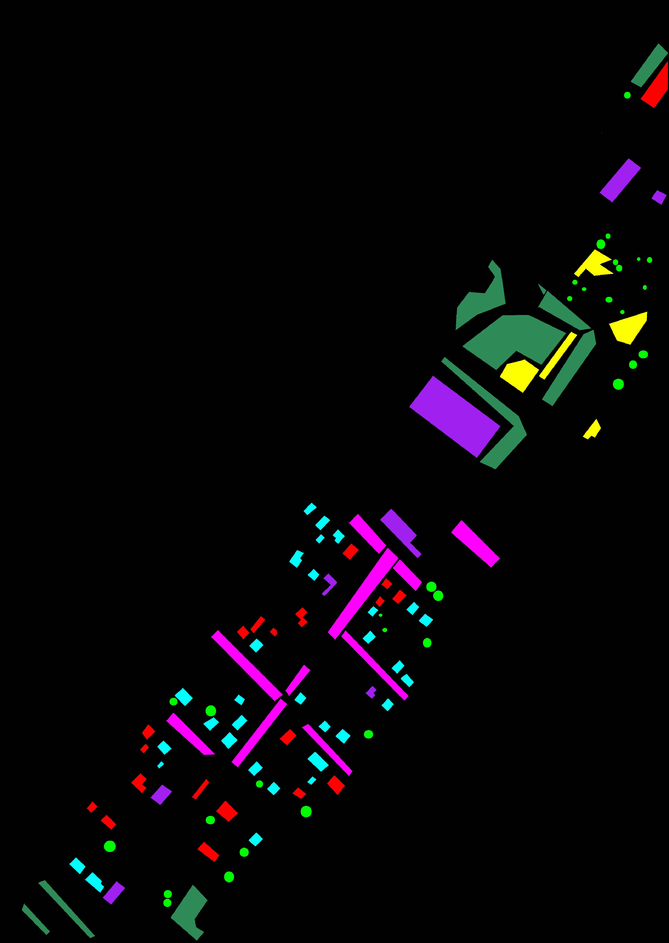}
		}
		\subfloat[Test]{
			\includegraphics[height=0.26\textwidth]{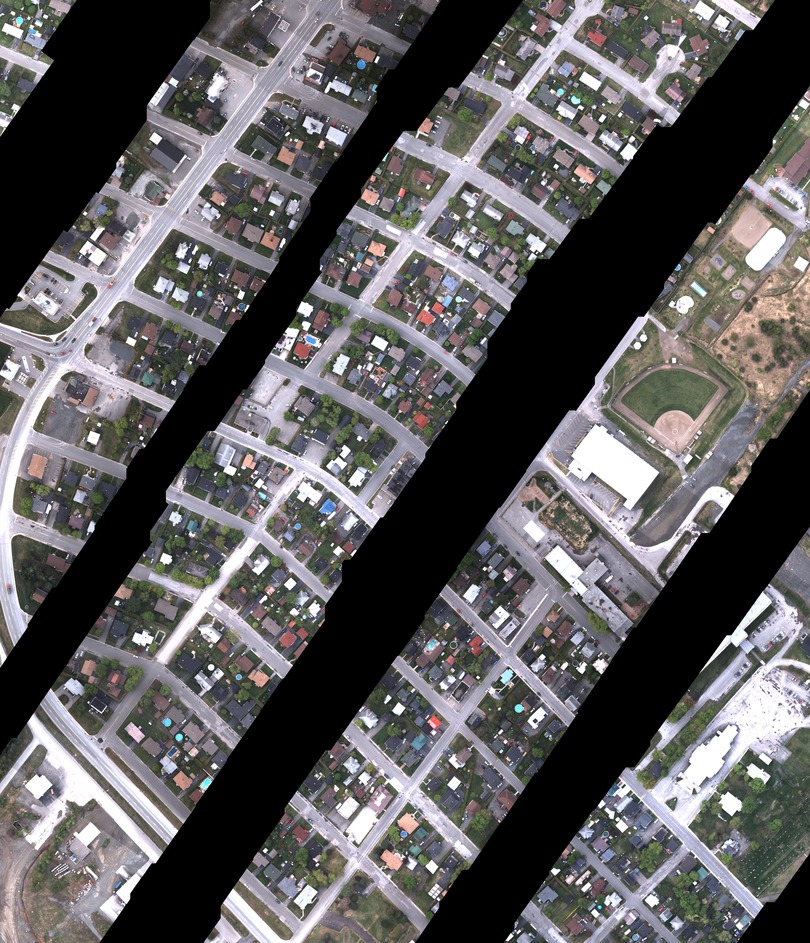}
			\includegraphics[height=0.26\textwidth]{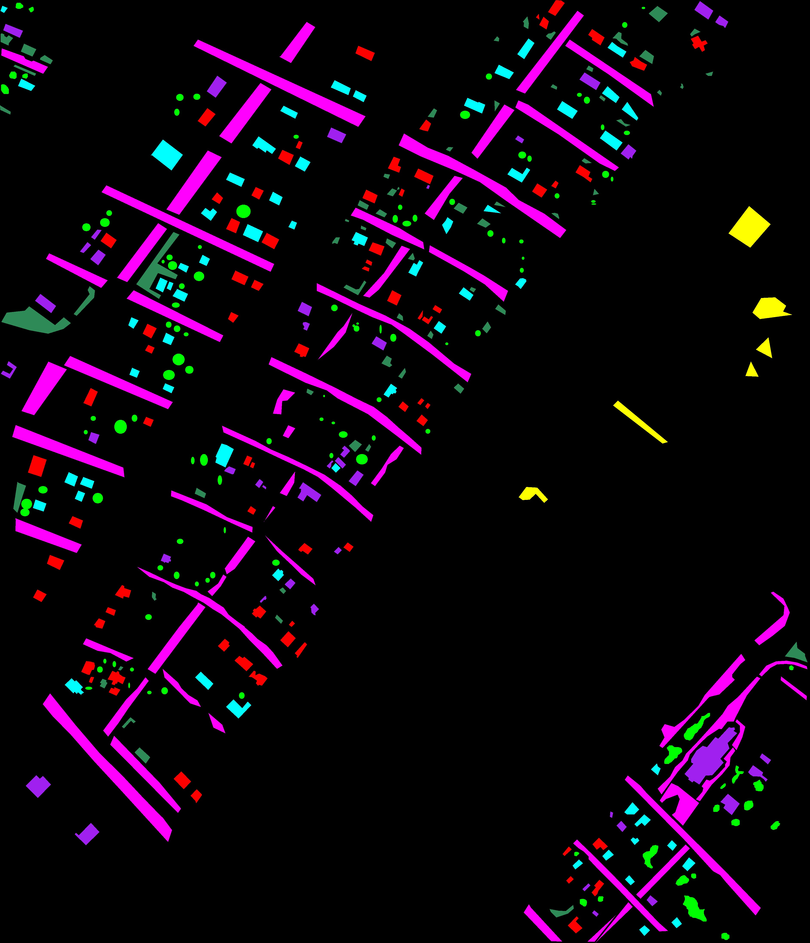}
		}
		\hspace{1em}
		\subfloat[Dilated6]{
			\includegraphics[height=0.26\textwidth]{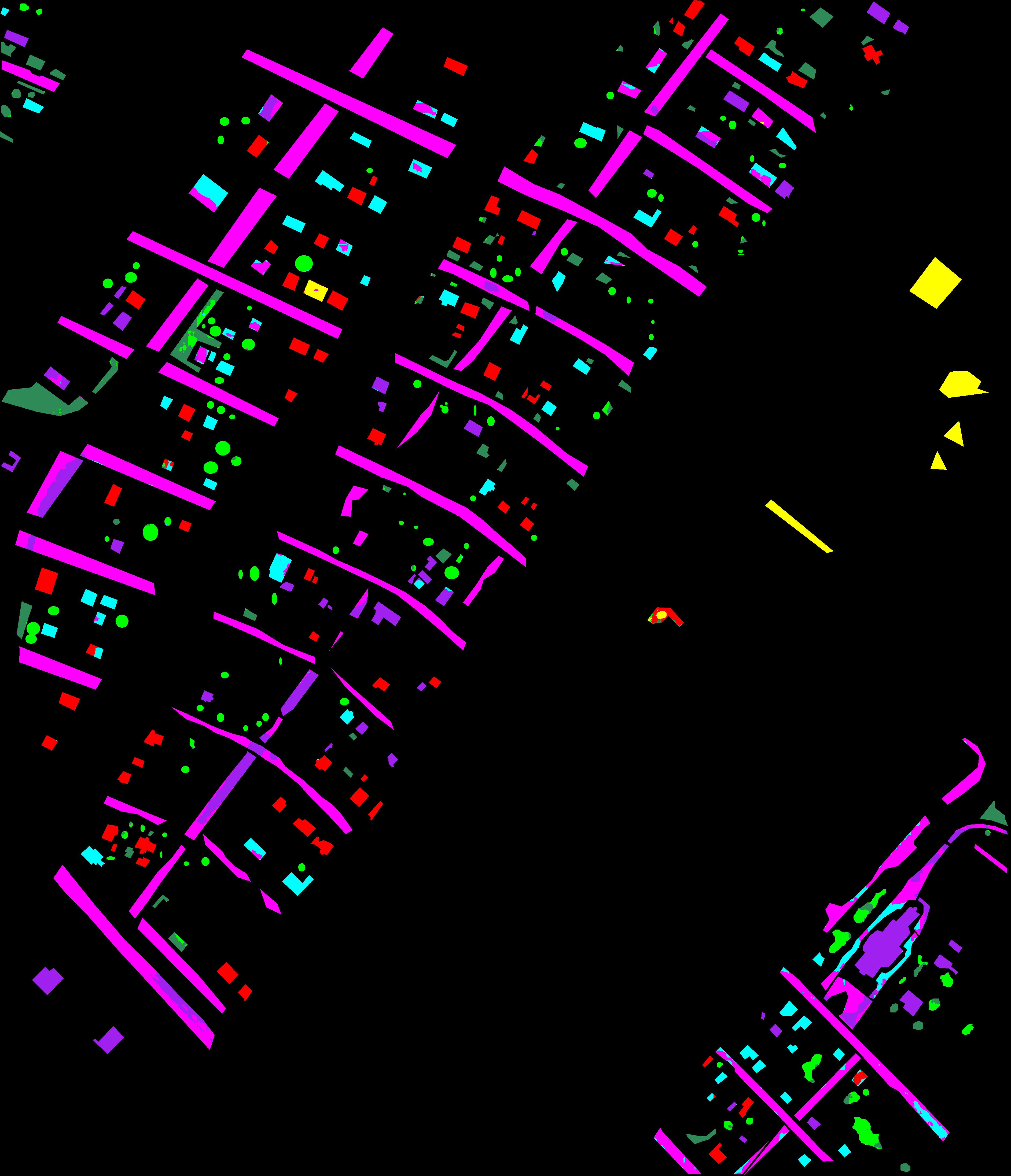}
		}
		\subfloat[DenseDilated6]{
			\includegraphics[height=0.26\textwidth]{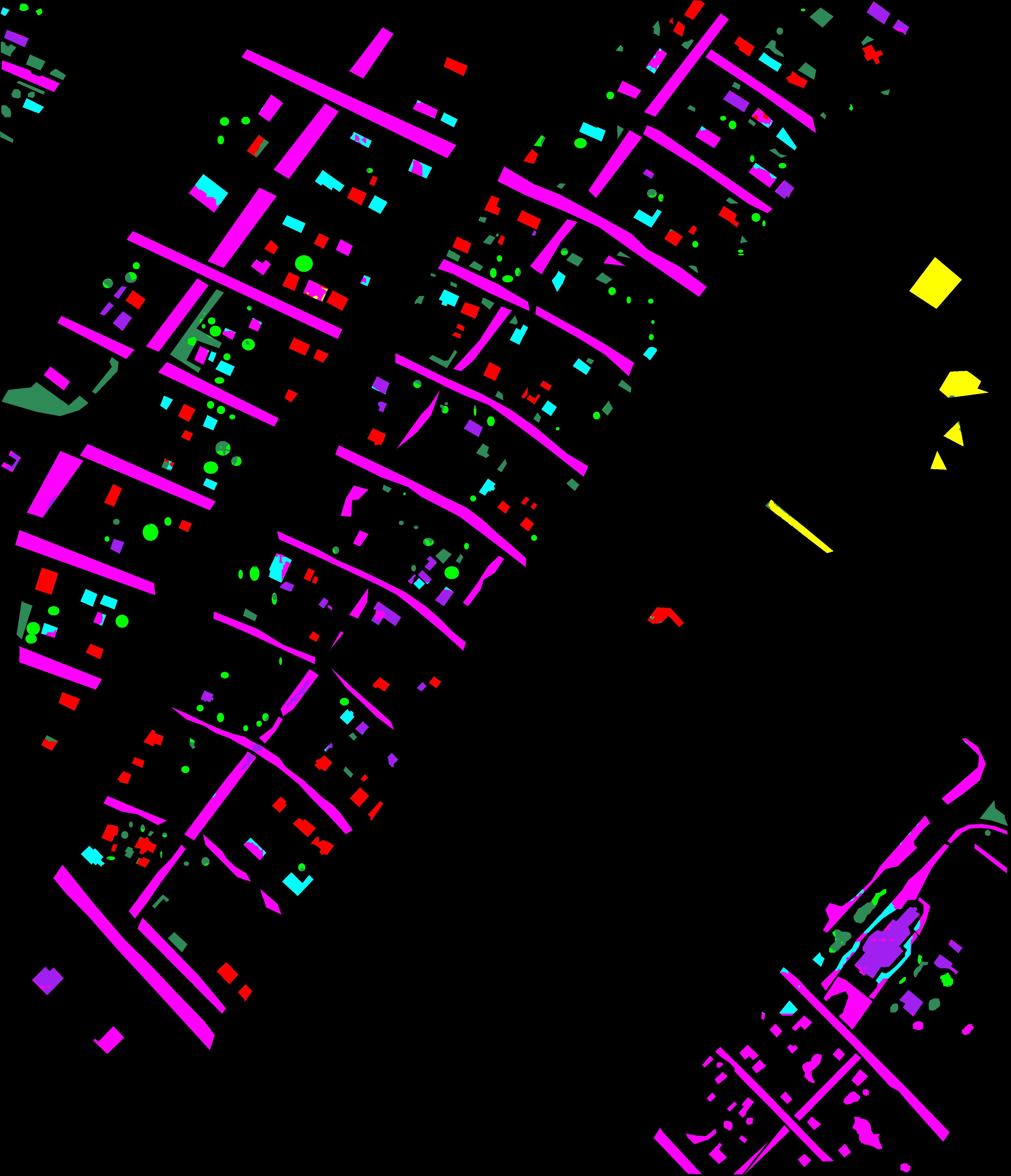}
		}
		\subfloat[Dilated6Pooling]{
			\includegraphics[height=0.26\textwidth]{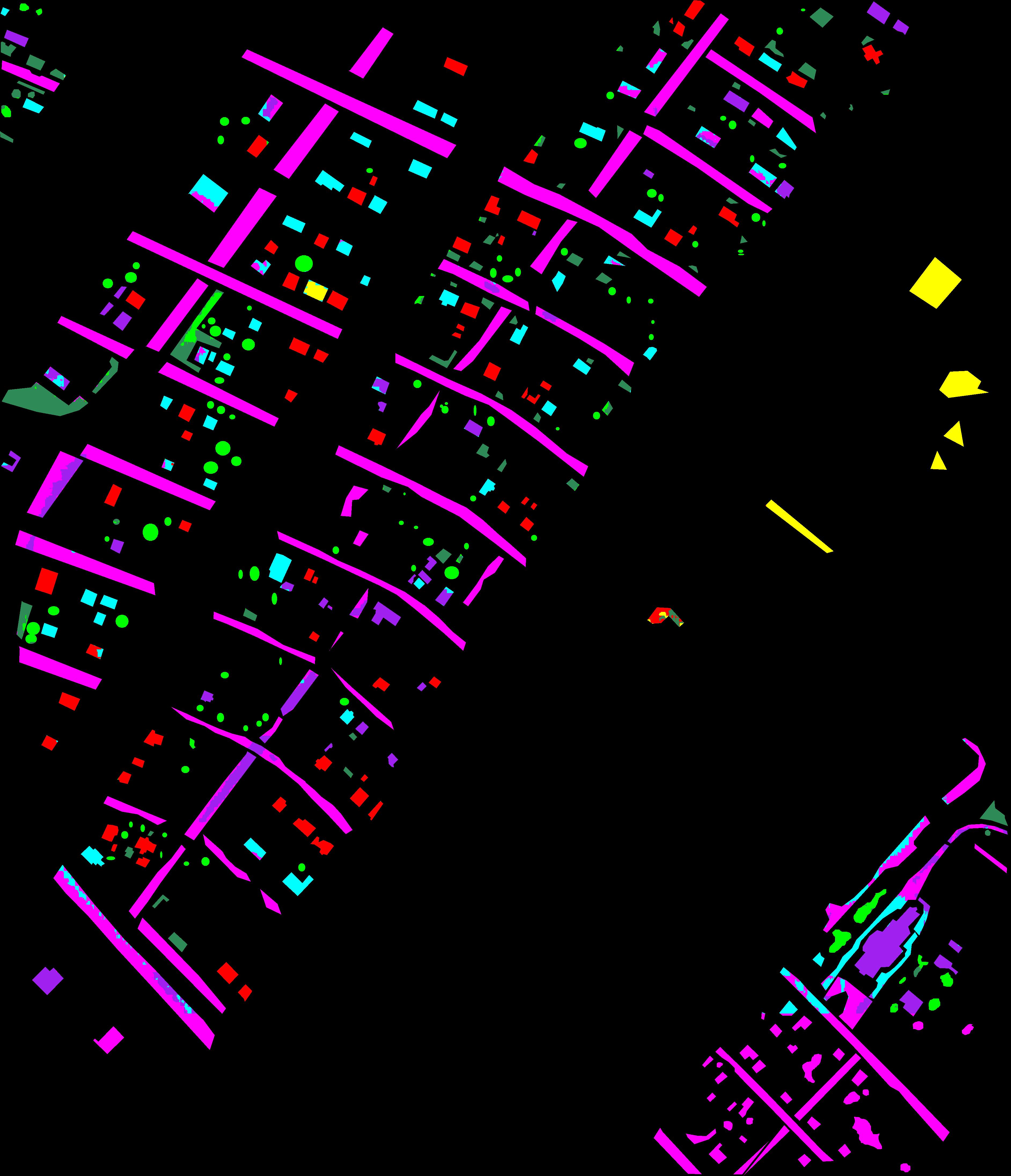}
		}
		\subfloat[Dilated8Pooling]{
			\includegraphics[height=0.26\textwidth]{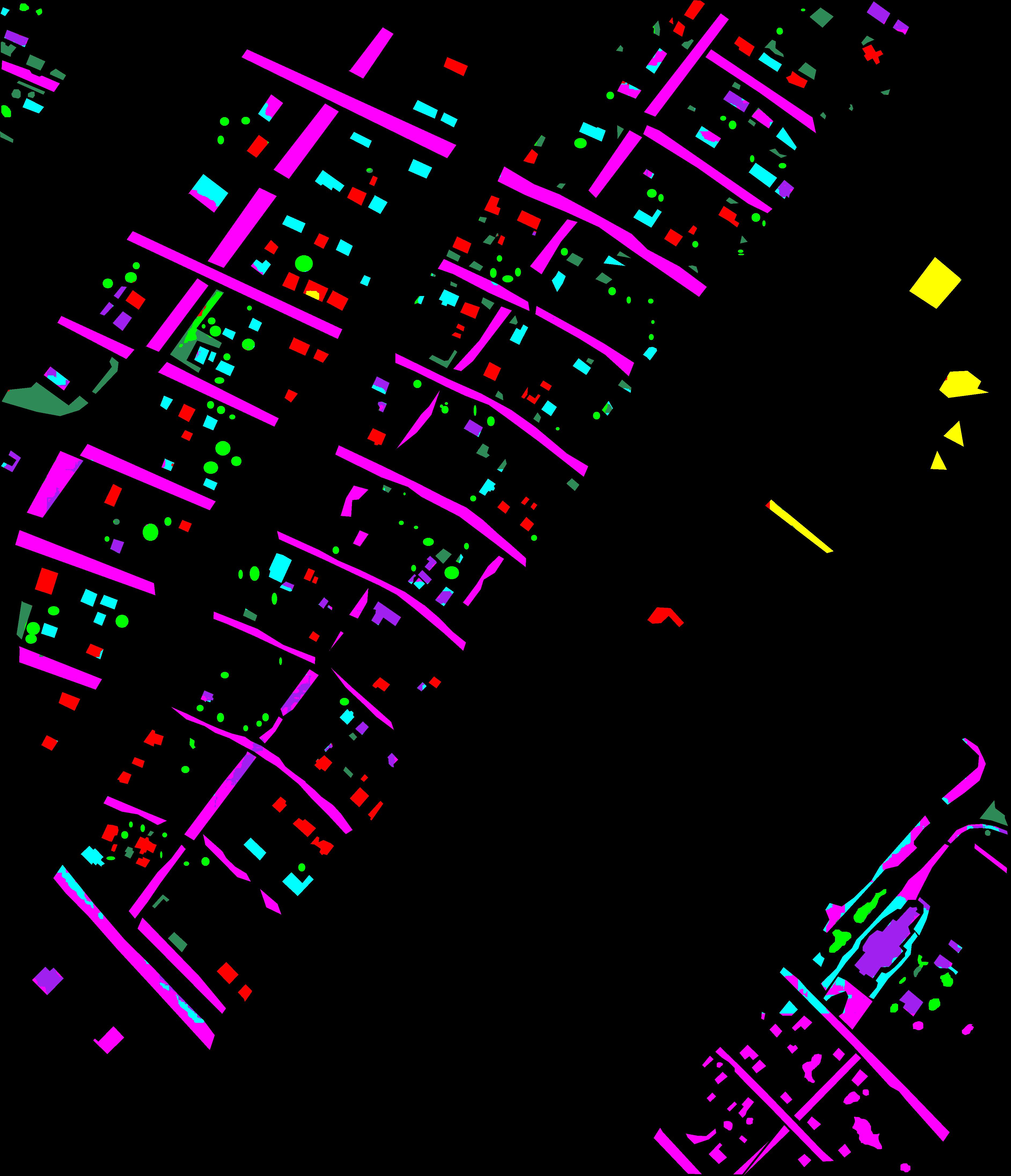}
		}
		\caption{The GRSS Data Fusion training and test images, their respective ground-truths and the prediction maps generated by the proposed algorithm. Legend -- Black: unclassified. Light purple: road. Light green: trees. Red: red roof. Cyan: gray roof. Dark purple: concrete roof. Dark green: vegetation. Yellow: bare soil.
		}
		\label{fig:grss_results}
	\end{figure*}

	We also performed several experiments on the GRSS Data Fusion Contest dataset~\cite{liao2015processing} considering all analysis carried out in previous sections.
	Experimental results, as well as baselines, are presented in Table~\ref{tab:grss_results}.
	The prediction maps obtained for the test set are presented in Figure~\ref{fig:grss_results}.
	
	Overall, Dilated6 produced the best result among all approaches.
	In general, networks trained with the proposed method outperformed the baselines.
	Moreover, the Dilated6Pooling trained with the proposed dynamic technique outperformed the baseline composed of the same network trained using traditional training process, corroborating with previous conclusions.
	
	Among the baseline methods, although all of them achieved comparable results, the best outcome was yielded by the Deep Contextual~\cite{santanadeep}.
	This method also leverages from multi-context information, since it combines features extracted from distinct layers of pre-trained ConvNets.
	When comparing this method with the best result of the proposed technique (Dilated6), one can clearly observe the advantage of the proposed approach, which improves the results for all metrics when compared to the Deep Contextual~\cite{santanadeep} approach.
	This reaffirms the effectiveness of the proposed dynamic method, corroborating with previous conclusions.

	\begin{table}[h!]
		\centering
		\caption{Results for the GRSS Data Fusion dataset.}
		\label{tab:grss_results}
		\resizebox{\columnwidth}{!}{
			\begin{tabular}{@{}clrrr@{}}
				\toprule
				\multicolumn{1}{c}{\textbf{\begin{tabular}[c]{@{}c@{}}Training\\Process\end{tabular}}} & \multicolumn{1}{c}{\textbf{Network}} &
				\multicolumn{1}{c}{\textbf{\begin{tabular}[c]{@{}c@{}}Overall\\Accuracy\end{tabular}}} & \multicolumn{1}{c}{\textbf{\begin{tabular}[c]{@{}c@{}}Average\\Accuracy\end{tabular}}} & \multicolumn{1}{c}{\textbf{\begin{tabular}[c]{@{}c@{}}Kappa\\Index\end{tabular}}} \\
				\midrule
				\multirow{5}{*}{\textbf{Traditional}} & Pixelwise~\cite{keiller2016icpr}  							& 85.04	&	86.52	&	78.18 \\
				& FCN~\cite{long2015fully}  							  		& 83.27	&	87.45	&	76.10 \\
				& Deconvolution Network~\cite{badrinarayanan2015segnet}		& 82.15	&	86.24	&	75.04 \\
				& Dilated network (Dilated6Pooling)							& 83.96	&	83.83	&	76.12 \\
				& Deep Contextual~\cite{santanadeep}  						& 85.45 &   88.33 	&   79.01 \\
				\midrule
				\multirow{4}{*}{\textbf{Dynamic}}  & Dilated6	  												& 90.10	&	90.13	&	85.22 \\
				& DenseDilated6			  									& 88.66	&	80.62	&	81.80 \\
				& Dilated6Pooling				    						& 88.05	&	86.12	&	81.81 \\
				& Dilated8Pooling			   								& 89.03	&	85.31	&	83.08 \\
				\bottomrule
			\end{tabular}
		}
	\end{table}

	\subsubsection{Vaihingen Dataset} \label{subsec:vaihingen_state_results}
	
	As introduced in Section~\ref{subsec:baselines}, official results for the Vaihingen dataset are reported only by the challenge organization that held some images that are used for testing the submitted algorithms.
	Therefore, one must submit the outcomes of the proposed algorithm to have them evaluated.
	In our case, following previous analysis, we submitted five approaches: 
	the first four are related to each network presented in Section~\ref{sec:methodology} trained with the 6 classes (which are represented in the official results as UFMG\_1 to 4),
	and the fifth one, represented in the official results as UFMG\_5, is the Dilated8 network (Figure~\ref{dilated4}) trained with only 5 classes, i.e., all labels except the clutter/background one.
	This last submission is due to the lack of training data for that class which corresponds to only 0.67\% of the dataset (as stated in Table~\ref{tab:isprs_stat}).
	It is important to note that all submissions related to the proposed work do not use any post-processing, such as Conditional Random Fields (CRF)~\cite{lafferty2001conditional}.

	Some official results reported by the organization are summarized in Table~\ref{tab:vaihingen_results}.
	In addition to our results, this table also compiles the \textbf{best results of each work} with enough information 
	to make a fair comparison, i.e., in which the proposed approach is minimally explained.
	In order to allow a visual comparison, examples of the proposed method, for the validation and test sets, are presented in Figures~\ref{fig:vaihingein_validation_results} and~\ref{fig:vaihingein_test_results}, respectively.
	
	It is possible to notice that the proposed work yielded competitive results.
	The best result, in terms of overall accuracy, was $90.3\%$ achieved by DLR\_9~\cite{marmanis2016classification} and GSN3~\cite{wang2017gated}.
	Our best result (UFMG\_4) appears in fifth place by yielding $89.4\%$ of overall accuracy, outperforming several methods, such as ADL\_3~\cite{paisitkriangkrai2015effective} and RIT\_L8~\cite{liu2017dense}, that also tried to aggregate multi-context information.
	However, as can be seen in Table~\ref{tab:vaihingen_results} and Figure~\ref{vaihingen}, while the other approaches have a larger number of trainable parameters, our network has only 2 millions, which makes it less pruned to overfitting and, consequently, easier to train, showing that the proposed method really helps in extracting all feasible information of the data even if using limited architectures (in terms of parameters).
	In fact, the number of parameters of the network is so relevant that authors of DLR\_9 submission~\cite{marmanis2016classification}, one of the best results but with a higher number of parameters, do not recommend their proposed method for practical use because of the memory consumption and expensive training phase.
	Furthermore, the obtained results, that do not have any post-processing, are better than others, such as DST\_2~\cite{sherrah2016fully}, that employ CRF as post-processing method, which shows the potential of dilated convolutions in aggregate refined information.

	Aside from this, the proposed work (UFMG\_5) achieved the best result ($82.5\%$ of F1 Score) in the car class, which is one of the most difficult classes (of this dataset) when compared to others (such as building) because of its composition (small objects) and its high intraclass variance (caused by a great variety of models and colors).
	This may be justified by the fact that the proposed network does not downsample the input image preserving important details for such classes composed of small objects.
	However, this submission ignores the clutter/background class, which could be considered as an advantage, making the comparison unfair.
	But, there are other works doing the same training protocol (i.e., ignoring the clutter/background class), such as INR~\cite{maggiori2016high}.
	Yet such works have not achieved good accuracy in the car class as the proposed work.
	Furthermore, still considering the car class, the second best result ($81.3\%$ of F1 Score) is also yielded by our proposed work (UFMG\_4), which employs all classes during the training phase, which shows the effectiveness and robustness of our work mainly for classes related to small objects.
	
	\begin{table}[]
		\centering
		\caption{Official results for the Vaihingen dataset.}
		\label{tab:vaihingen_results}
		\resizebox{\columnwidth}{!}{ 
			\begin{tabular}{@{}lrrrrrrr@{}}
				\toprule
				\multicolumn{1}{c}{\multirow{3}{*}{\textbf{Method}}} & \multicolumn{1}{c}{\multirow{3}{*}{\textbf{\#Parameters}}} & \multicolumn{5}{c}{\textbf{F1 Score}} & \multicolumn{1}{c}{\textbf{\multirow{3}{*}{\begin{tabular}[c]{@{}c@{}}Overall\\Accuracy\end{tabular}}}} \\
				\cmidrule(lr){3-7}
				\multicolumn{1}{c}{} & \multicolumn{1}{c}{} & \multicolumn{1}{c}{\textbf{\begin{tabular}[c]{@{}c@{}}Impervious\\Surface\end{tabular}}} & \multicolumn{1}{c}{\textbf{Building}} & \multicolumn{1}{c}{\textbf{\begin{tabular}[c]{@{}c@{}}Low\\Vegetation\end{tabular}}} & \multicolumn{1}{c}{\textbf{Tree}} & \multicolumn{1}{c}{\textbf{Car}} & \multicolumn{1}{c}{} \\ 
				\midrule
				DLR\_9~\cite{marmanis2016classification}					& $806\cdot10^6$ & 92.4 & 95.2 & 83.9 & 89.9 & 81.2 & 90.3 \\
				GSN3~\cite{wang2017gated} 									& $44\cdot10^6$  & 92.3 & 95.2 & 84.1 & 90.0 & 79.3 & 90.3 \\
				ONE\_7~\cite{audebert2016semantic} 							& $28\cdot10^6$  & 91.0 & 94.5 & 84.4 & 89.9 & 77.8 & 89.8 \\
				INR~\cite{maggiori2016high} 								& $4\cdot10^6$   & 91.1 & 94.7 & 83.4 & 89.3 & 71.2 & 89.5 \\
				\textbf{UFMG\_4} 							& $2\cdot10^6$   & 91.1 & 94.5 & 82.9 & 88.8 & 81.3 & 89.4 \\  
				\textbf{UFMG\_5}							& $2\cdot10^6$   & 91.0 & 94.6 & 82.7 & 88.9 & 82.5 & 89.3 \\  
				\textbf{UFMG\_1}		 							& $1.3\cdot10^6$ & 90.5 & 94.1 & 82.5 & 89.0 & 78.5 & 89.1 \\  
				DST\_2~\cite{sherrah2016fully} 								& $3.5\cdot10^6$ & 90.5 & 93.7 & 83.4 & 89.2 & 72.6 & 89.1 \\ 
				\textbf{UFMG\_2} 							& $0.8\cdot10^6$ & 90.7 & 94.3 & 82.5 & 88.5 & 77.4 & 89.0 \\  
				\textbf{UFMG\_3} 							& $1.3\cdot10^6$ & 90.6 & 93.4 & 82.4 & 88.5 & 79.8 & 88.8 \\  
				ADL\_3~\cite{paisitkriangkrai2015effective} 				& $0.5\cdot10^6$ & 89.5 & 93.2 & 82.3 & 88.2 & 63.3 & 88.0 \\ 
				RIT\_2~\cite{piramanayagam2016classification} 				& $138\cdot10^6$ & 90.0 & 92.6 & 81.4 & 88.4 & 61.1 & 88.0 \\
				RIT\_L8~\cite{liu2017dense} 								& $134\cdot10^6$ & 89.6 & 92.2 & 81.6 & 88.6 & 76.0 & 87.8 \\
				UZ\_1~\cite{volpi2017dense} 								& $2.5\cdot10^6$ & 89.2 & 92.5 & 81.6 & 86.9 & 57.3 & 87.3 \\
				\bottomrule
			\end{tabular}
		}
	\end{table}
	
	\newcommand{\exVaihingen}{0.125}
	\begin{table*}[t]
		\begin{center}
			\begin{tabular}{>{\centering\arraybackslash} m{1.95cm} >{\centering\arraybackslash}m{1.95cm} >{\centering\arraybackslash}m{1.95cm} >{\centering\arraybackslash}m{1.95cm} >{\centering\arraybackslash}m{1.95cm} >{\centering\arraybackslash}m{1.95cm} >{\centering\arraybackslash}m{1.95cm} >{\centering\arraybackslash} m{1.95cm} @{}m{0pt}@{} } 
				\textbf{Image} & \textbf{nDSM} & \textbf{Ground-Truth} & \textbf{Dilated6} & \textbf{DenseDilated6} & \textbf{Dilated6 Pooling} & \textbf{Dilated8 Pooling} & \textbf{Dilated8 Pooling} & \\
				\cincludegraphics[width=\exVaihingen\textwidth]{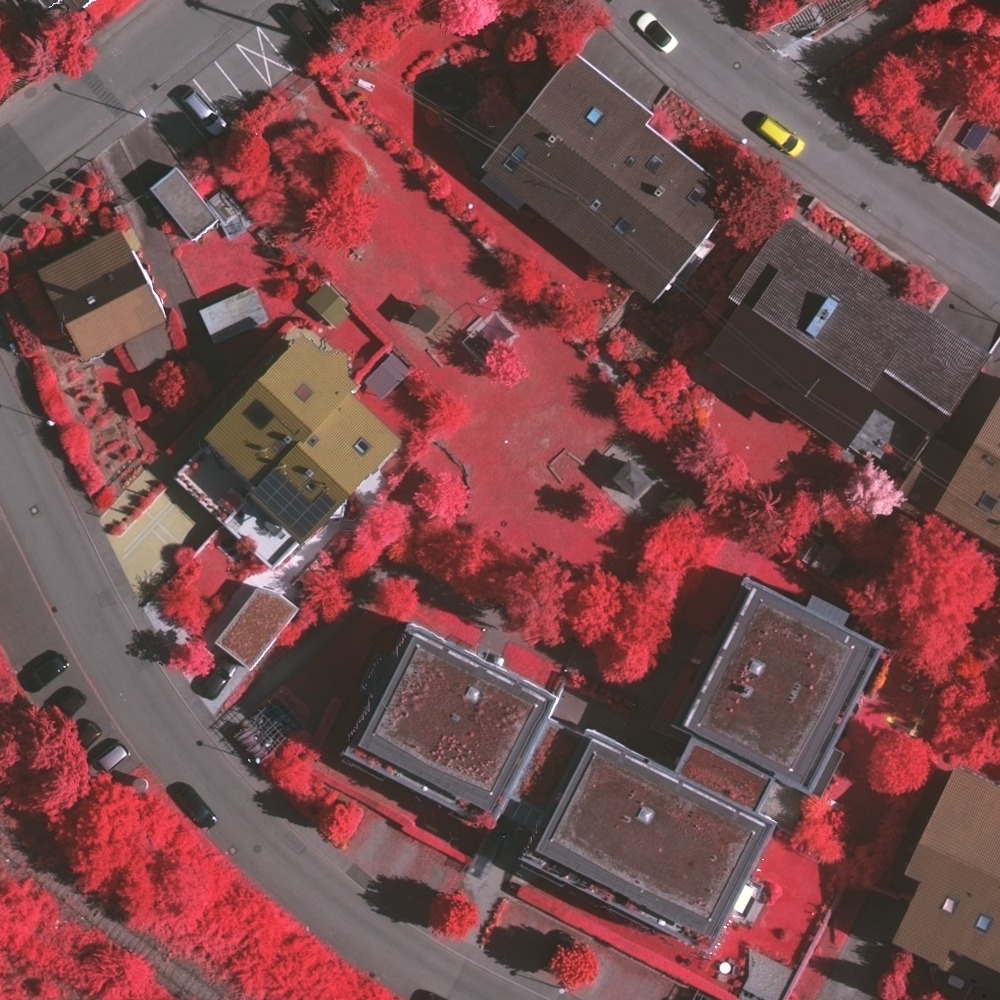} &
				\cincludegraphics[width=\exVaihingen\textwidth]{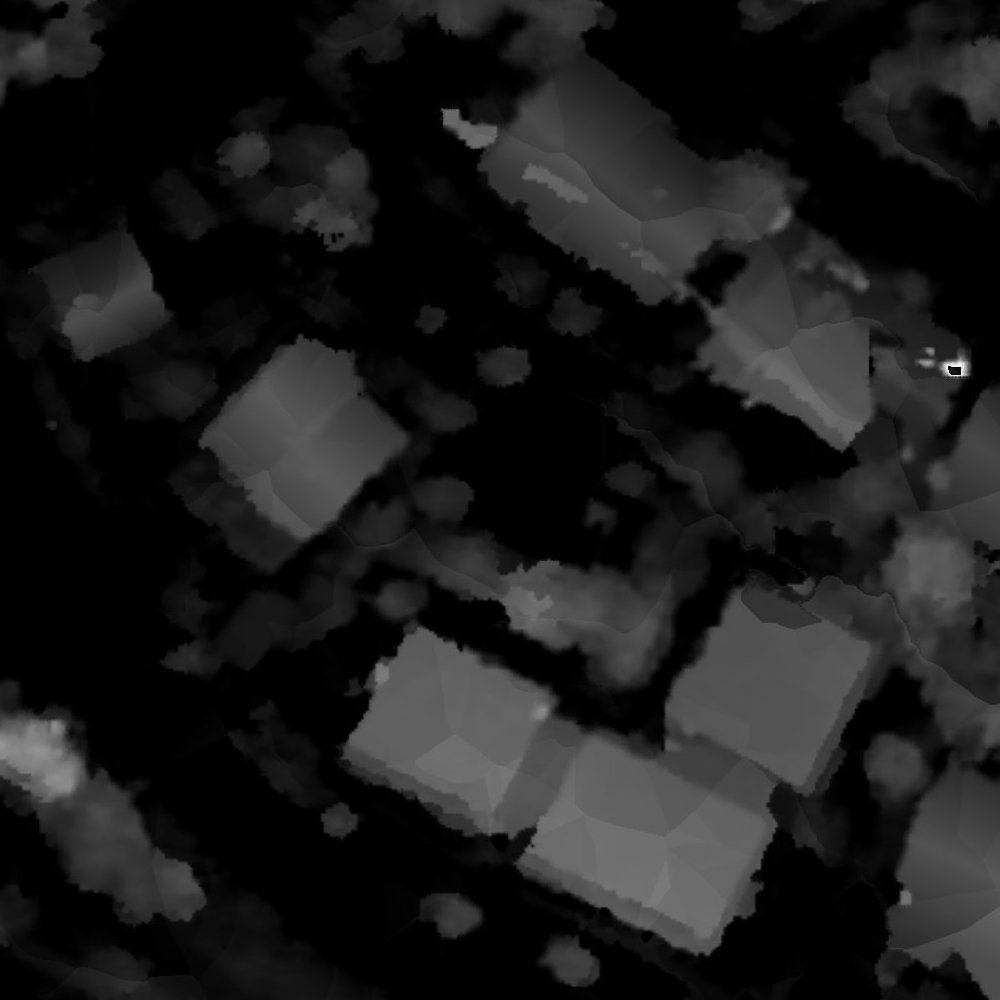} &
				\cincludegraphics[width=\exVaihingen\textwidth]{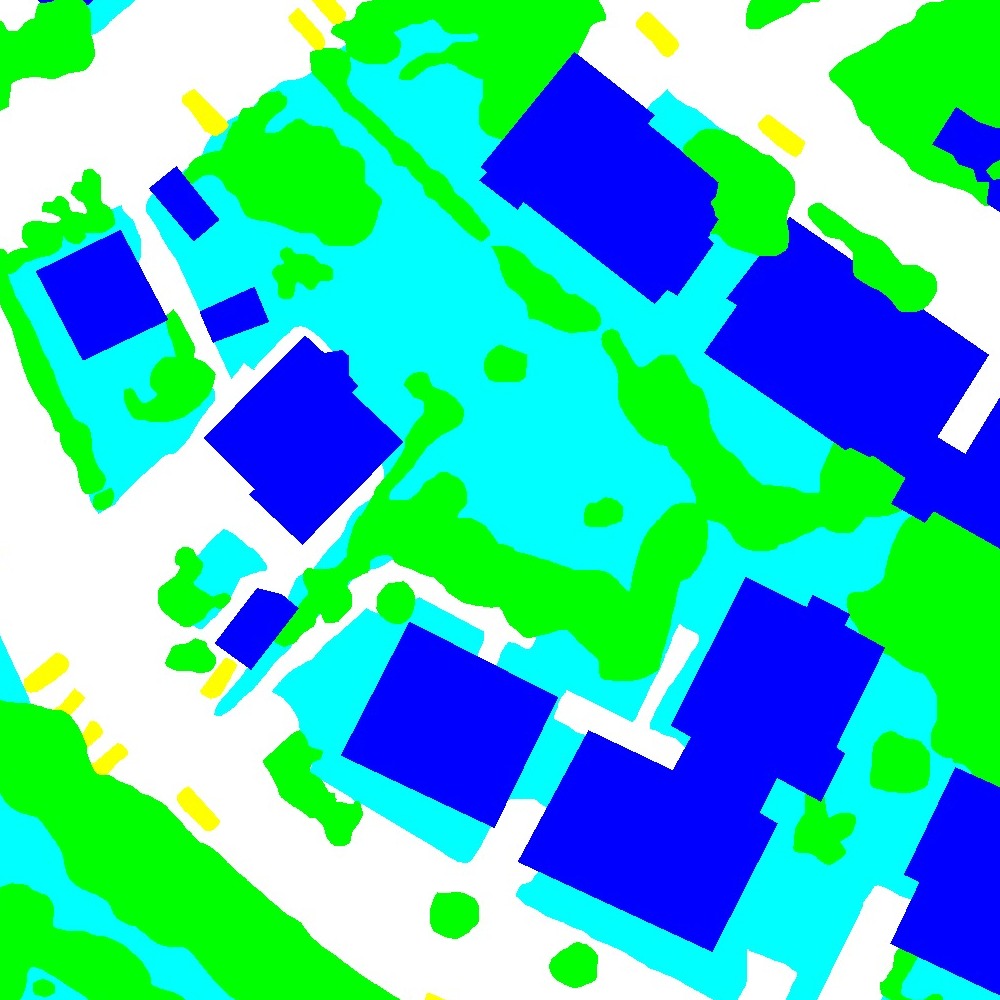} &
				\cincludegraphics[width=\exVaihingen\textwidth]{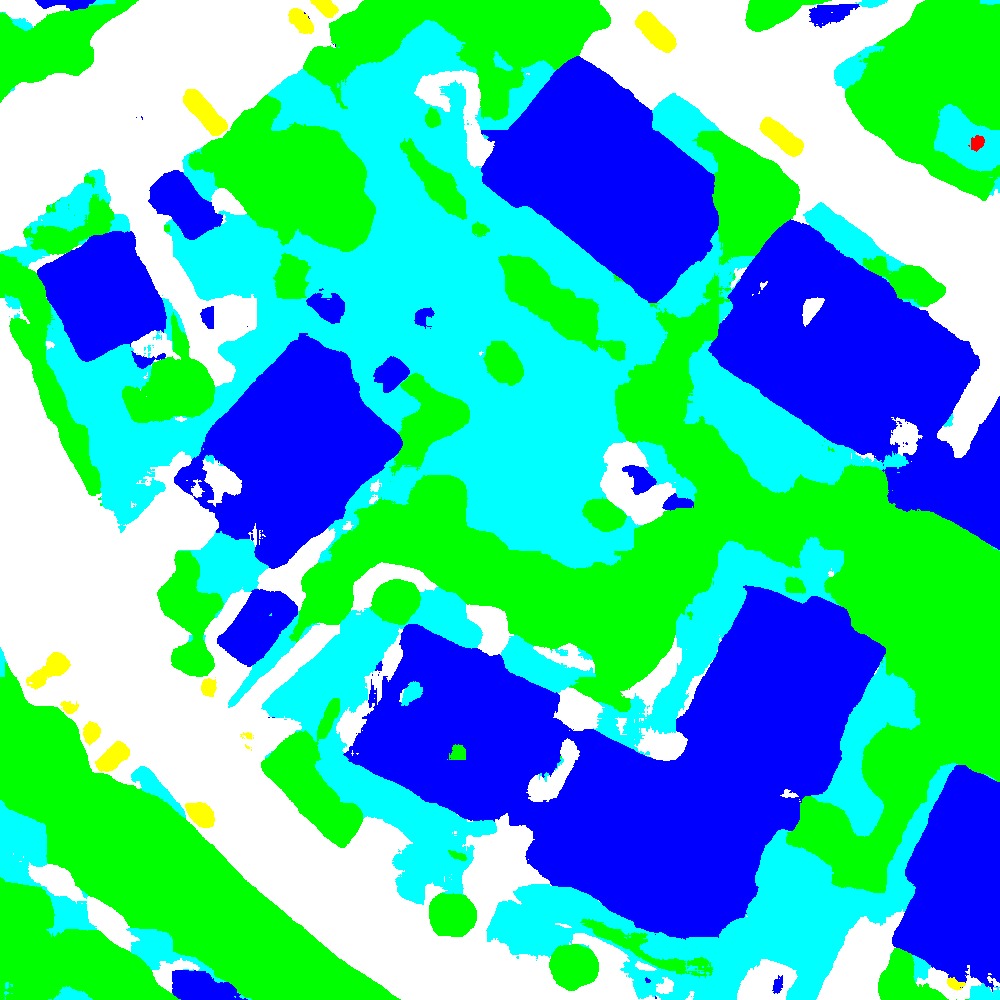} &
				\cincludegraphics[width=\exVaihingen\textwidth]{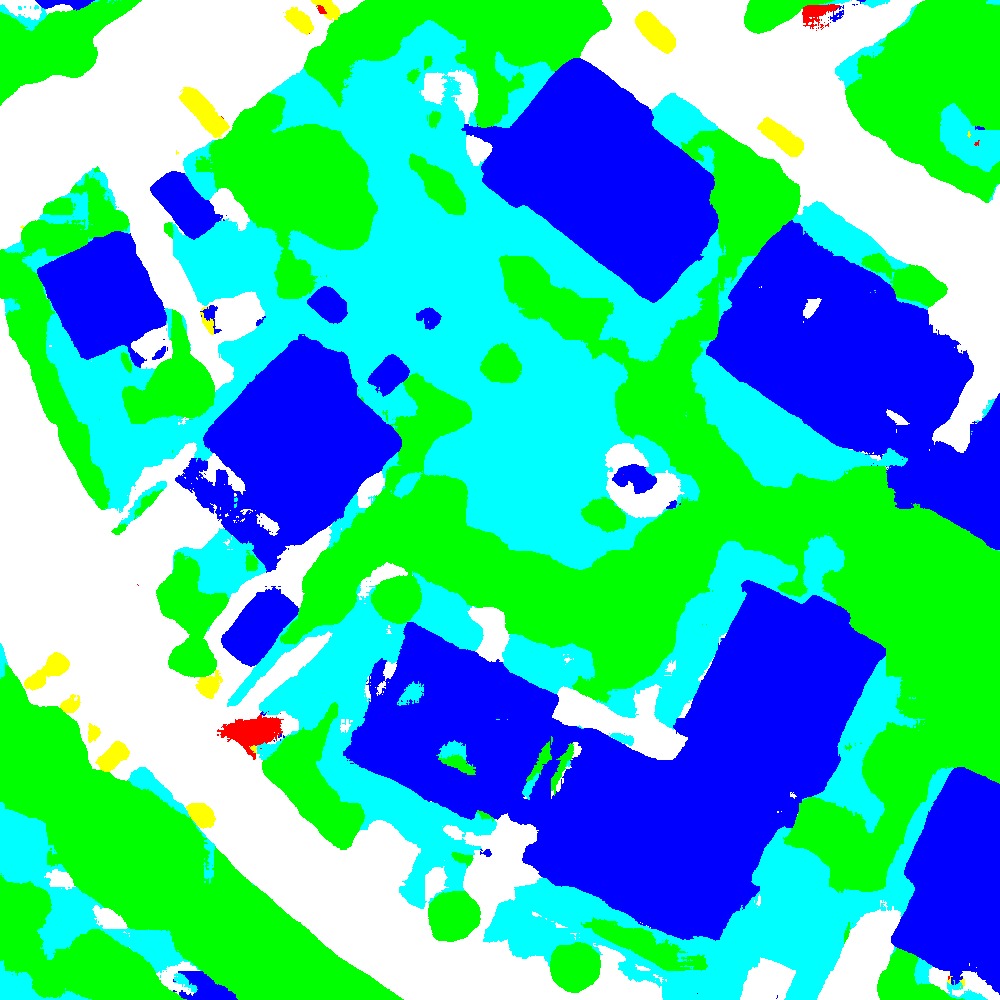} &
				\cincludegraphics[width=\exVaihingen\textwidth]{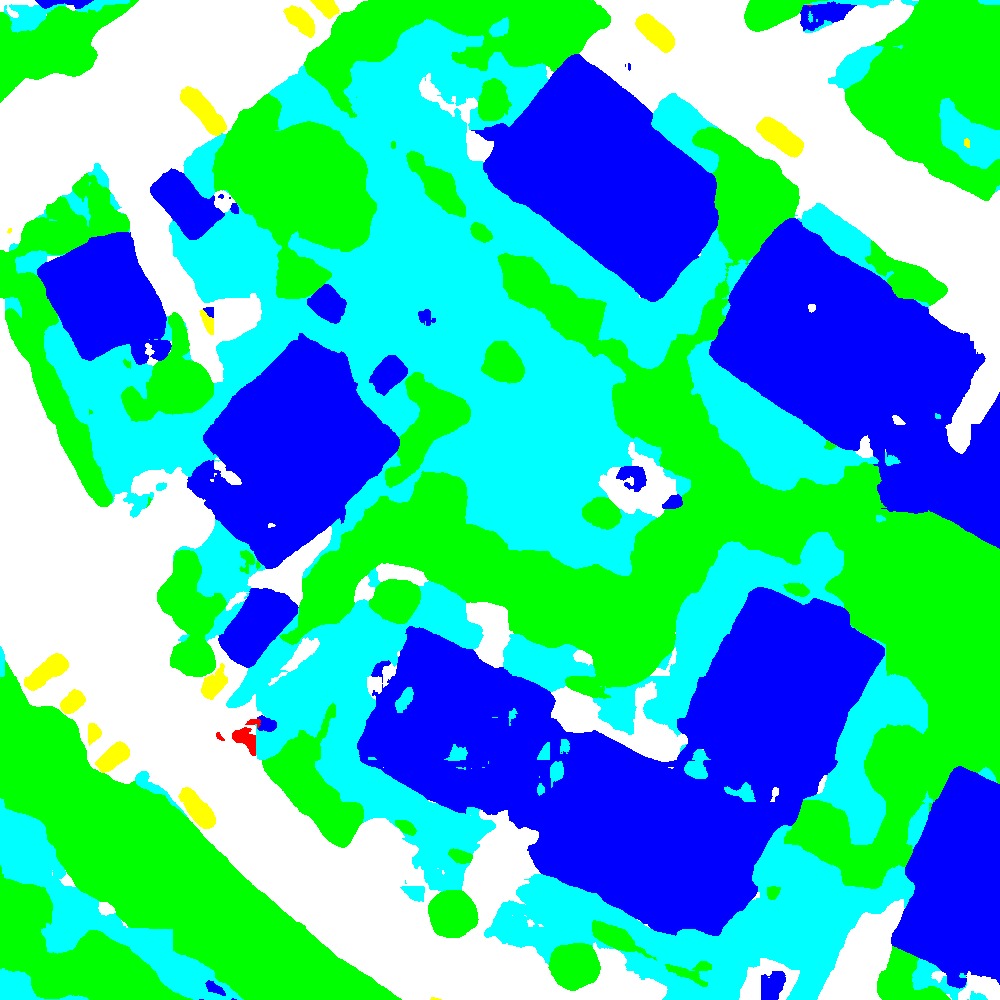} &
				\cincludegraphics[width=\exVaihingen\textwidth]{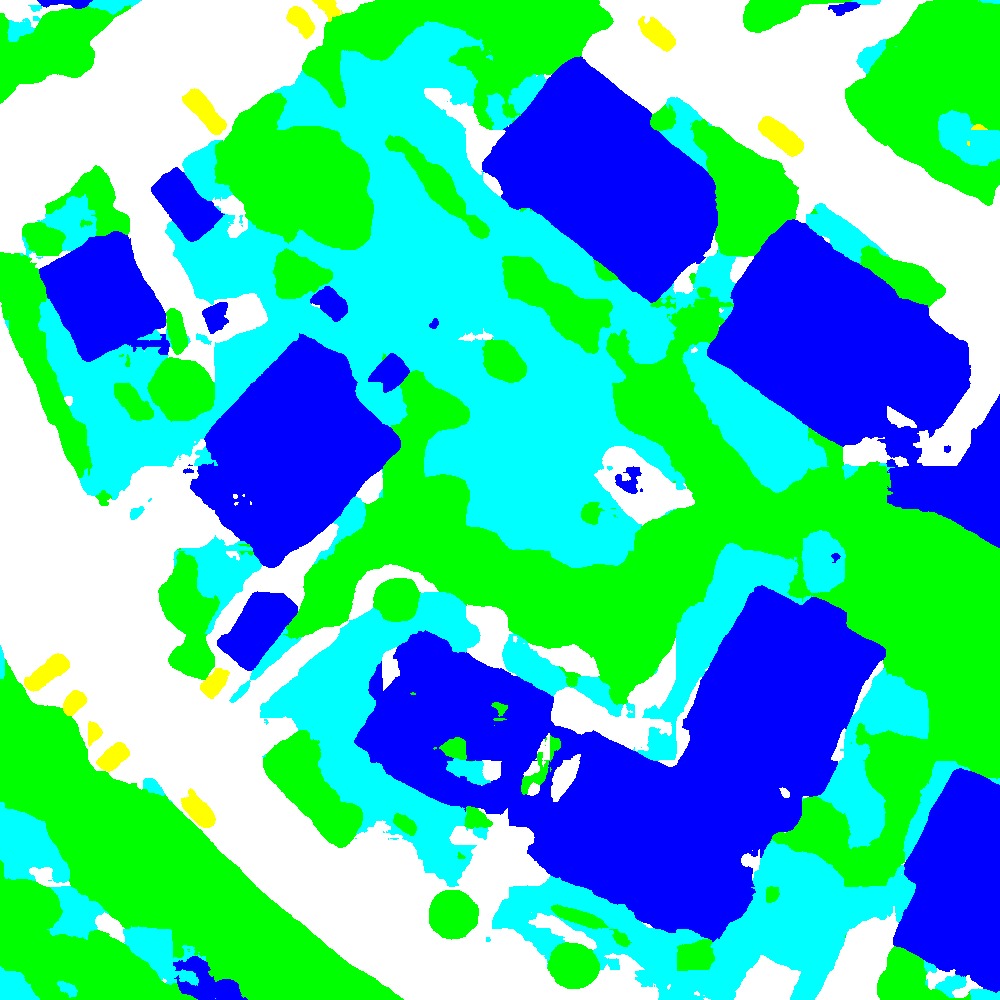} &
				\cincludegraphics[width=\exVaihingen\textwidth]{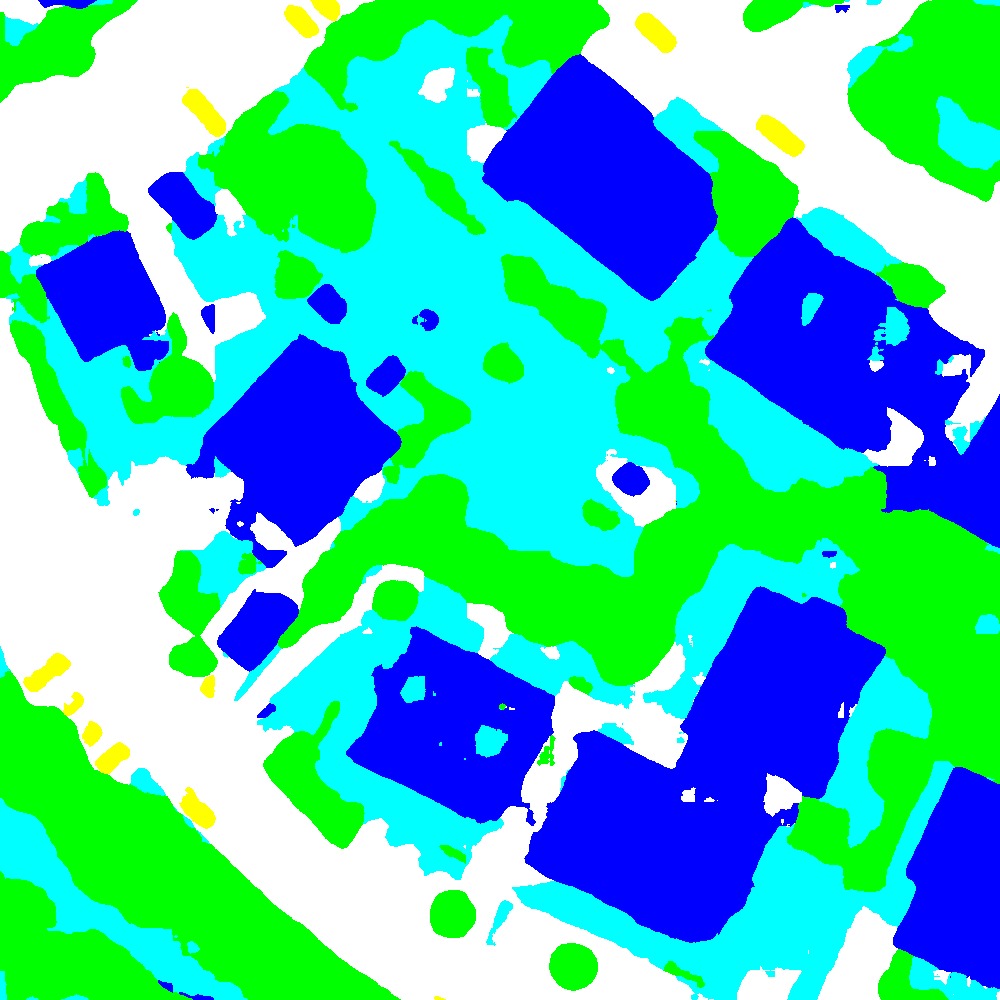} & \\[1.15cm]
				\cincludegraphics[width=\exVaihingen\textwidth]{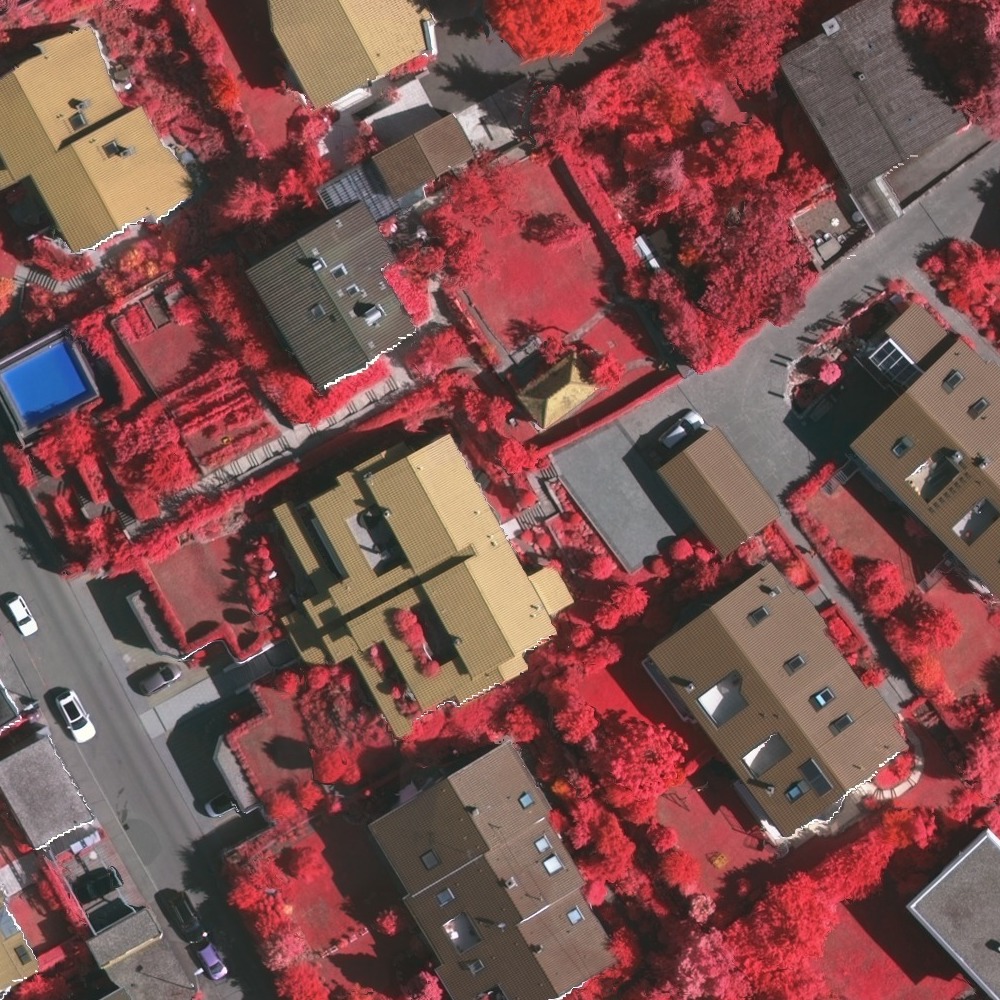} &
				\cincludegraphics[width=\exVaihingen\textwidth]{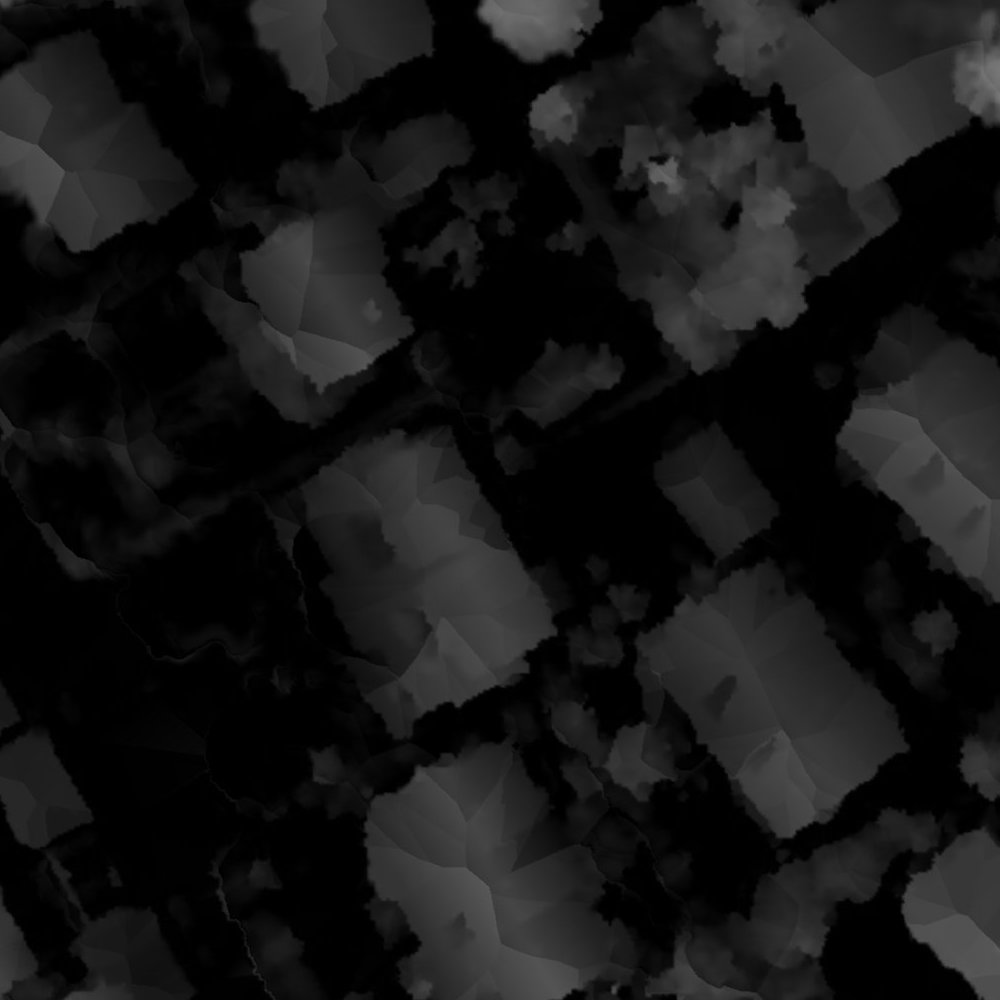} &
				\cincludegraphics[width=\exVaihingen\textwidth]{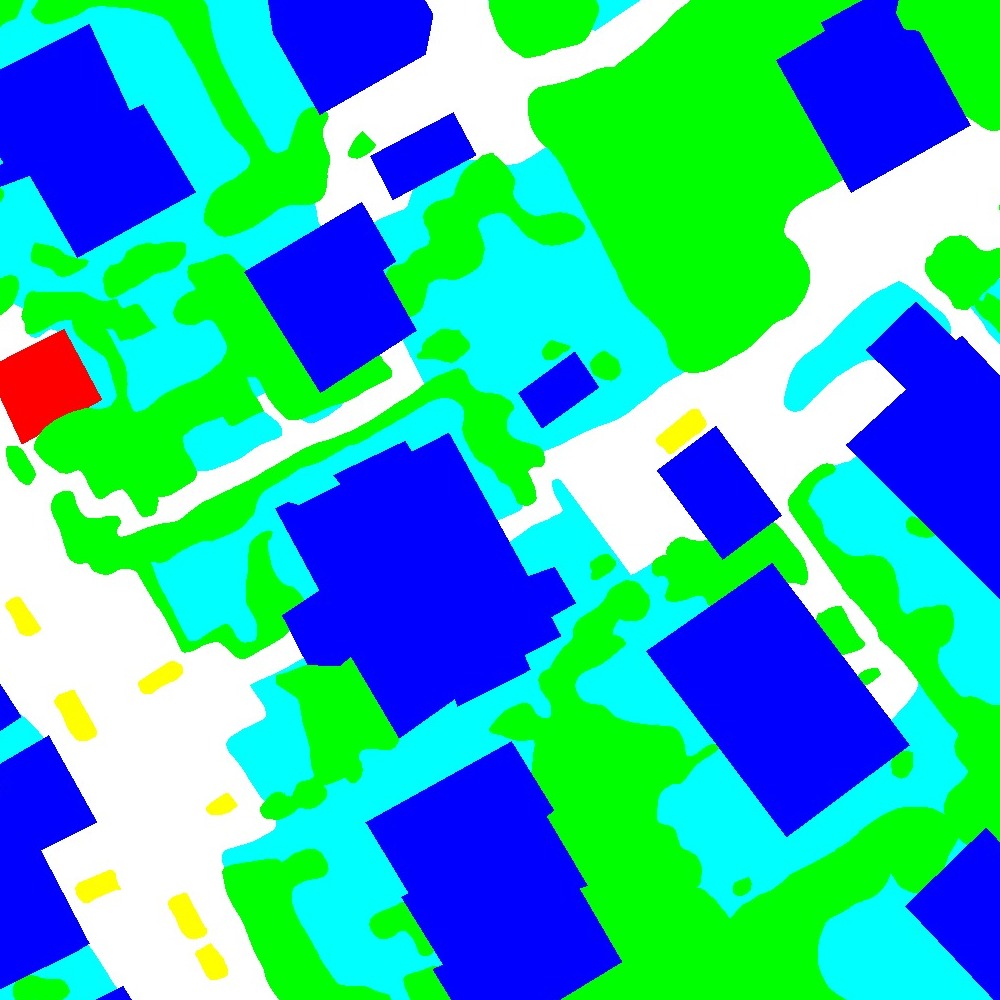} &
				\cincludegraphics[width=\exVaihingen\textwidth]{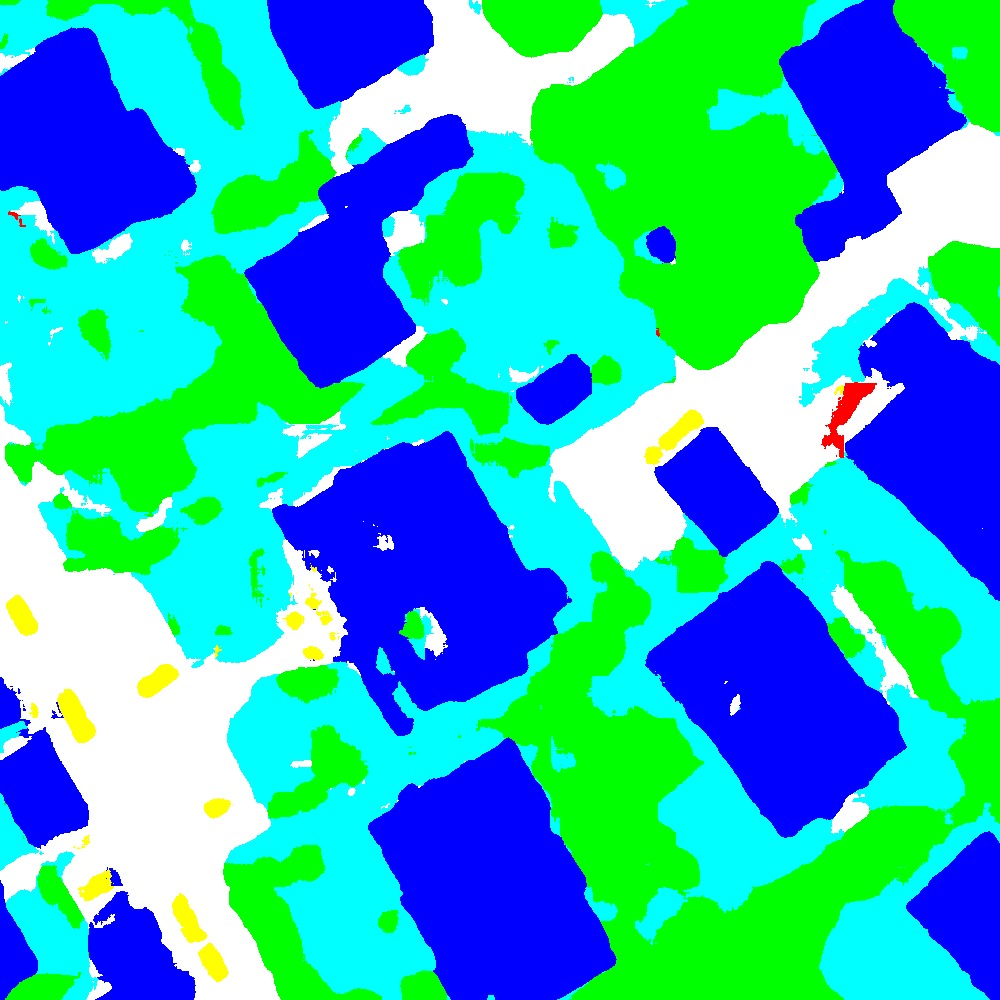} &
				\cincludegraphics[width=\exVaihingen\textwidth]{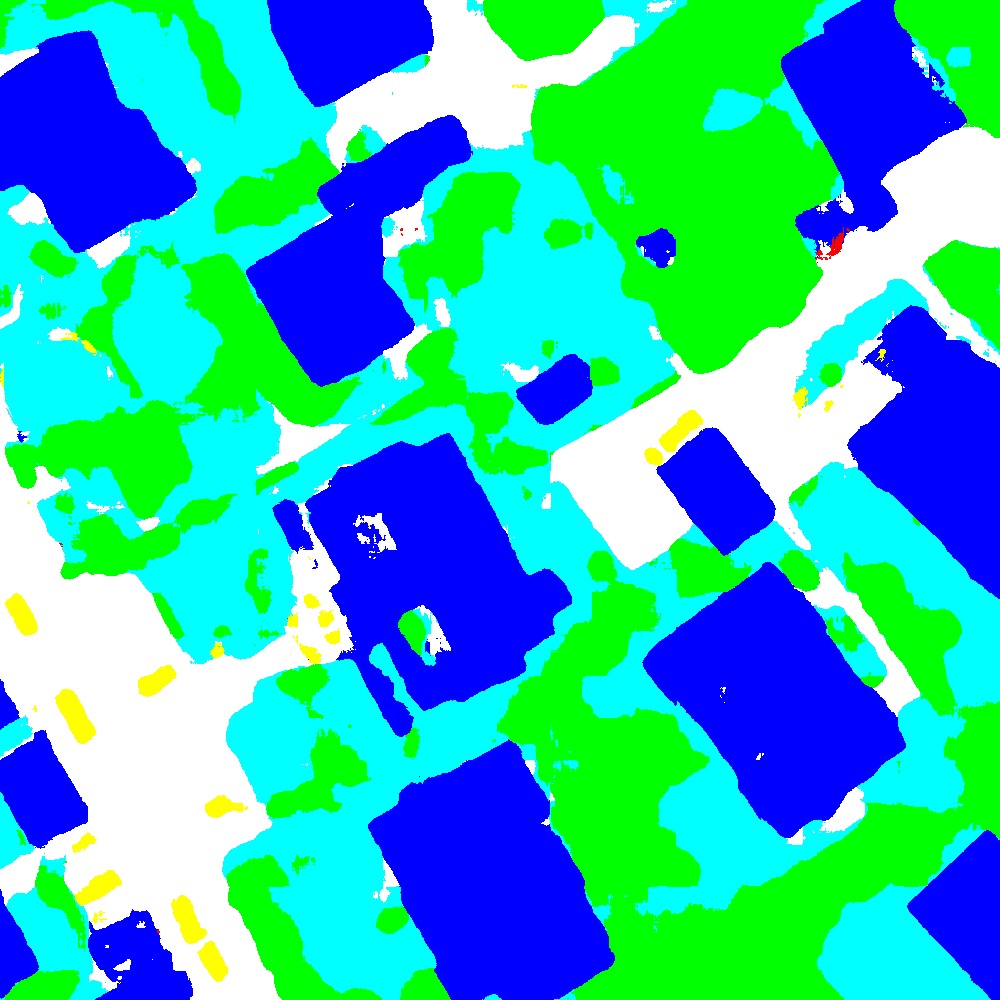} &
				\cincludegraphics[width=\exVaihingen\textwidth]{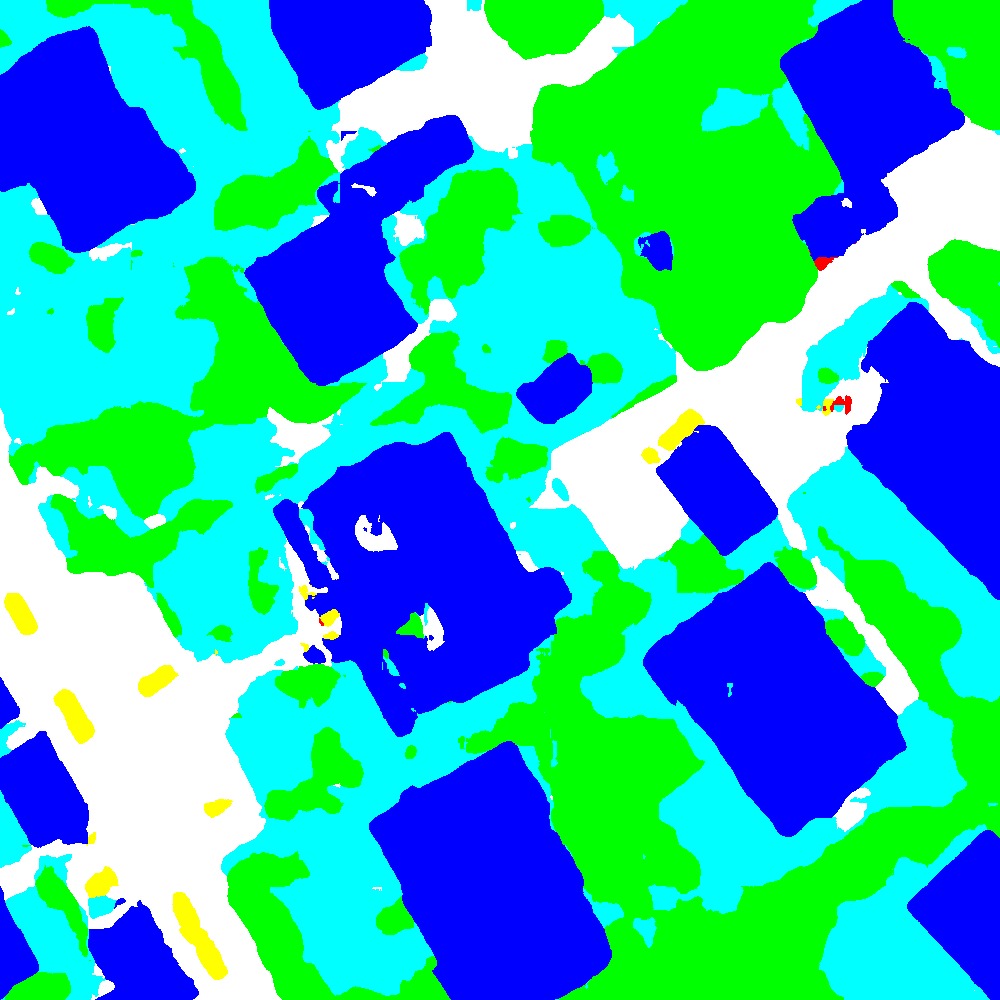} &
				\cincludegraphics[width=\exVaihingen\textwidth]{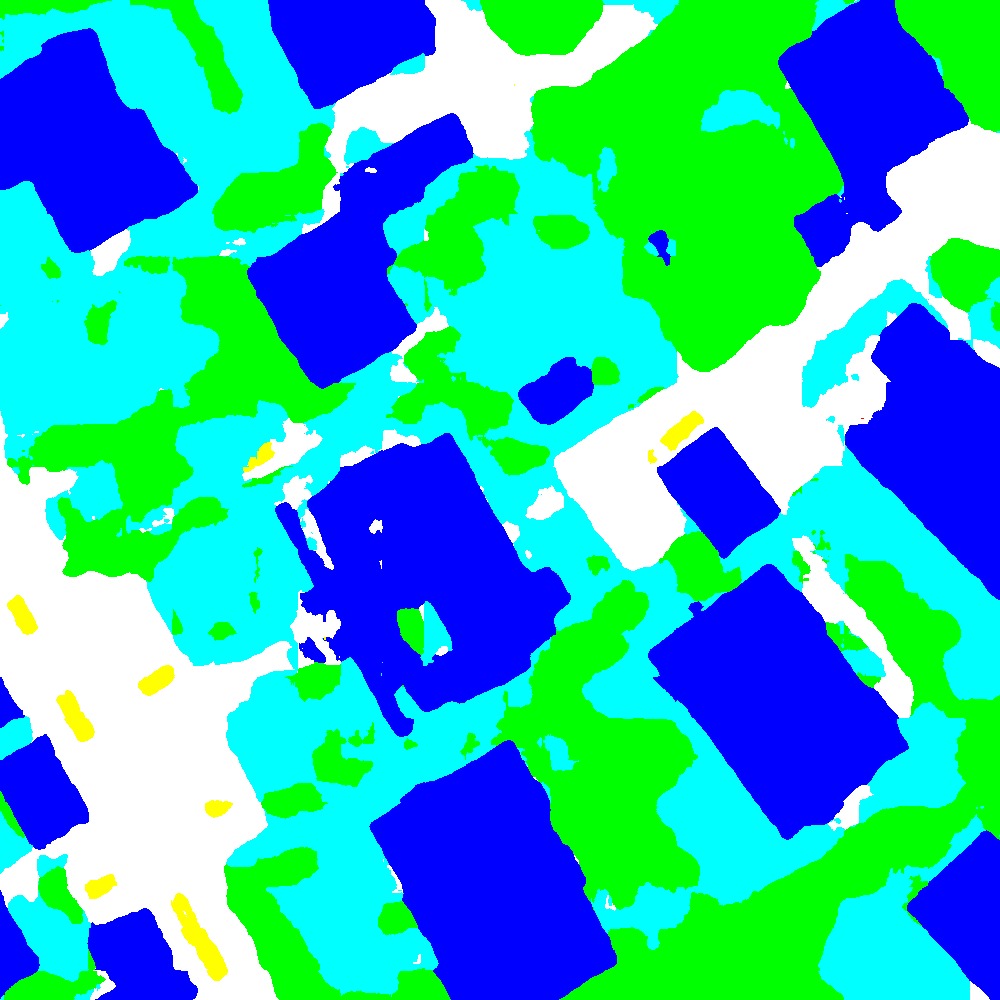} &
				\cincludegraphics[width=\exVaihingen\textwidth]{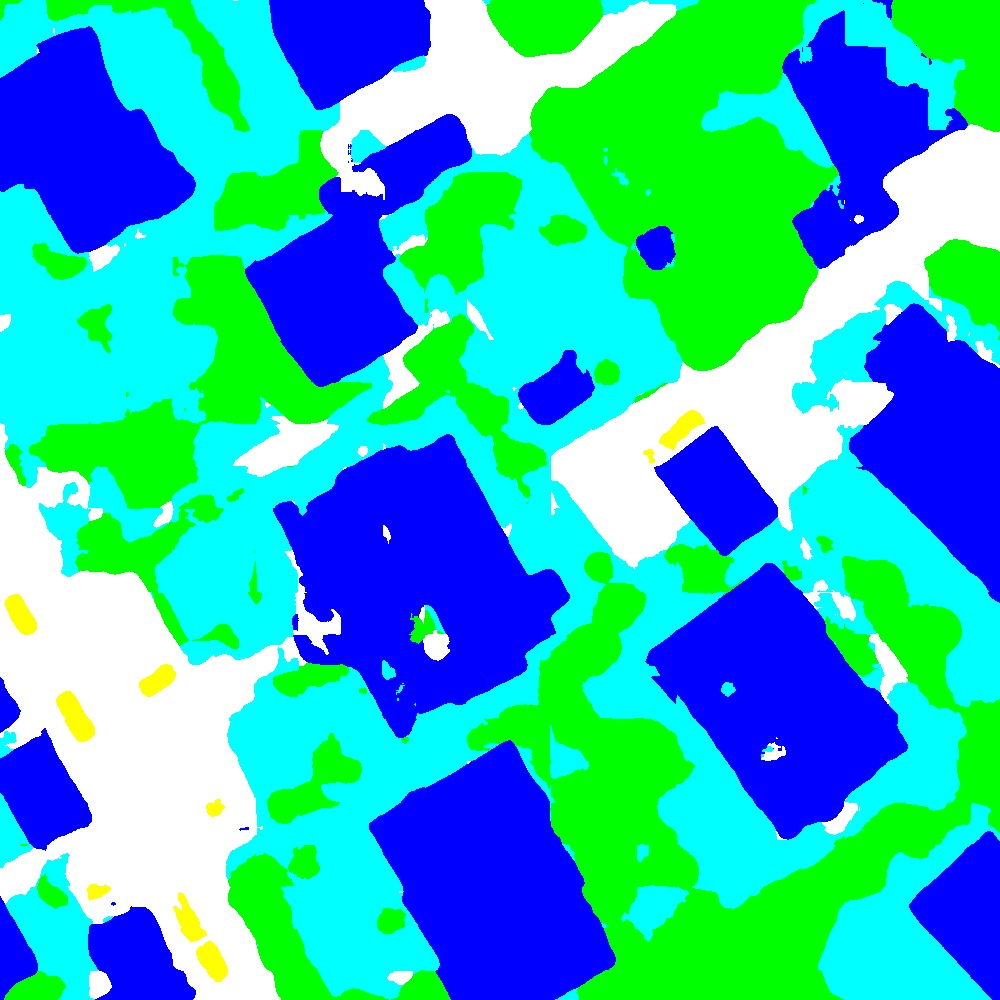} & \\[1.15cm] 
			\end{tabular}
		\end{center}
		\captionof{figure}{Example predictions for the validation set of the Vaihingen dataset.
			Legend -- White: impervious surfaces. Blue: buildings. Cyan: low vegetation. Green: trees. Yellow: cars. Red: clutter, background.
		}
		\label{fig:vaihingein_validation_results}
	\end{table*}
	
	\begin{table*}[t]
		\begin{center}
			\begin{tabular}{>{\centering\arraybackslash} m{1.95cm} >{\centering\arraybackslash}m{1.95cm} >{\centering\arraybackslash}m{1.95cm} >{\centering\arraybackslash}m{1.95cm} >{\centering\arraybackslash}m{1.95cm} >{\centering\arraybackslash}m{1.95cm} >{\centering\arraybackslash}m{1.95cm} @{}m{0pt}@{} } 
				\textbf{Image} & \textbf{nDSM} & \textbf{Dilated6} & \textbf{DenseDilated6} & \textbf{Dilated6 Pooling} & \textbf{Dilated8 Pooling} & \textbf{Dilated8 Pooling} & \\
				\cincludegraphics[width=\exVaihingen\textwidth]{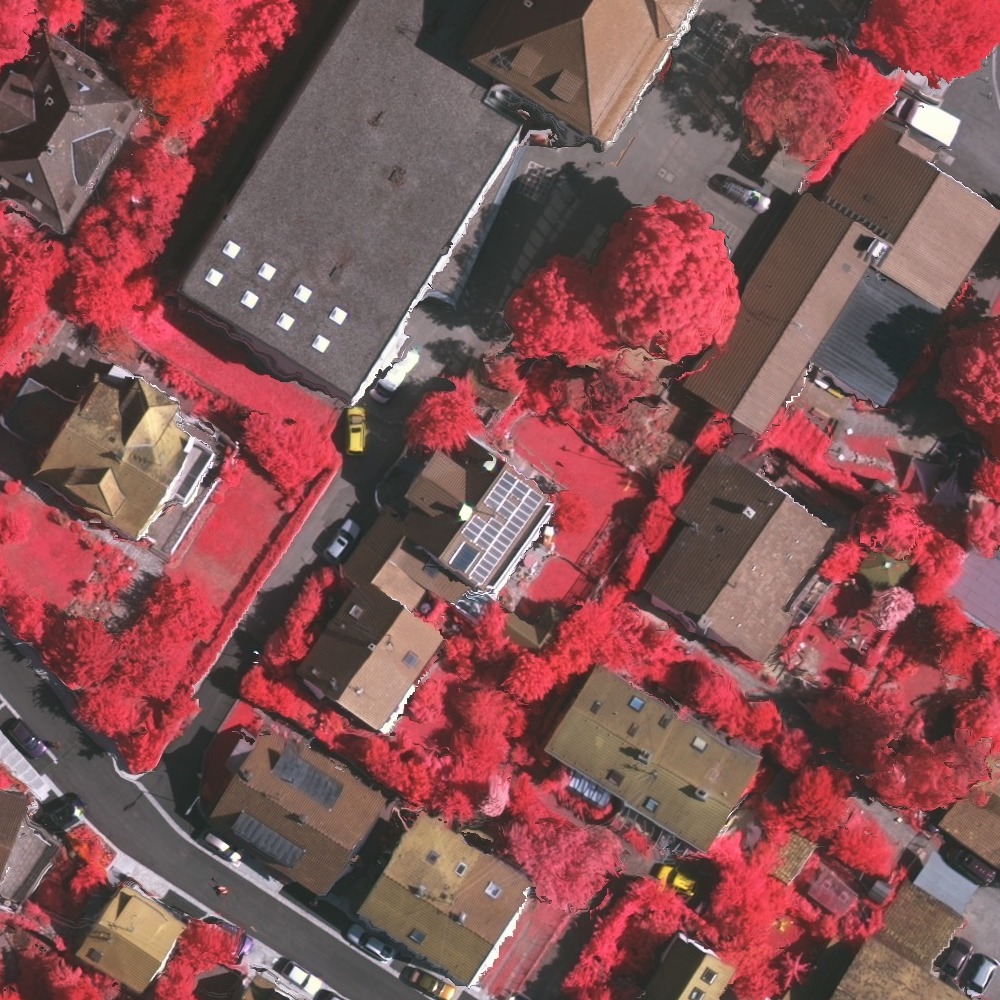} &
				\cincludegraphics[width=\exVaihingen\textwidth]{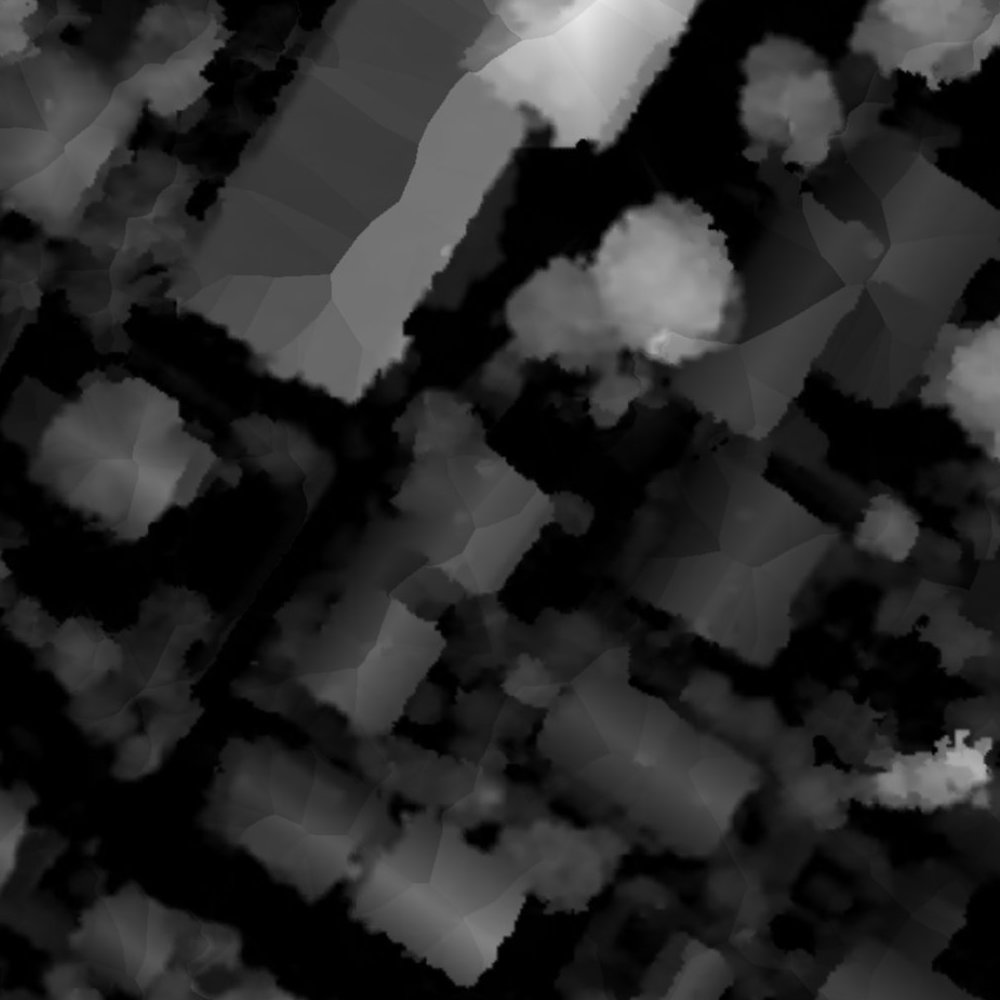} &
				\cincludegraphics[width=\exVaihingen\textwidth]{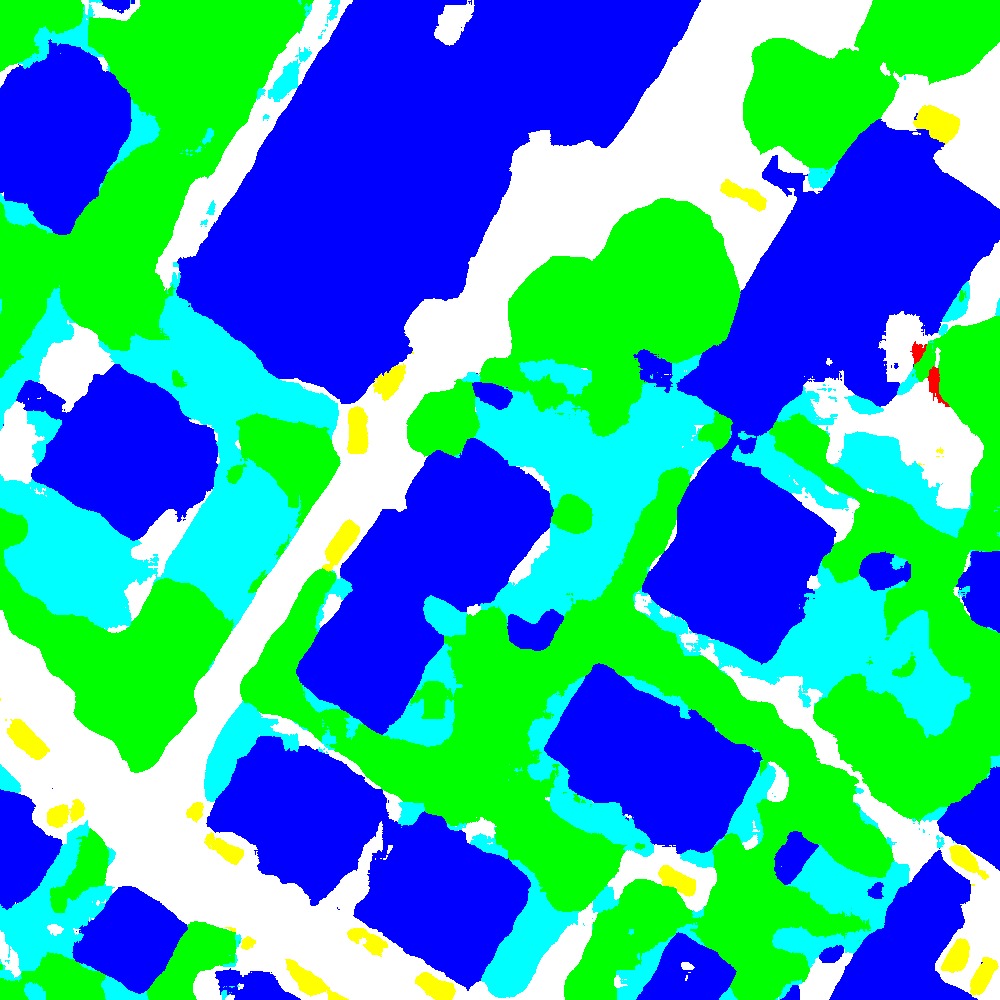} &
				\cincludegraphics[width=\exVaihingen\textwidth]{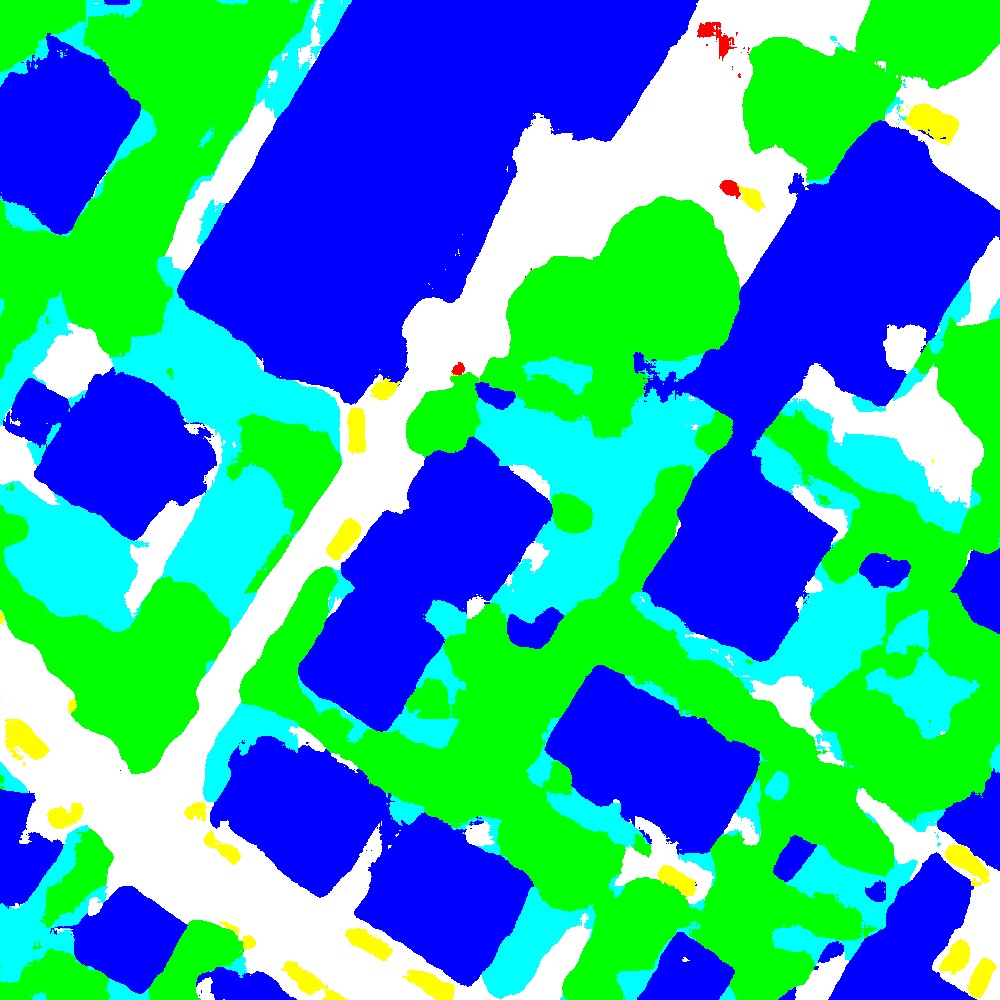} &
				\cincludegraphics[width=\exVaihingen\textwidth]{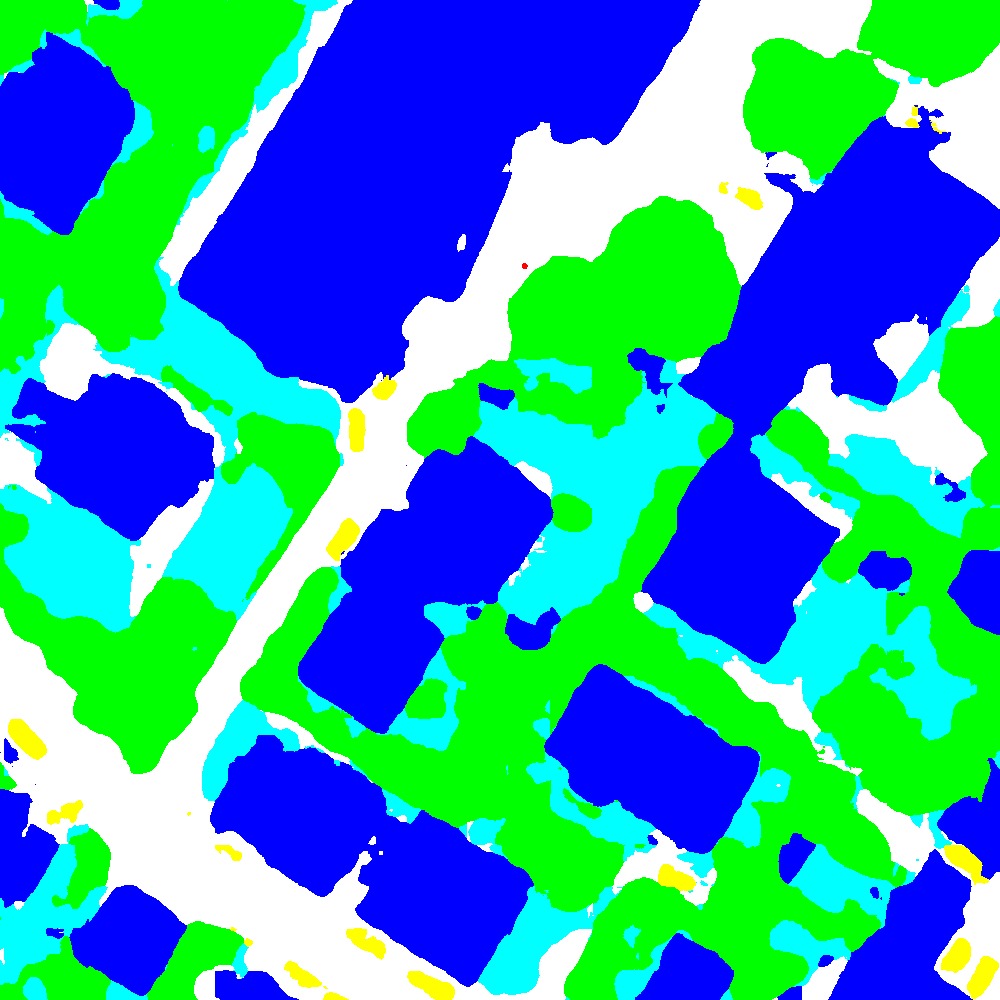} &
				\cincludegraphics[width=\exVaihingen\textwidth]{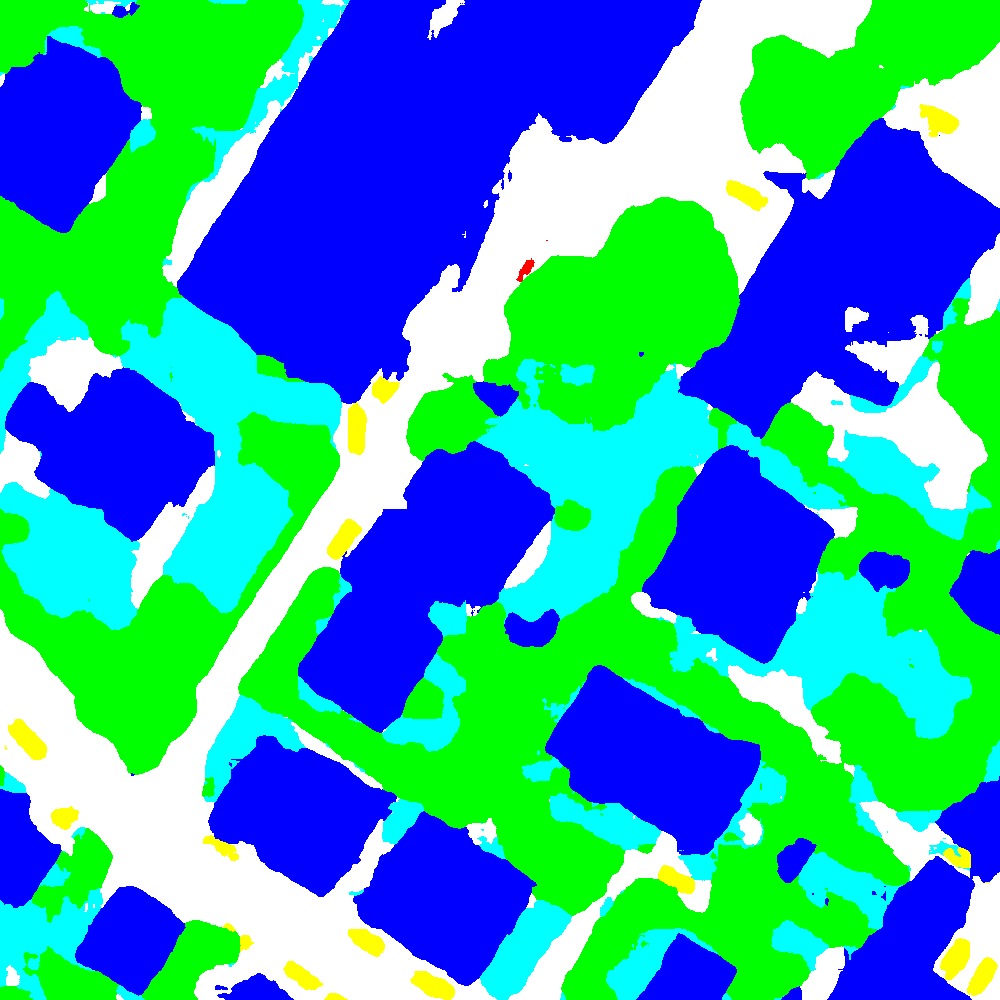} &
				\cincludegraphics[width=\exVaihingen\textwidth]{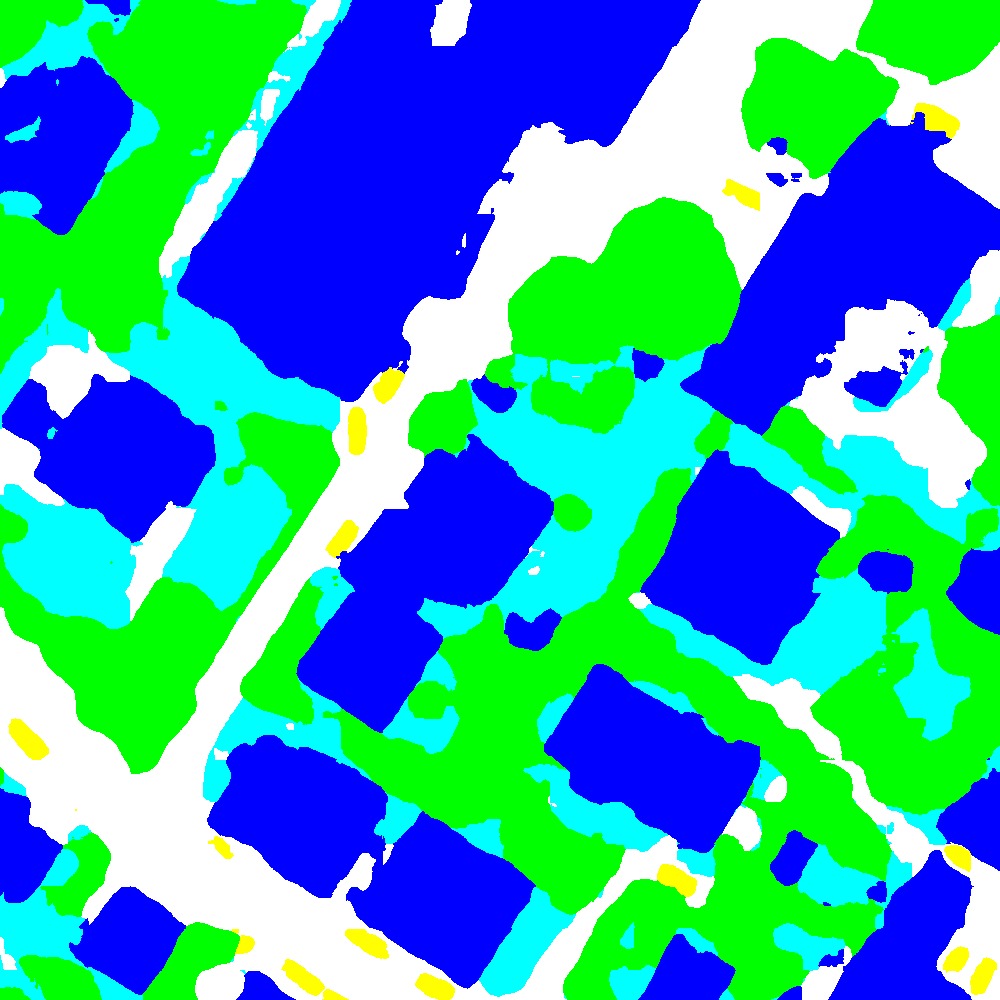} & \\[1.15cm]
				\cincludegraphics[width=\exVaihingen\textwidth]{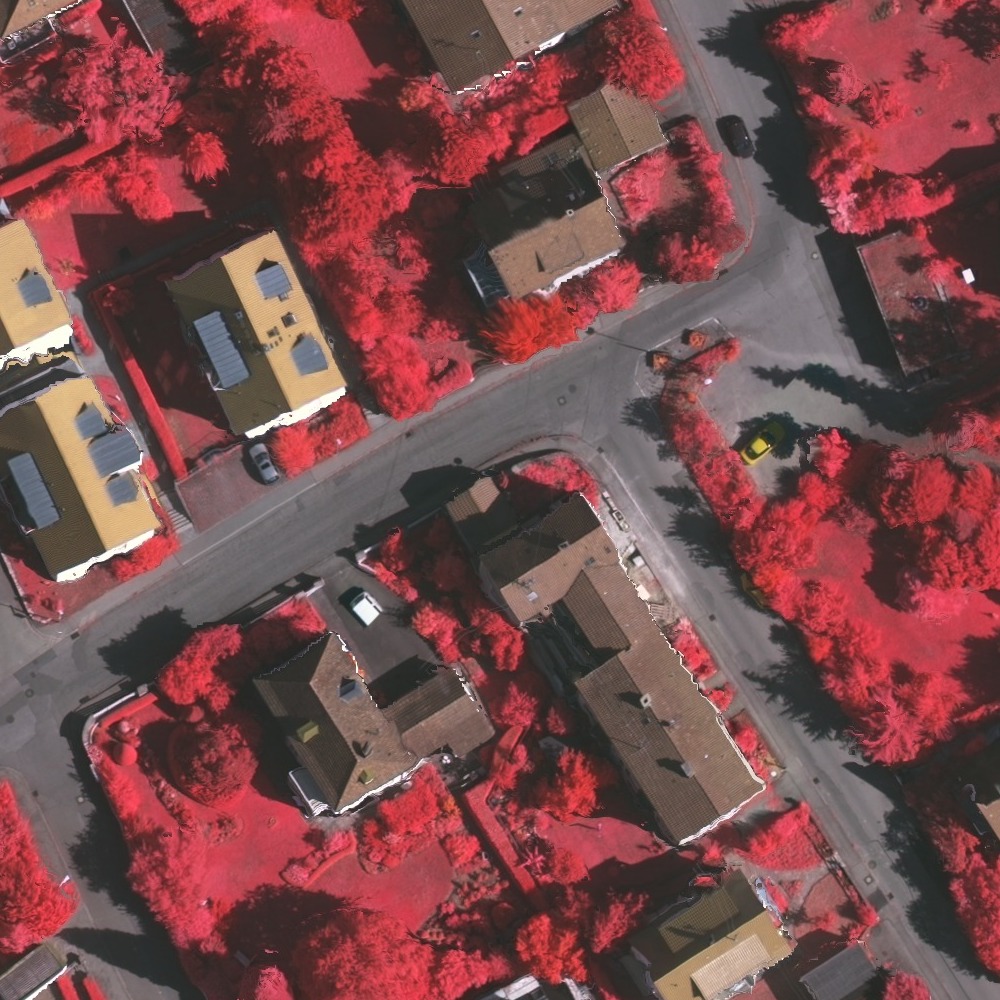} &
				\cincludegraphics[width=\exVaihingen\textwidth]{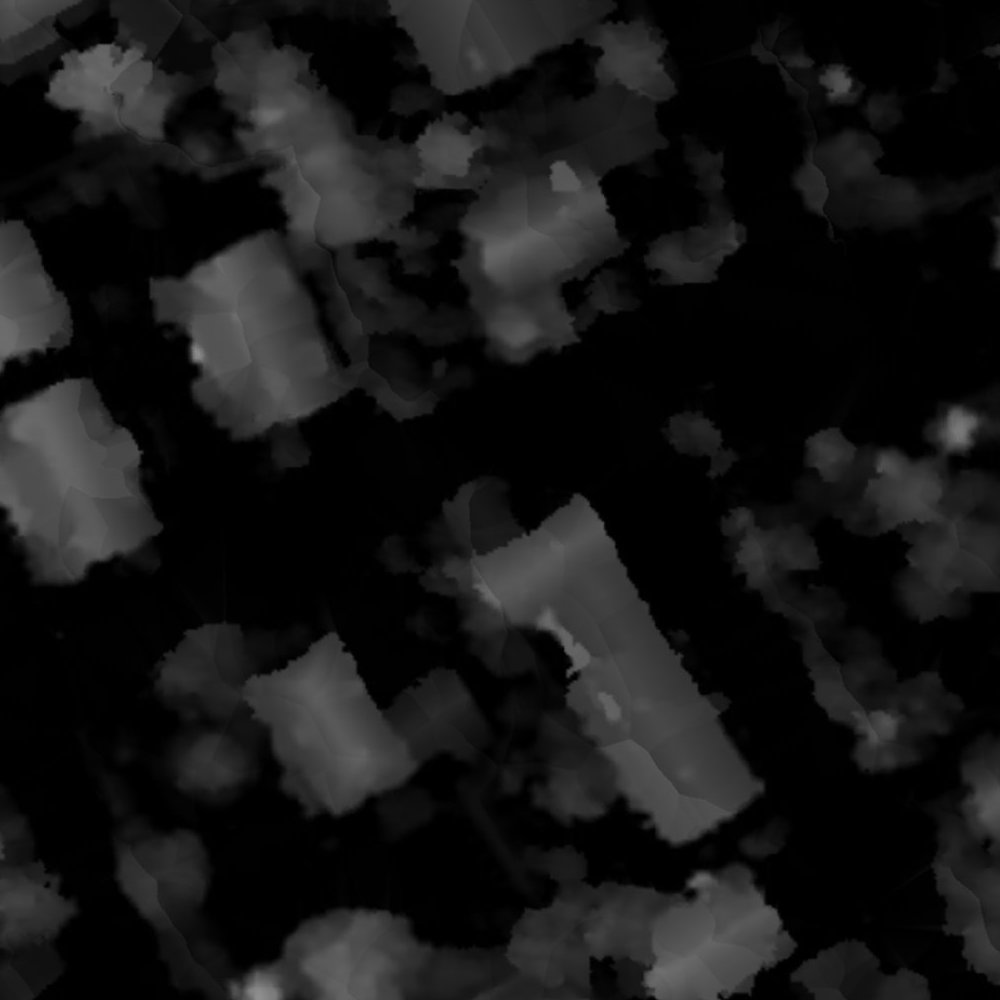} &
				\cincludegraphics[width=\exVaihingen\textwidth]{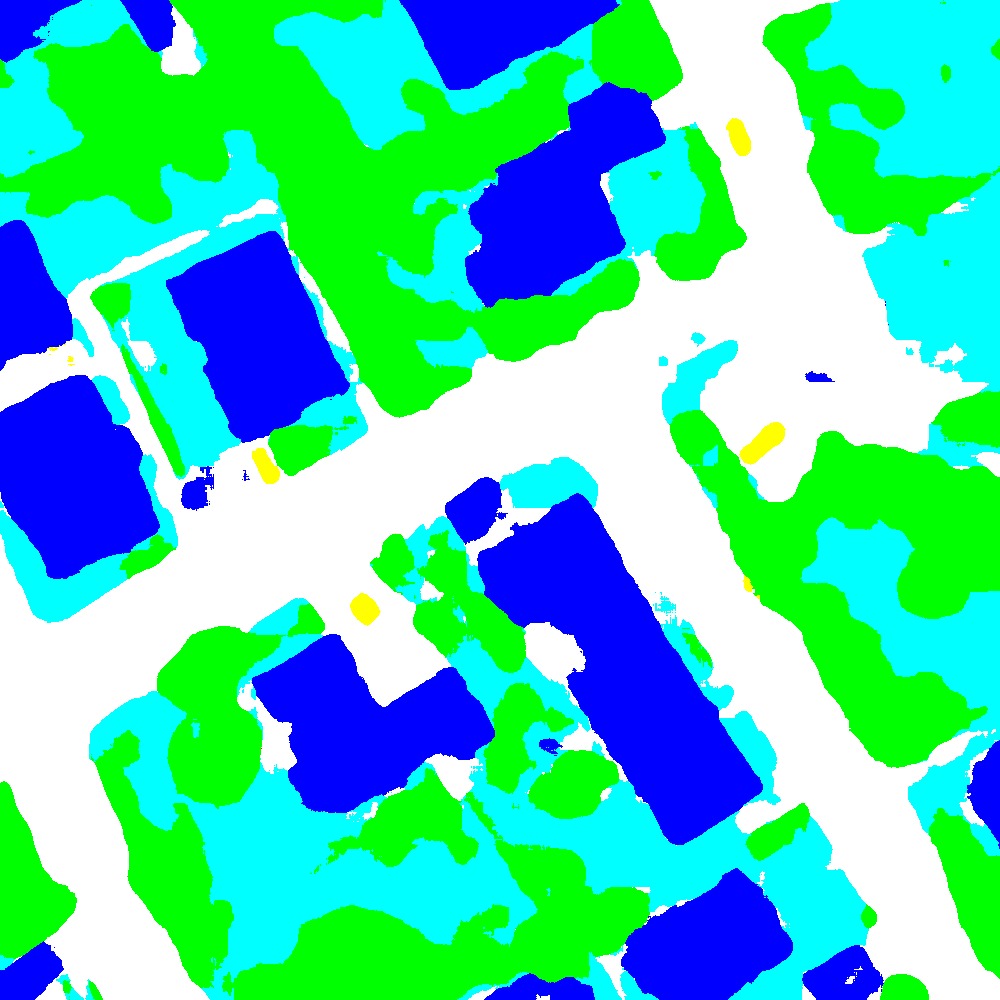} &
				\cincludegraphics[width=\exVaihingen\textwidth]{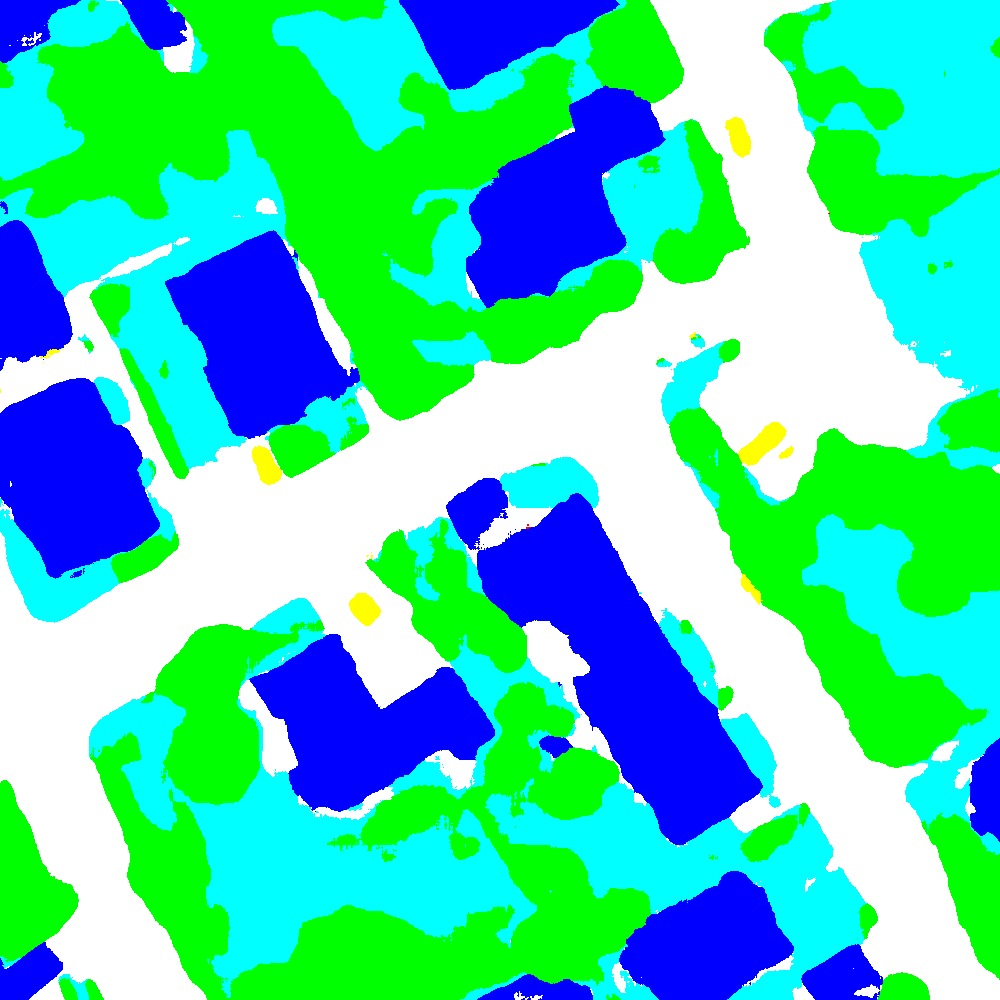} &
				\cincludegraphics[width=\exVaihingen\textwidth]{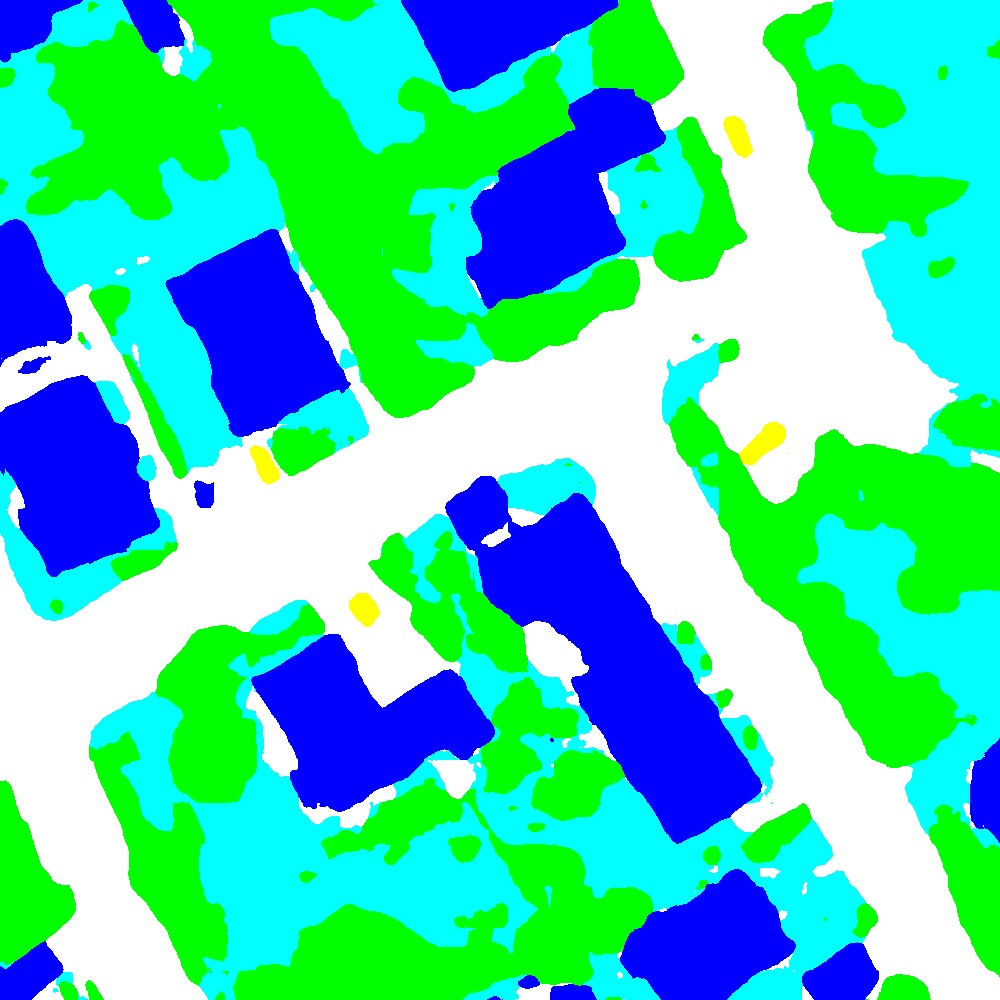} &
				\cincludegraphics[width=\exVaihingen\textwidth]{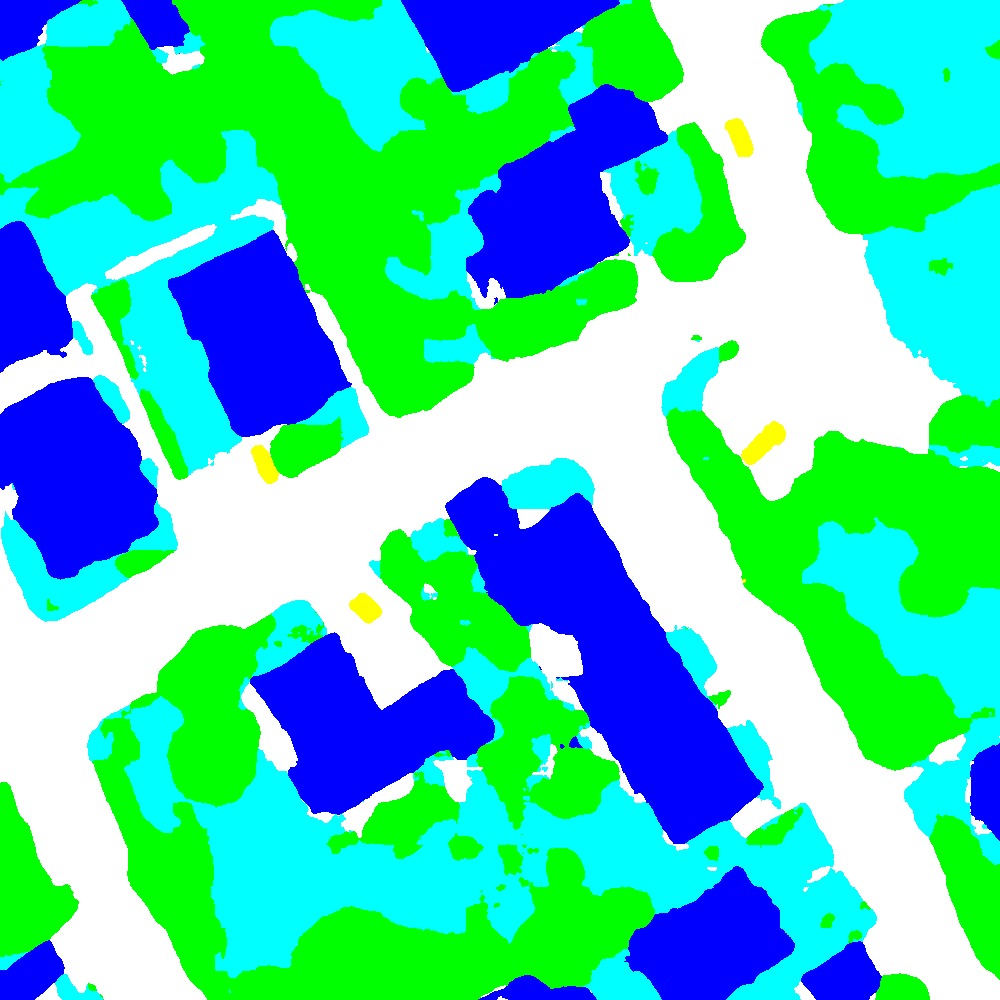} &
				\cincludegraphics[width=\exVaihingen\textwidth]{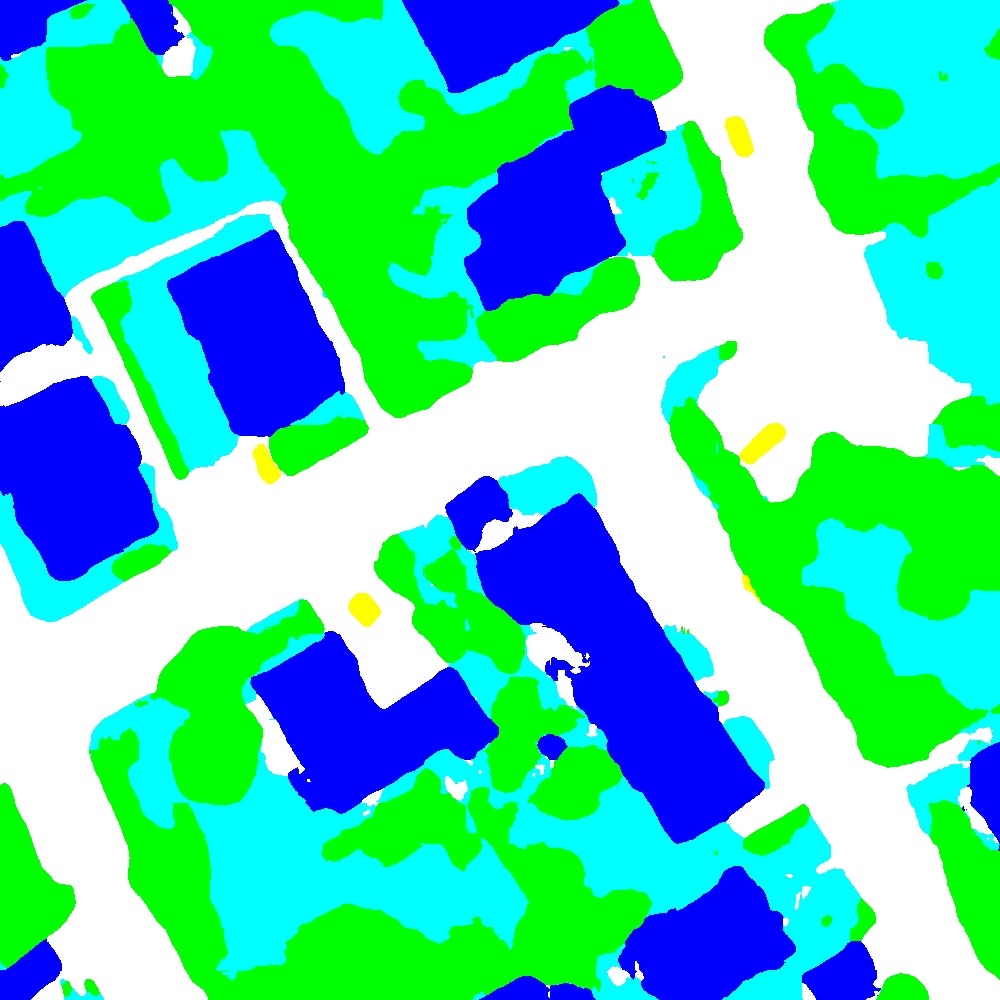} & \\[1.15cm]
			\end{tabular}
		\end{center}
		\captionof{figure}{Example predictions for the test set of the Vaihingen dataset.
			Legend -- White: impervious surfaces. Blue: buildings. Cyan: low vegetation. Green: trees. Yellow: cars. Red: clutter, background.
		}
		\label{fig:vaihingein_test_results}
	\end{table*}
	
	\newcommand{\accVsParamFigSize}{0.24}
	\begin{figure}[t]
		\centering
		\subfloat[Vaihingen]{
			\includegraphics[width=\accVsParamFigSize\textwidth]{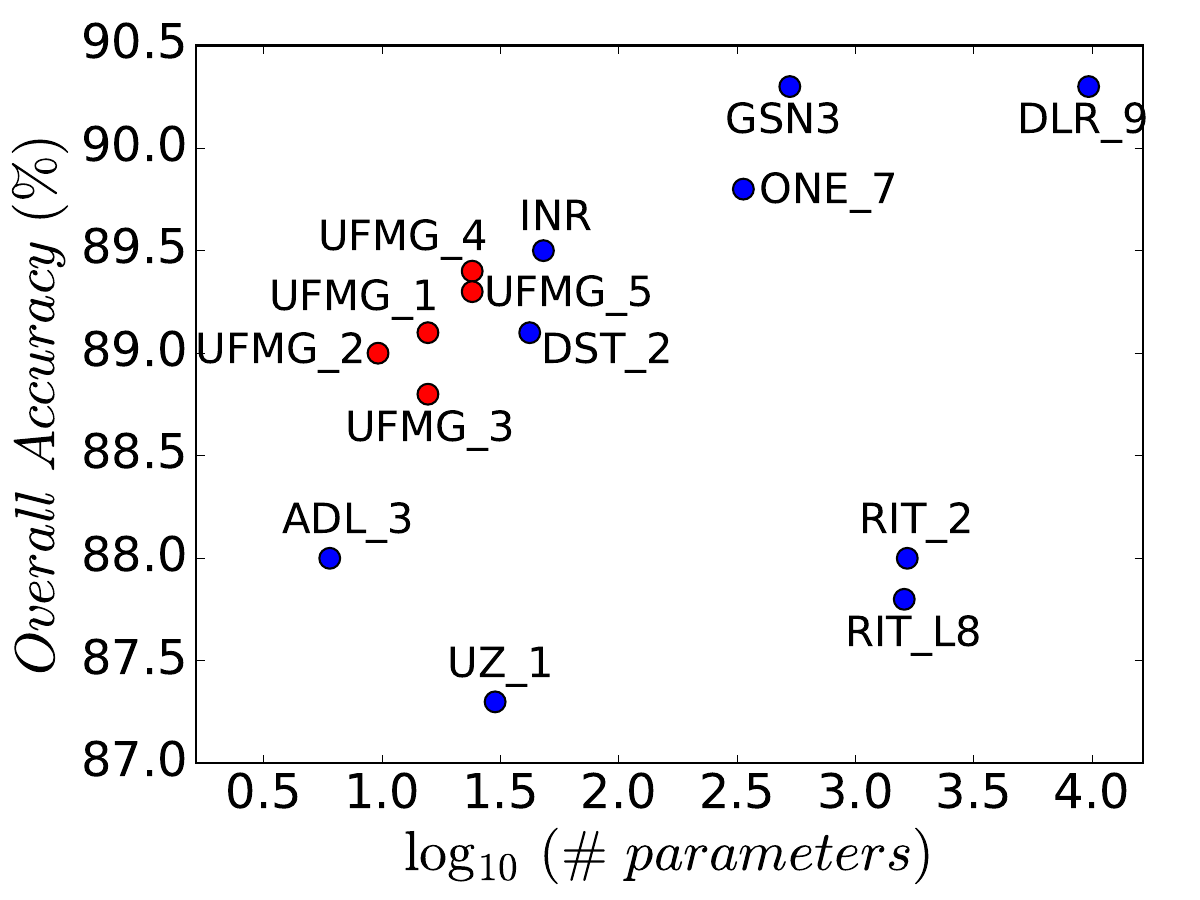}
			\label{vaihingen}
		}
		\subfloat[Potsdam]{
			\includegraphics[width=\accVsParamFigSize\textwidth]{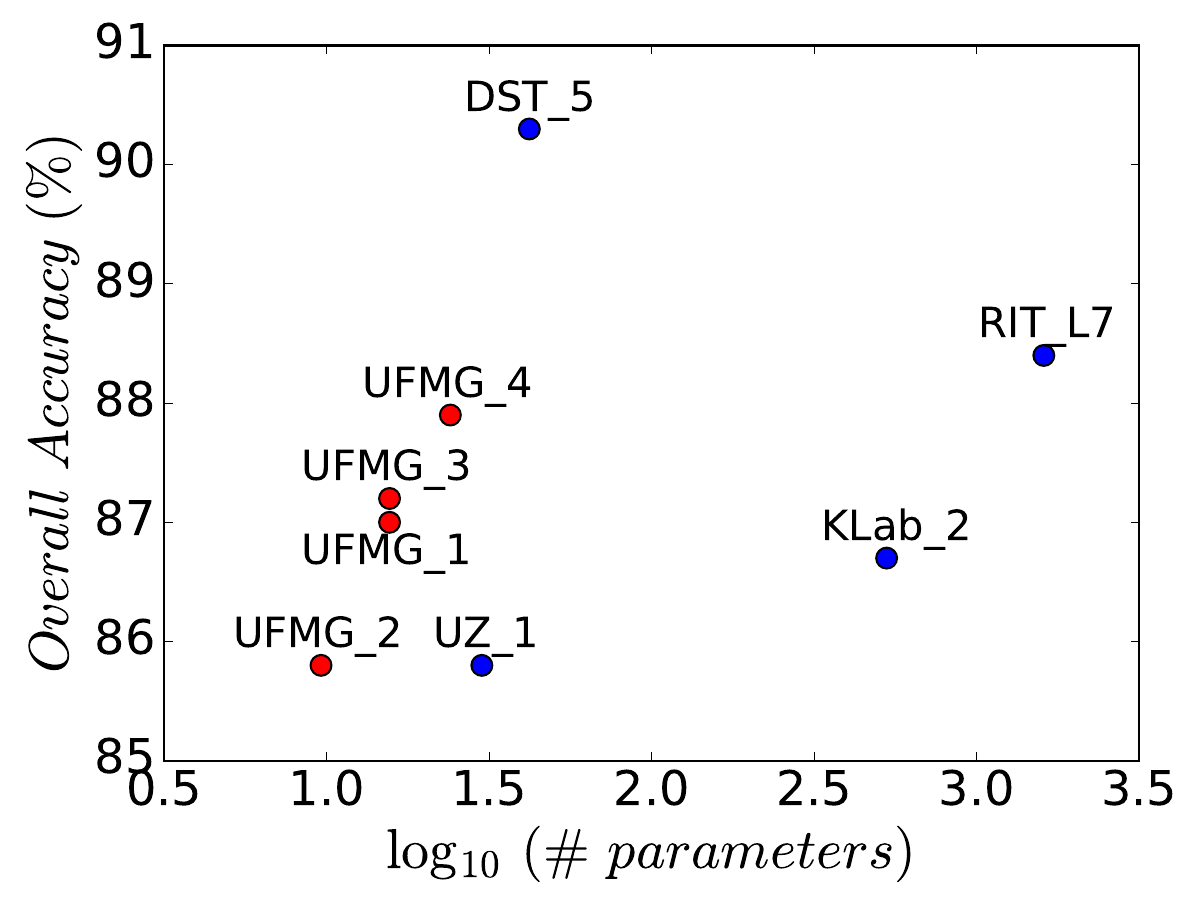}
			\label{potsdam}
		}
		\caption{Comparison, in terms of overall accuracy and number of trainable parameters, between proposed and existing networks for Vaihingen and Potsdam datasets.
			Ideal architectures should be in the top left corner, with fewer parameters but higher accuracy.
			Since the x axis is logarithmic, a change of only 0.3 in this axis is equivalent to more than 1 million new parameters in the model.
		}
		\label{fig:parameters_vs_acc}
	\end{figure}
	
	\subsubsection{Potsdam Dataset} \label{subsec:potsdam_state_results}
	
	As for the Vaihingen dataset, official results for the Potsdam dataset are reported only by the challenge organization.
	For this dataset, we have four submissions, one for each network presented in Section~\ref{sec:methodology} trained with the 6 classes (which are represented, in the official results, as UFMG\_1 to 4).
	In this dataset, there is no need to disregard the clutter/background class, since it has a sufficient amount of samples (4.96\%).
	As before, all submissions related to the proposed work does not use any post-processing.

	Table~\ref{tab:potsdam_results} summarizes some results reported by the organizers.
	Again, besides our results, the table also compiles the \textbf{best results of each work} with enough information to make a fair comparison.
	Visual examples of the proposed method, for the validation and test sets, are presented in Figures~\ref{fig:postdam_validation_results} and~\ref{fig:postdam_test_results}, respectively.
	
	The proposed work achieved competitive results, appearing in third place according to the overall accuracy.
	DST\_5~\cite{sherrah2016fully} and RIT\_L7~\cite{liu2017dense} are the best result in terms of overall accuracy.
	However, they have a larger number of trainable parameters when compared to our proposed networks, as seen in Figure~\ref{potsdam}.
	This outcome corroborates with previous results, reaffirming obtained conclusions.

	\begin{table}[]
		\centering
		\caption{Official results for the Potsdam dataset.}
		\label{tab:potsdam_results}
		\resizebox{\columnwidth}{!}{ 
			\begin{tabular}{@{}lrrrrrrr@{}}
				\toprule
				\multicolumn{1}{c}{\multirow{3}{*}{\textbf{Method}}} & \multicolumn{1}{c}{\multirow{3}{*}{\textbf{\#Parameters}}} & \multicolumn{5}{c}{\textbf{F1 Score}} & \multicolumn{1}{c}{\textbf{\multirow{3}{*}{\begin{tabular}[c]{@{}c@{}}Overall\\Accuracy\end{tabular}}}} \\
				\cmidrule(lr){3-7}
				\multicolumn{1}{c}{} & \multicolumn{1}{c}{} & \multicolumn{1}{c}{\textbf{\begin{tabular}[c]{@{}c@{}}Impervious\\Surface\end{tabular}}} & \multicolumn{1}{c}{\textbf{Building}} & \multicolumn{1}{c}{\textbf{\begin{tabular}[c]{@{}c@{}}Low\\Vegetation\end{tabular}}} & \multicolumn{1}{c}{\textbf{Tree}} & \multicolumn{1}{c}{\textbf{Car}} & \multicolumn{1}{c}{} \\ 
				\midrule
				DST\_5~\cite{sherrah2016fully} 			& $3.5\cdot10^6$ & 92.5 & 96.4 & 86.7 & 88.0 & 94.7 & 90.3 \\
				RIT\_L7~\cite{liu2017dense} 			& $134\cdot10^6$ & 91.2 & 94.6 & 85.1 & 85.1 & 92.8 & 88.4 \\
				\textbf{UFMG\_4} 		& $2\cdot10^6$   & 90.8 & 95.6 & 84.4 & 84.3 & 92.4 & 87.9 \\  
				\textbf{UFMG\_3} 		& $1.3\cdot10^6$ & 90.5 & 95.6 & 83.3 & 82.6 & 90.8 & 87.2 \\  
				\textbf{UFMG\_1}		 		& $1.3\cdot10^6$ & 90.1 & 95.6 & 83.7 & 82.4 & 91.3 & 87.0 \\  
				KLab\_2~\cite{kemker2017}				& $44\cdot10^6$  & 89.7 & 92.7 & 83.7 & 84.0 & 92.1 & 86.7 \\
				\textbf{UFMG\_2} 		& $0.8\cdot10^6$ & 88.7 & 95.3 & 83.1 & 80.8 & 90.8 & 85.8 \\  
				UZ\_1~\cite{volpi2017dense} 			& $2.5\cdot10^6$ & 89.3 & 95.4 & 81.8 & 80.5 & 86.5 & 85.8 \\
				\bottomrule
			\end{tabular}
		}
	\end{table}

	\begin{table*}[t]
		\begin{center}
			\begin{tabular}{>{\centering\arraybackslash} m{1.95cm} >{\centering\arraybackslash}m{1.95cm} >{\centering\arraybackslash}m{1.95cm} >{\centering\arraybackslash}m{1.95cm} >{\centering\arraybackslash}m{1.95cm} >{\centering\arraybackslash}m{1.95cm} >{\centering\arraybackslash}m{1.95cm} >{\centering\arraybackslash} m{1.95cm} @{}m{0pt}@{} } 
				\textbf{Image} & \textbf{nDSM} & \textbf{Ground-Truth} & \textbf{Dilated6} & \textbf{DenseDilated6} & \textbf{Dilated6 Pooling} & \textbf{Dilated8 Pooling} & \textbf{Dilated8 Pooling} & \\
				
				\cincludegraphics[width=\exVaihingen\textwidth]{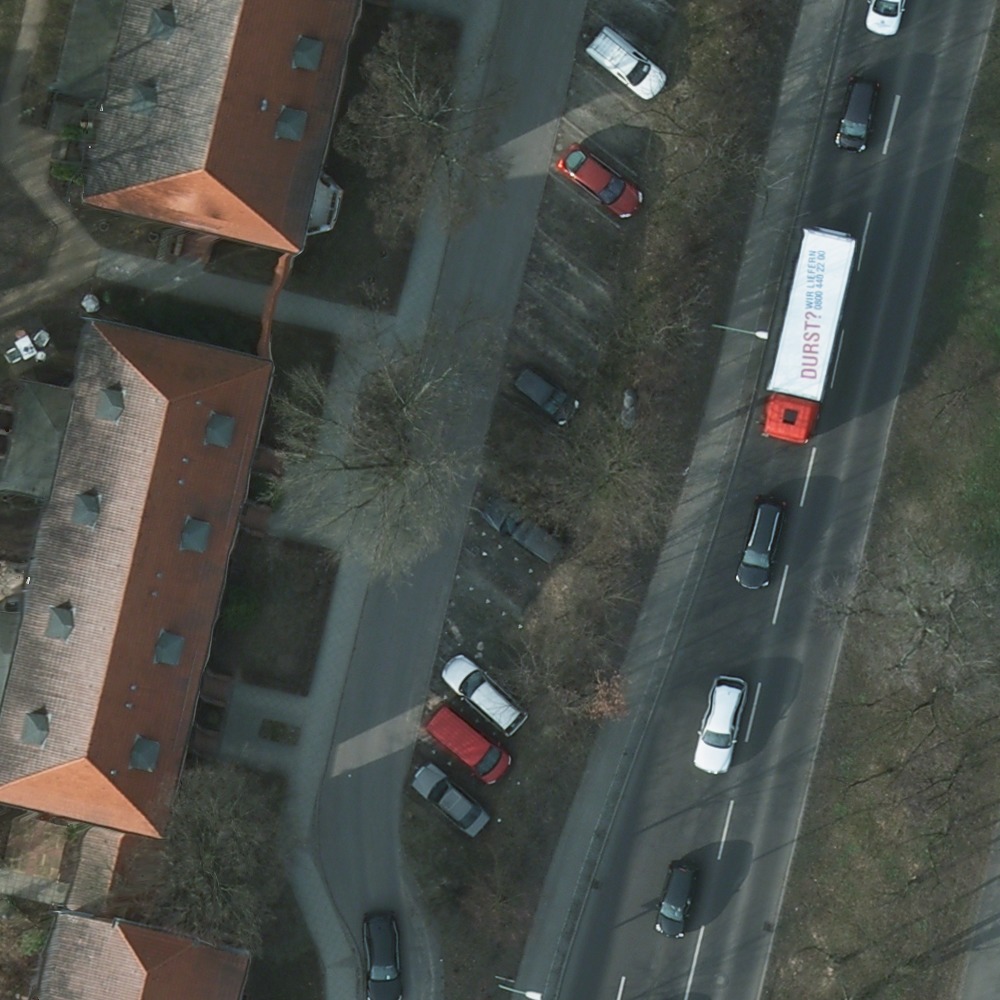} &
				\cincludegraphics[width=\exVaihingen\textwidth]{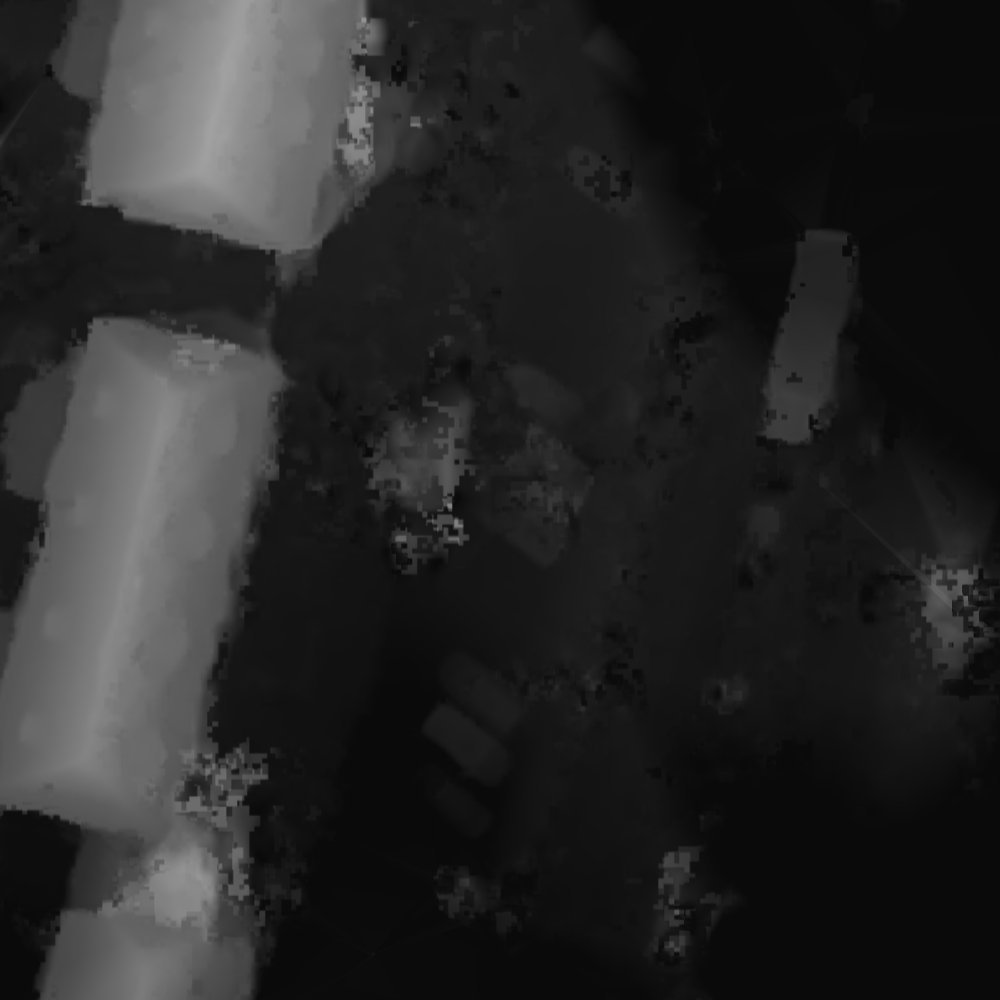} &
				\cincludegraphics[width=\exVaihingen\textwidth]{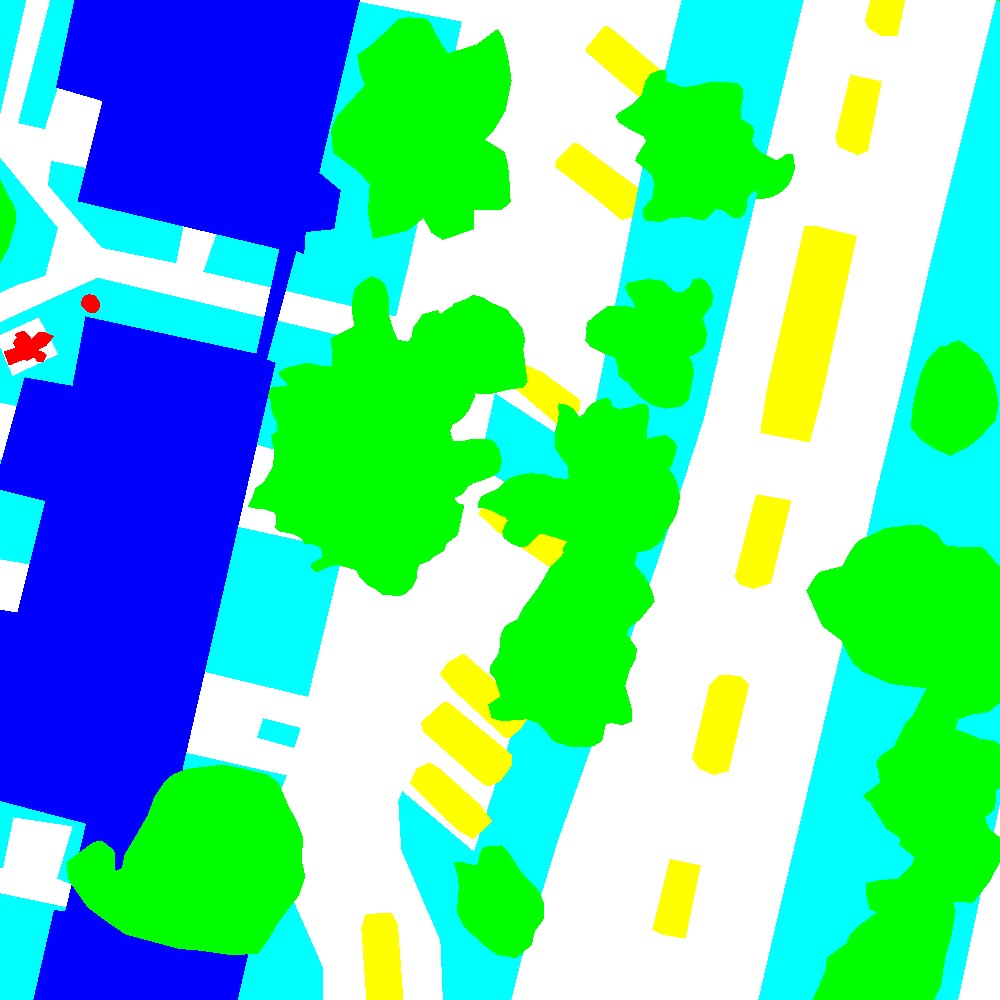} &
				\cincludegraphics[width=\exVaihingen\textwidth]{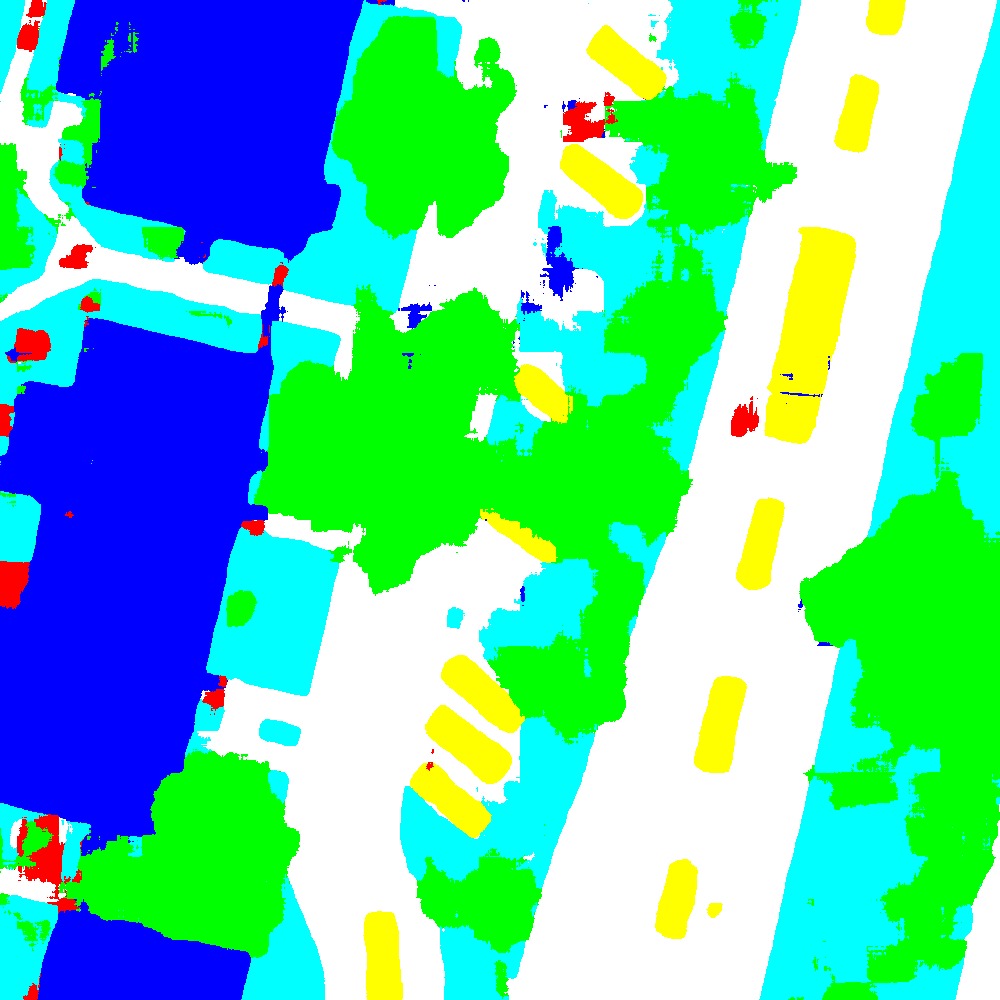} &
				\cincludegraphics[width=\exVaihingen\textwidth]{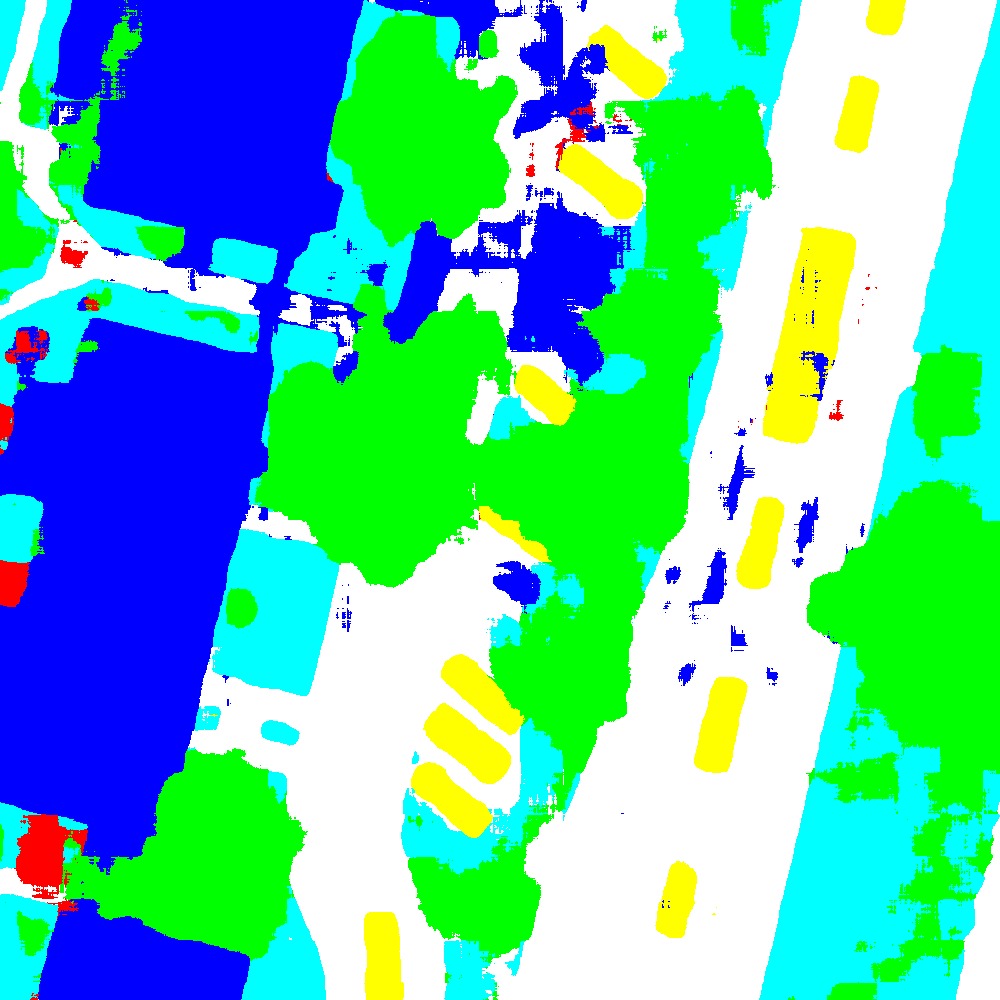} &
				\cincludegraphics[width=\exVaihingen\textwidth]{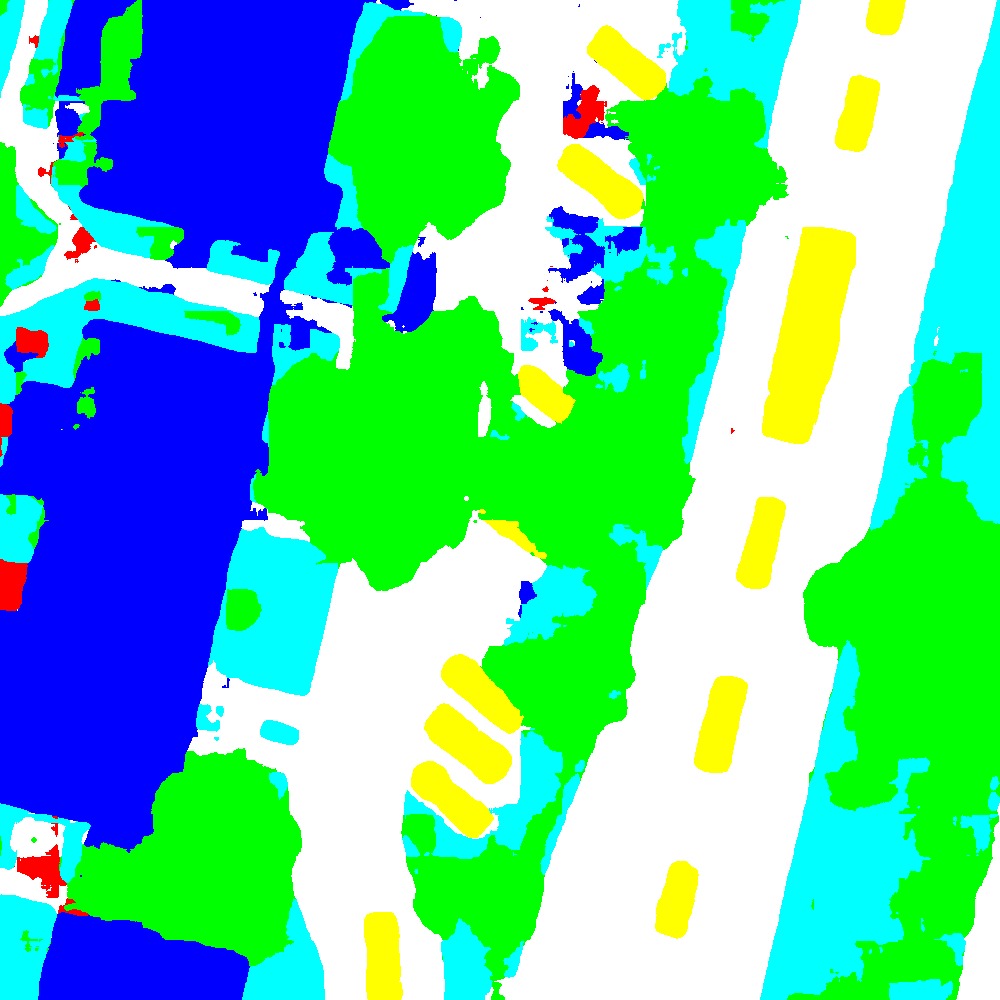} &
				\cincludegraphics[width=\exVaihingen\textwidth]{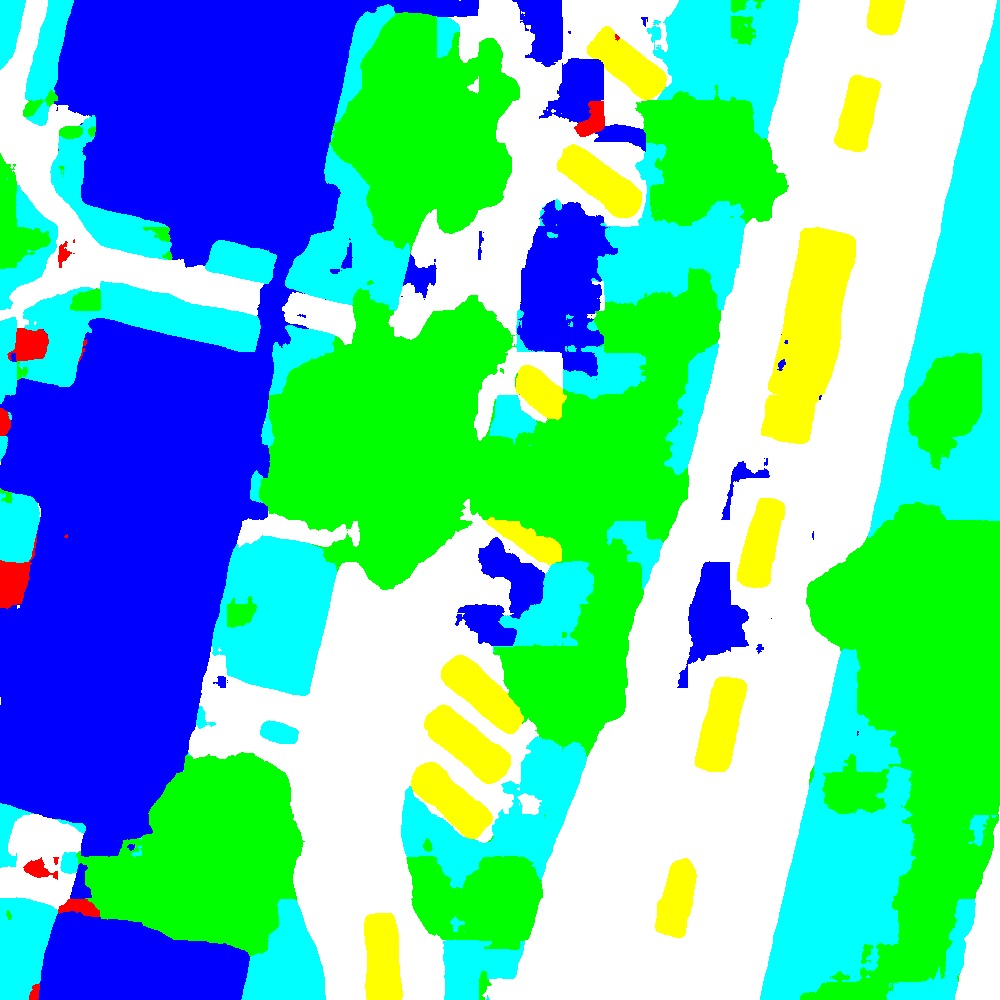} &
				\cincludegraphics[width=\exVaihingen\textwidth]{area2_12_dilated8_1000.jpeg} & \\[1.15cm]
				\cincludegraphics[width=\exVaihingen\textwidth]{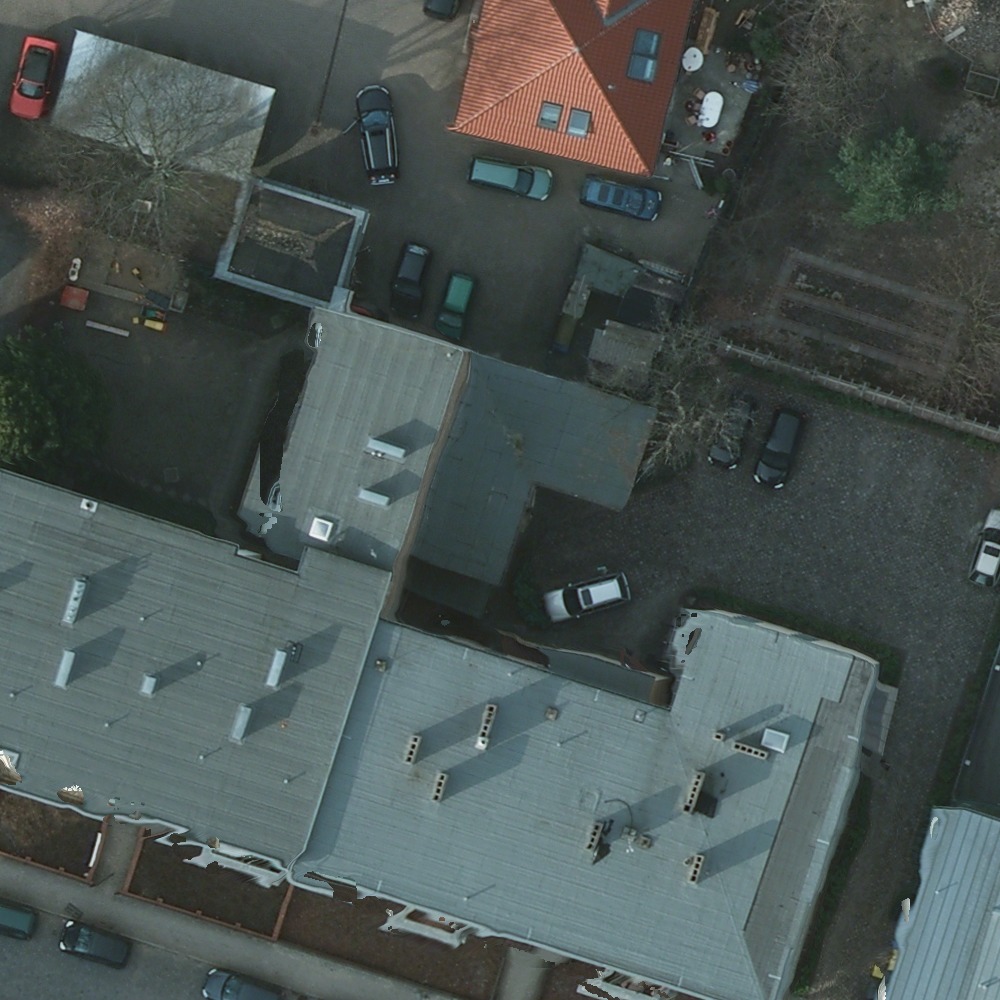} &
				\cincludegraphics[width=\exVaihingen\textwidth]{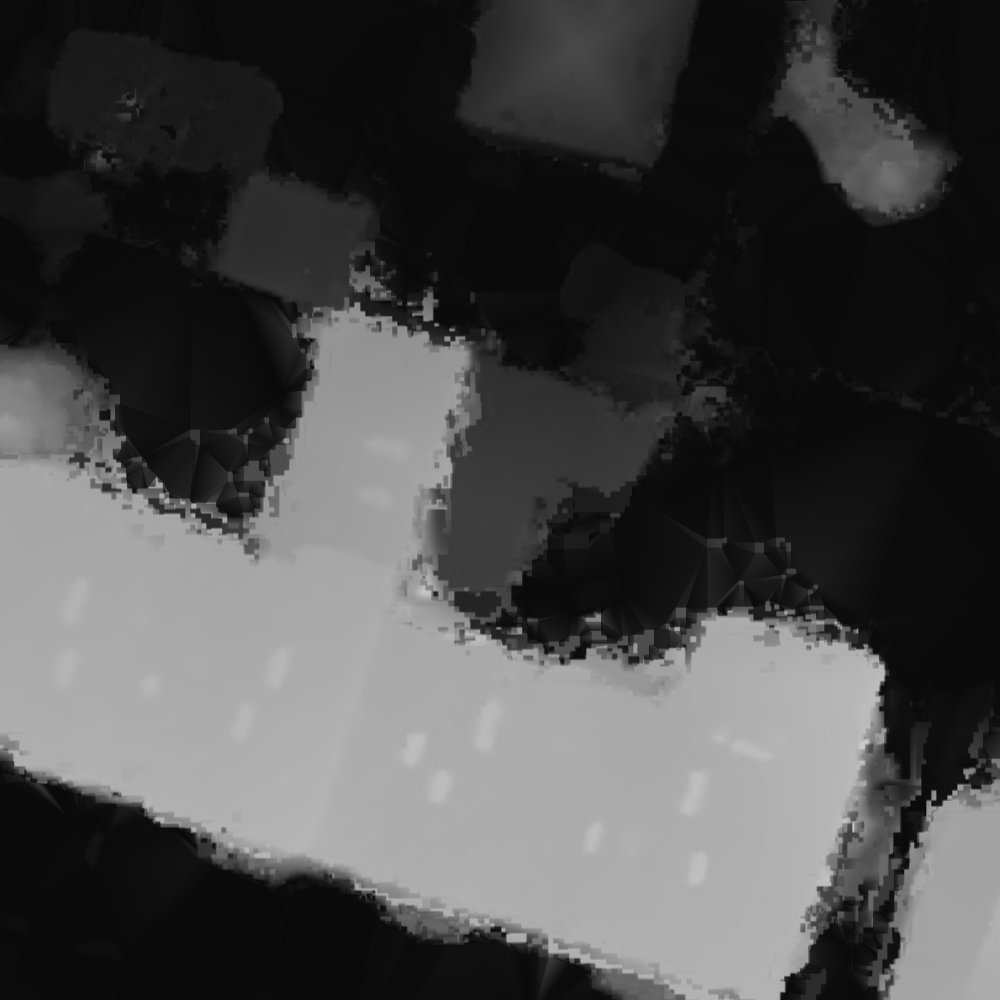} &
				\cincludegraphics[width=\exVaihingen\textwidth]{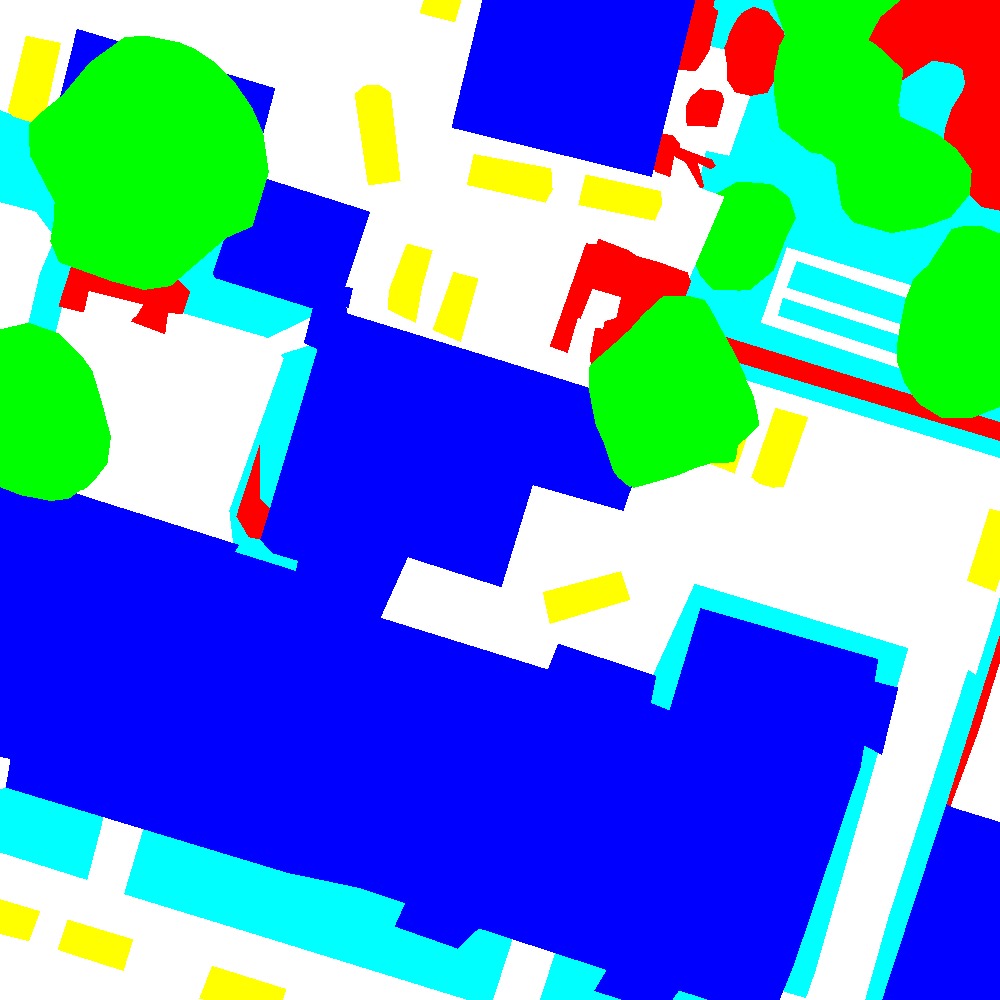} &
				\cincludegraphics[width=\exVaihingen\textwidth]{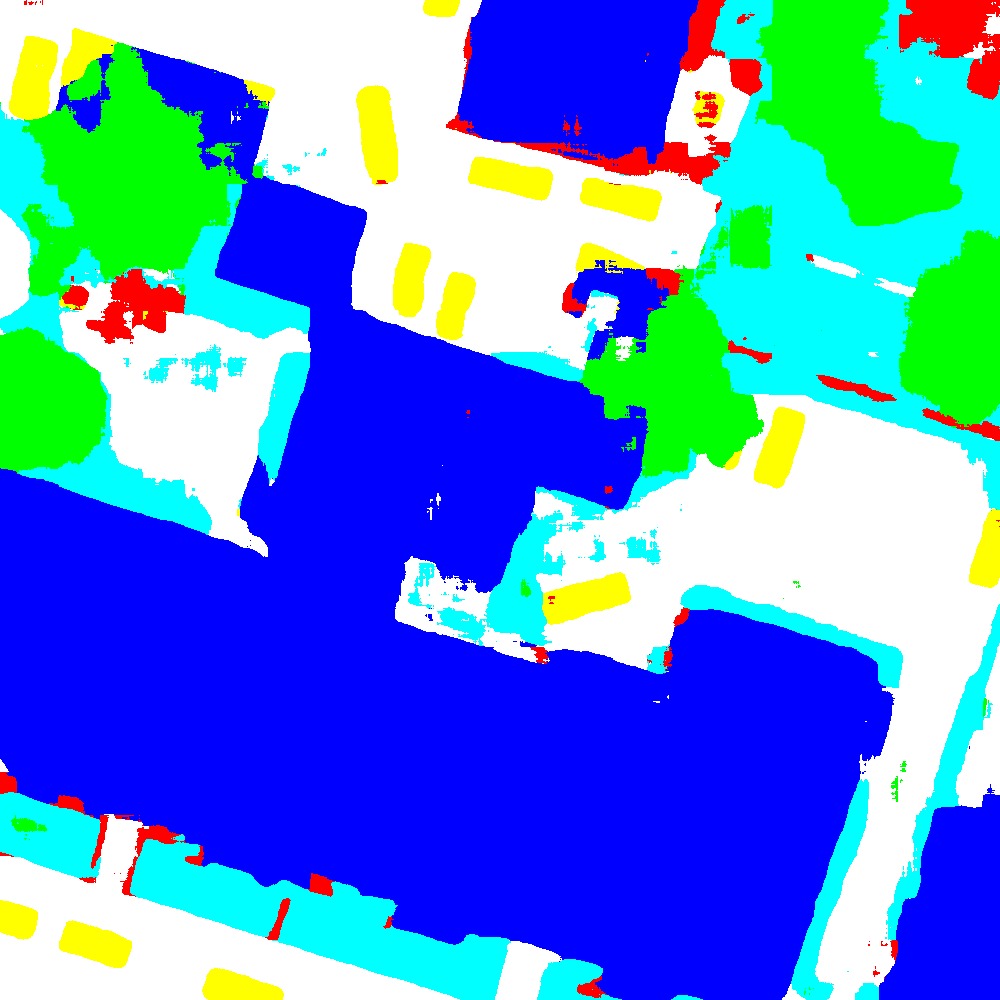} &
				\cincludegraphics[width=\exVaihingen\textwidth]{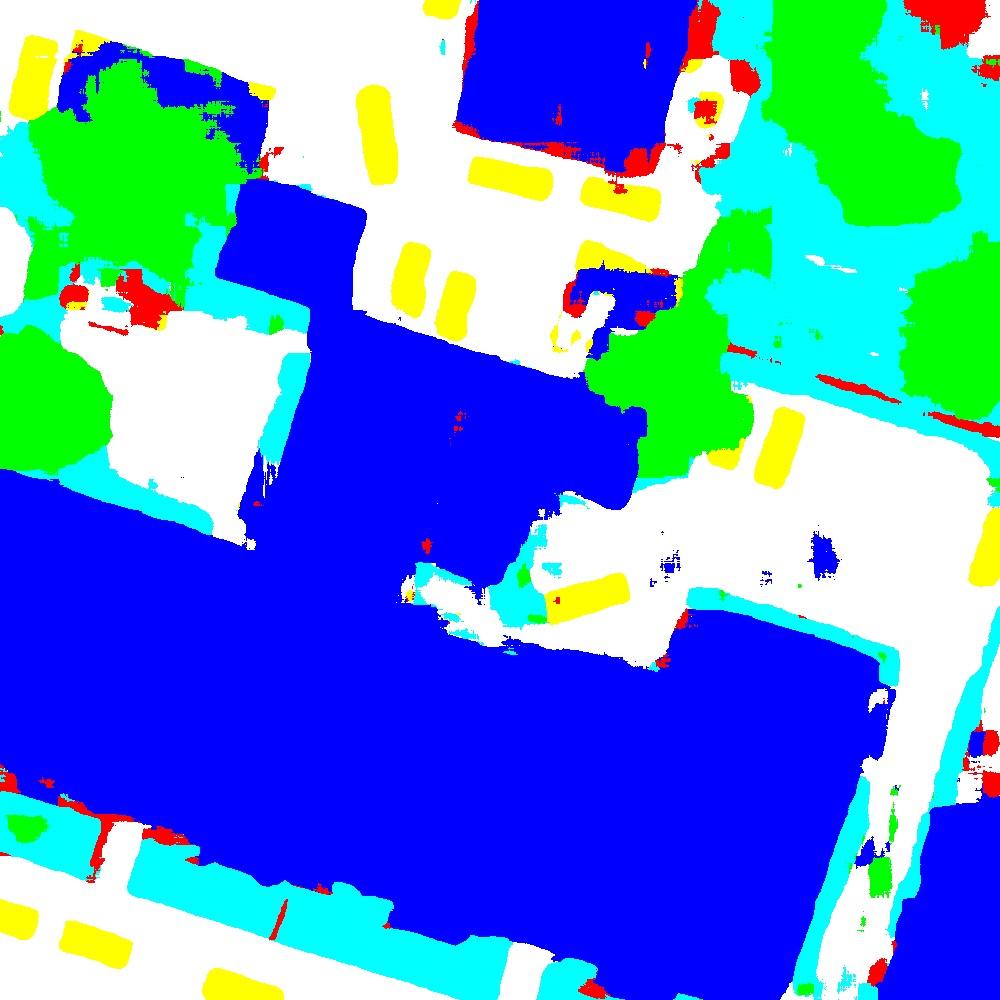} &
				\cincludegraphics[width=\exVaihingen\textwidth]{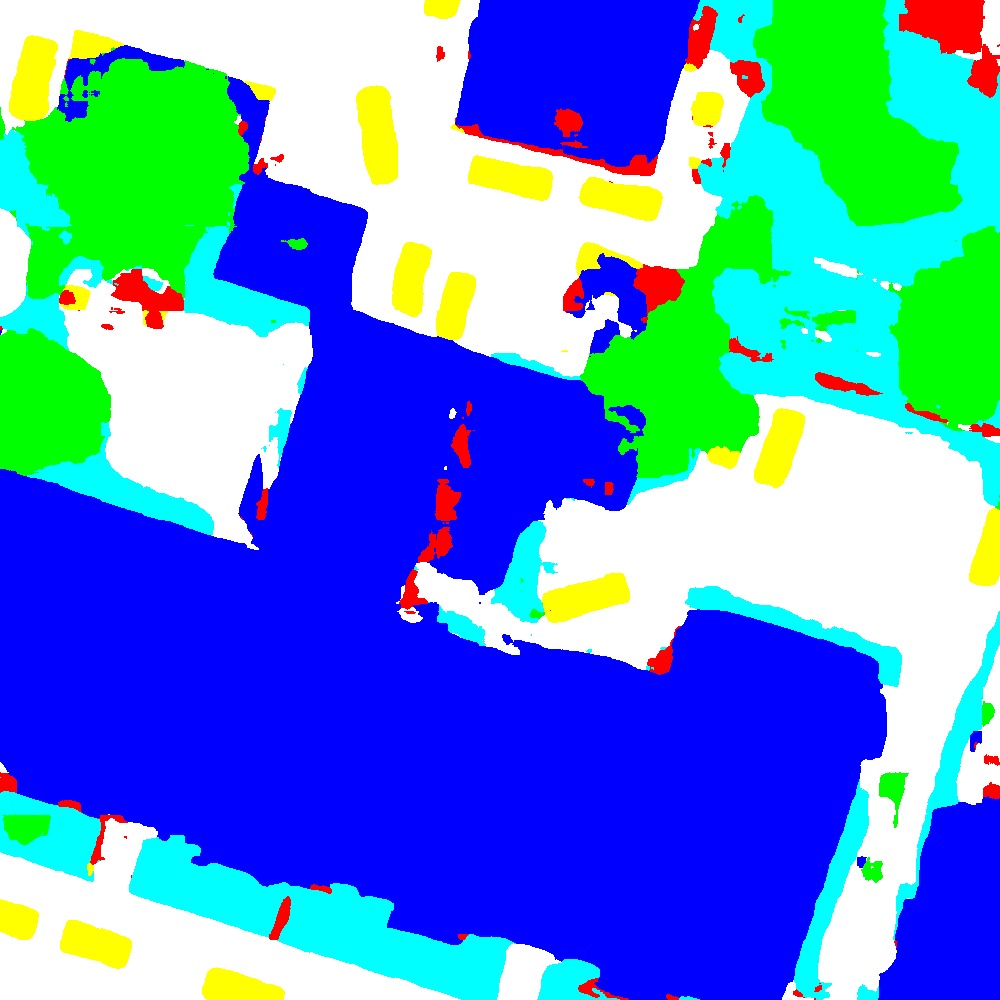} &
				\cincludegraphics[width=\exVaihingen\textwidth]{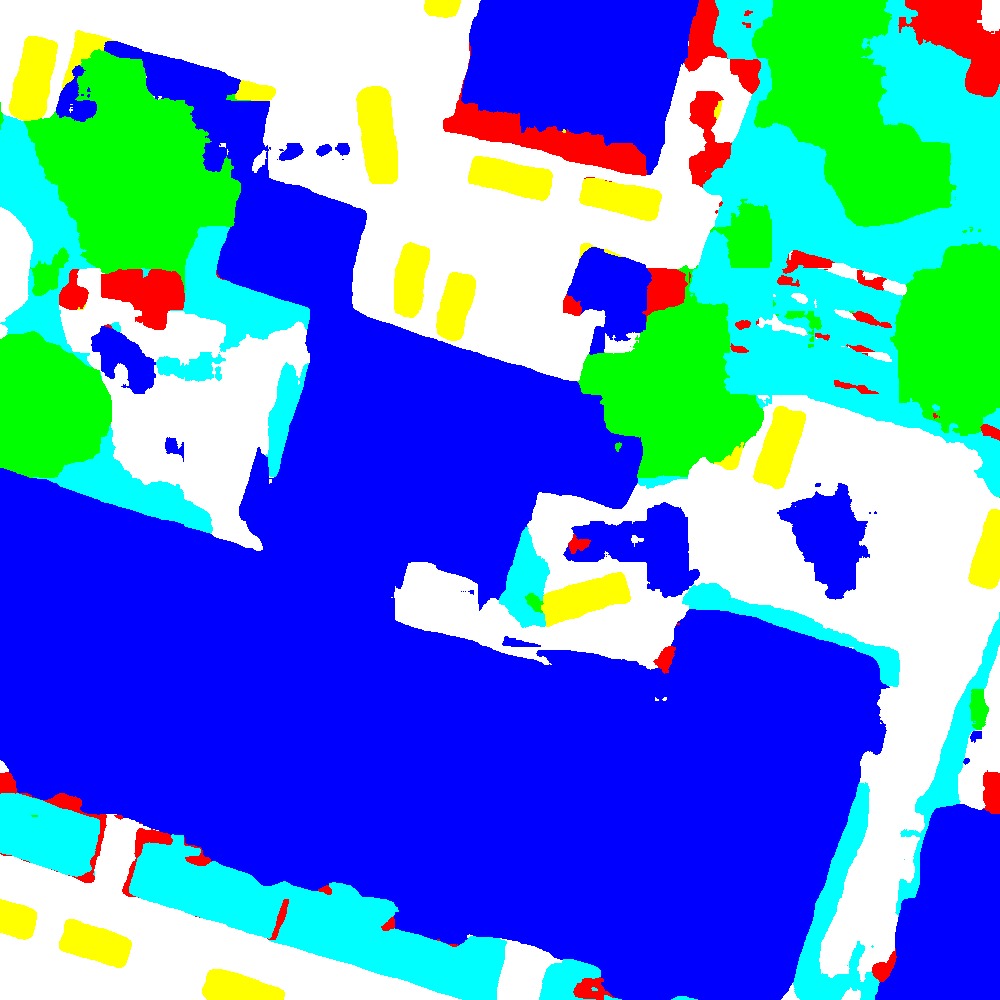} &
				\cincludegraphics[width=\exVaihingen\textwidth]{area3_12_dilated8_1000.jpeg} & \\[1.15cm]
			\end{tabular}
		\end{center}
		\captionof{figure}{Example predictions for the validation set of the Potsdam dataset.
			Legend -- White: impervious surfaces. Blue: buildings. Cyan: low vegetation. Green: trees. Yellow: cars. Red: clutter, background.
		}
		\label{fig:postdam_validation_results}
	\end{table*}
	
	\begin{table*}[t]
		\begin{center}
			\begin{tabular}{>{\centering\arraybackslash} m{1.95cm} >{\centering\arraybackslash}m{1.95cm} >{\centering\arraybackslash}m{1.95cm} >{\centering\arraybackslash}m{1.95cm} >{\centering\arraybackslash}m{1.95cm} >{\centering\arraybackslash}m{1.95cm} >{\centering\arraybackslash}m{1.95cm} @{}m{0pt}@{} } 
				\textbf{Image} & \textbf{nDSM} & \textbf{Dilated6} & \textbf{DenseDilated6} & \textbf{Dilated6 Pooling} & \textbf{Dilated8 Pooling} & \textbf{Dilated8 Pooling} & \\
				\cincludegraphics[width=\exVaihingen\textwidth]{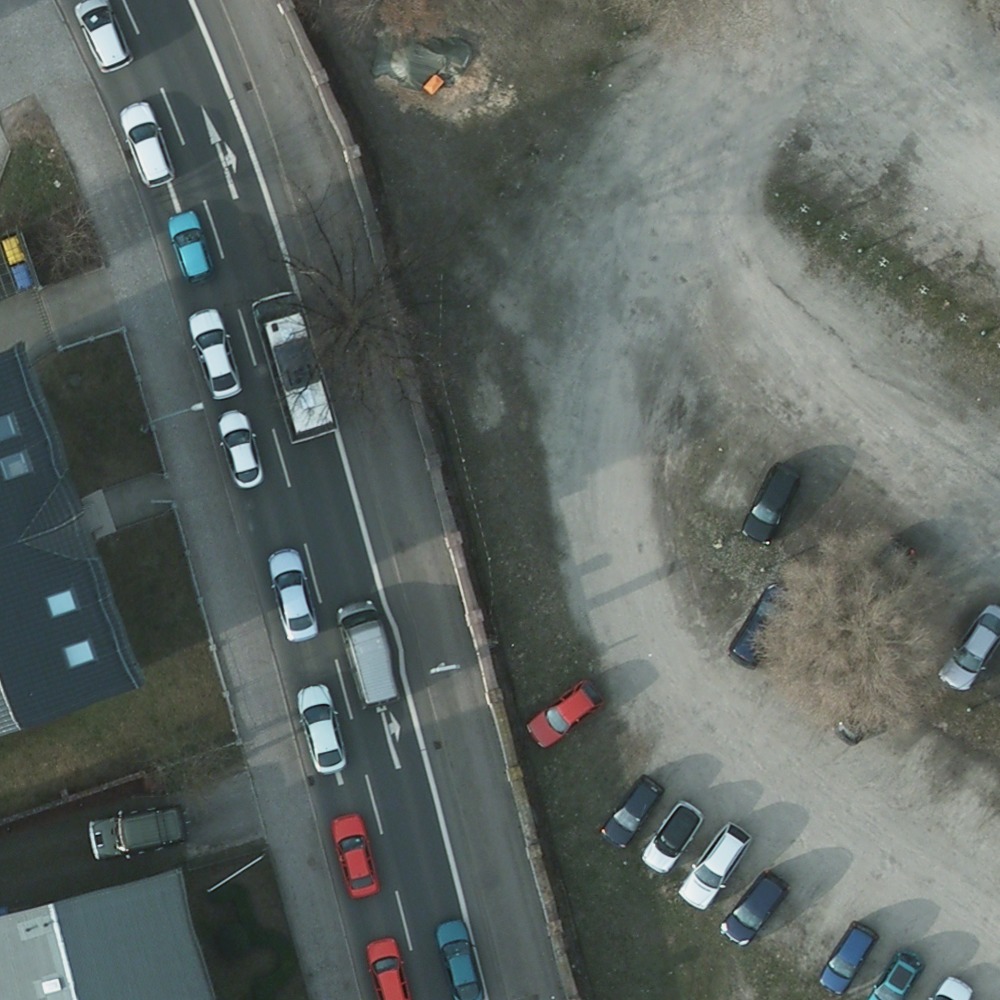} &
				\cincludegraphics[width=\exVaihingen\textwidth]{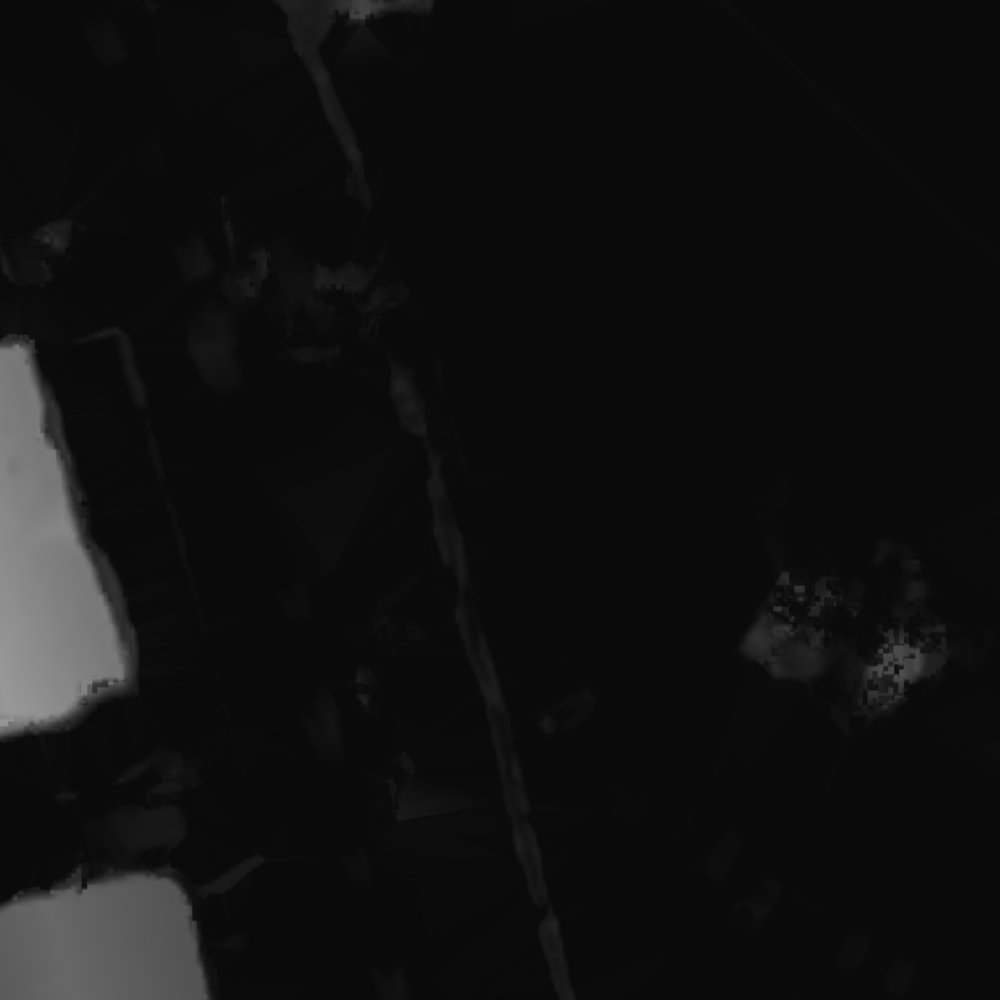} &
				\cincludegraphics[width=\exVaihingen\textwidth]{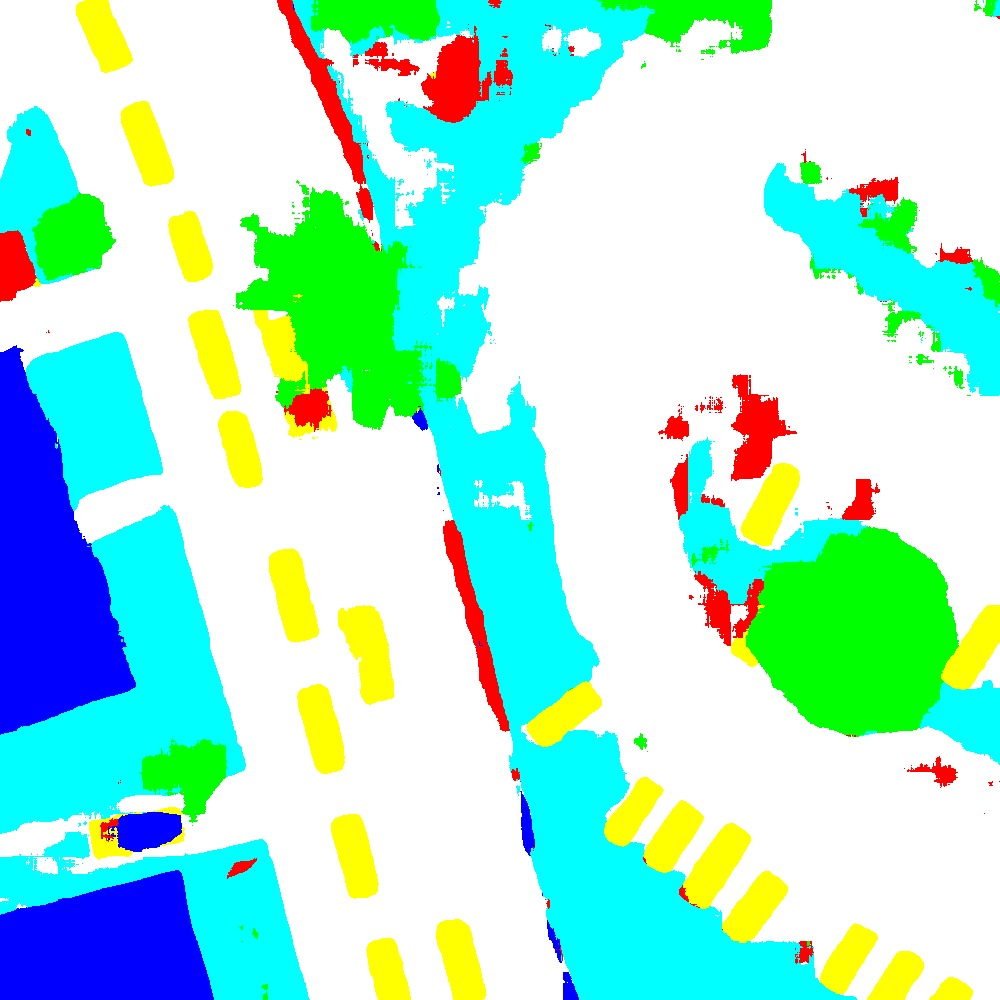} &
				\cincludegraphics[width=\exVaihingen\textwidth]{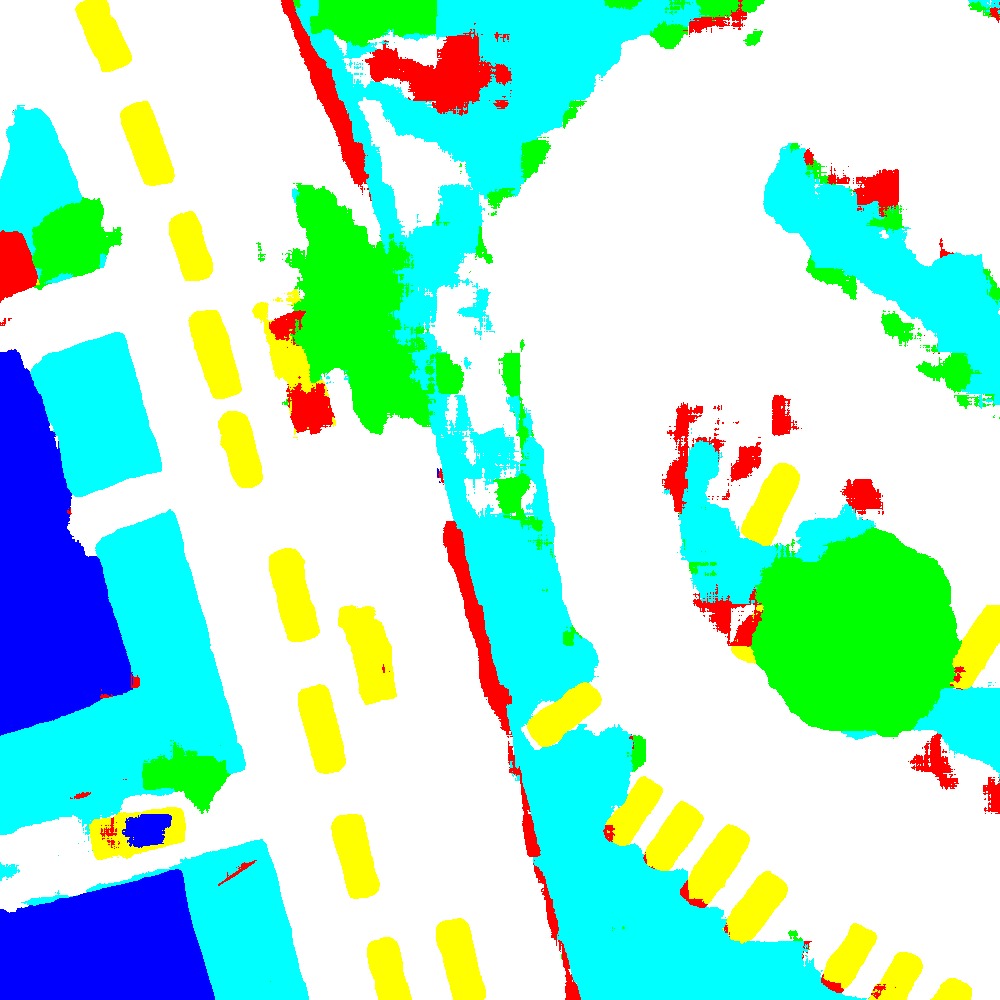} &
				\cincludegraphics[width=\exVaihingen\textwidth]{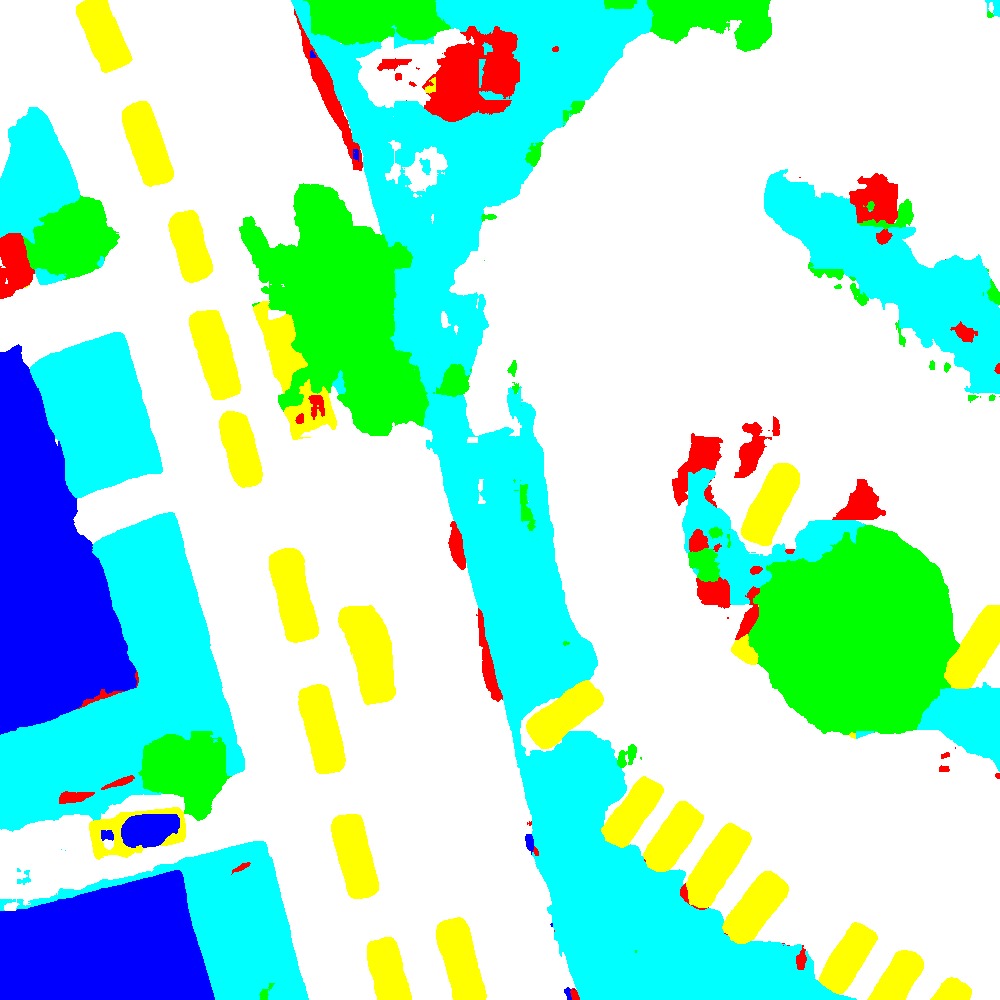} &
				\cincludegraphics[width=\exVaihingen\textwidth]{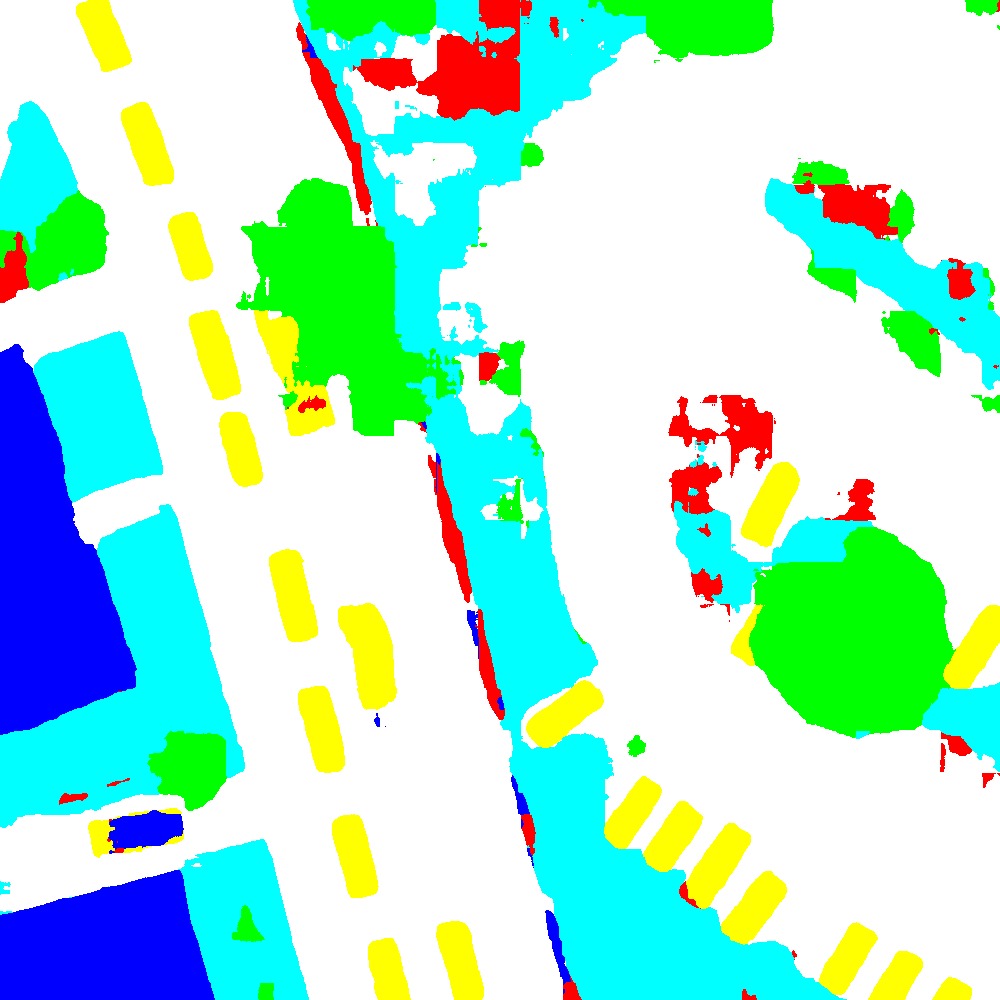} &
				\cincludegraphics[width=\exVaihingen\textwidth]{area4_15_dilated8_1000.jpeg} & \\[1.15cm]
				\cincludegraphics[width=\exVaihingen\textwidth]{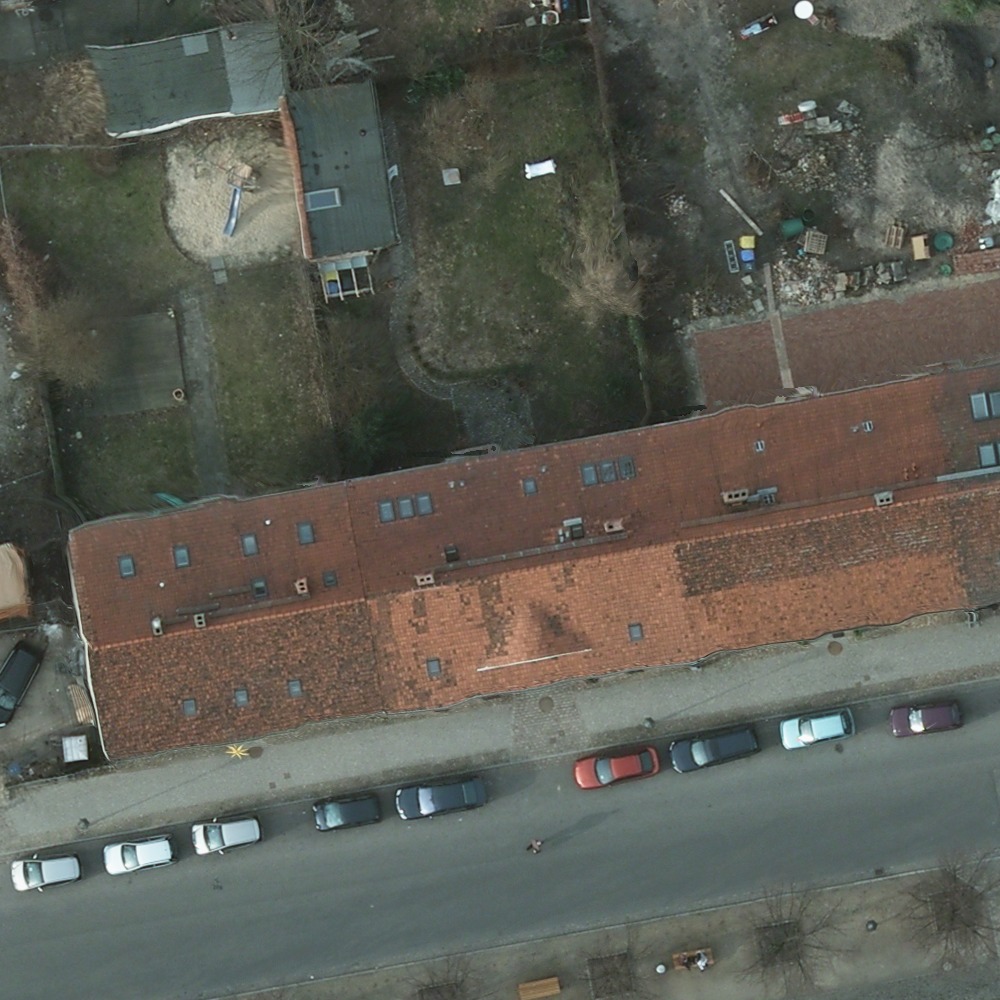} &
				\cincludegraphics[width=\exVaihingen\textwidth]{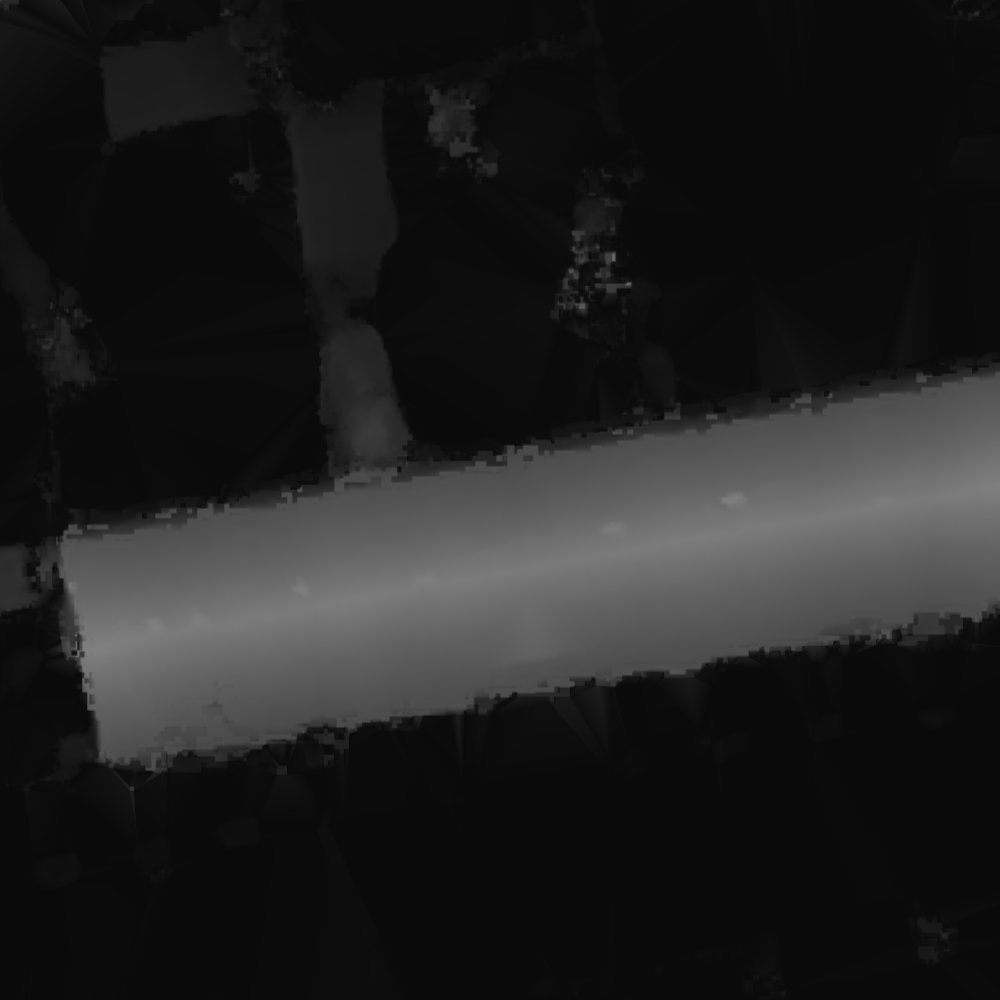} &
				\cincludegraphics[width=\exVaihingen\textwidth]{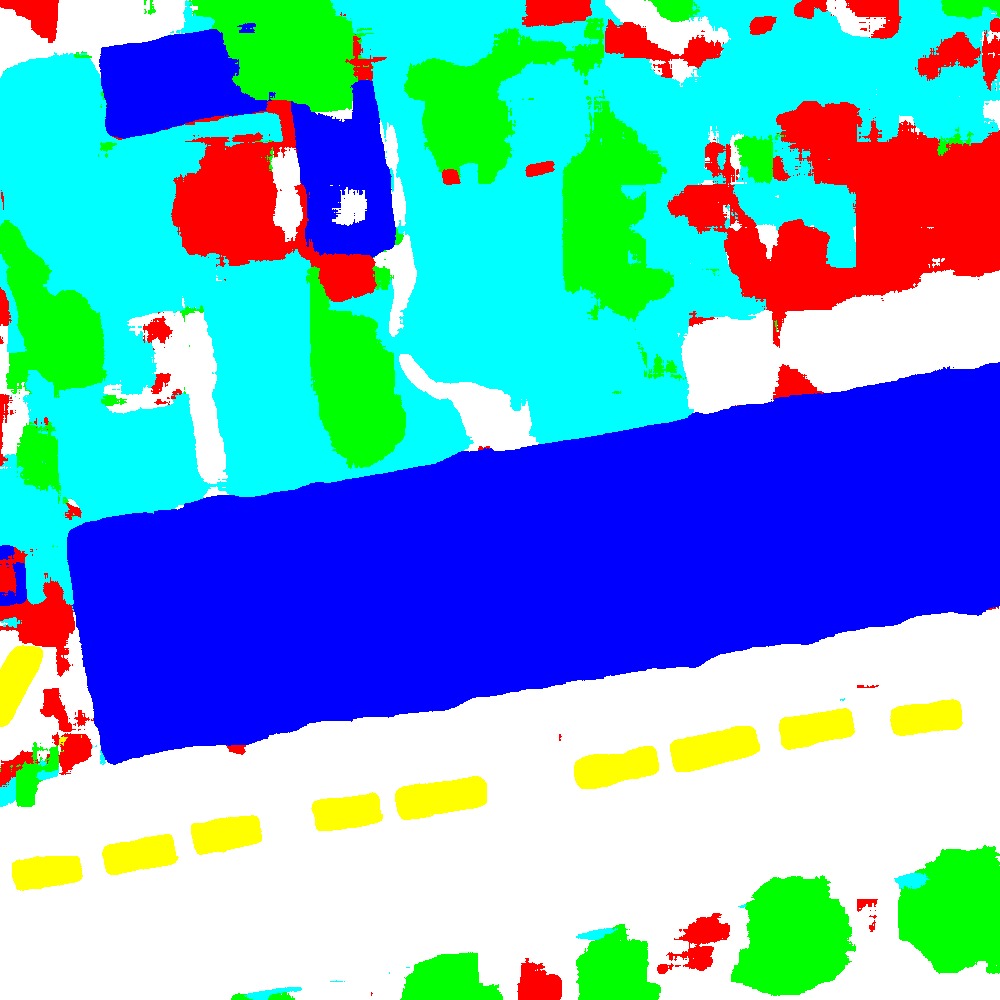} &
				\cincludegraphics[width=\exVaihingen\textwidth]{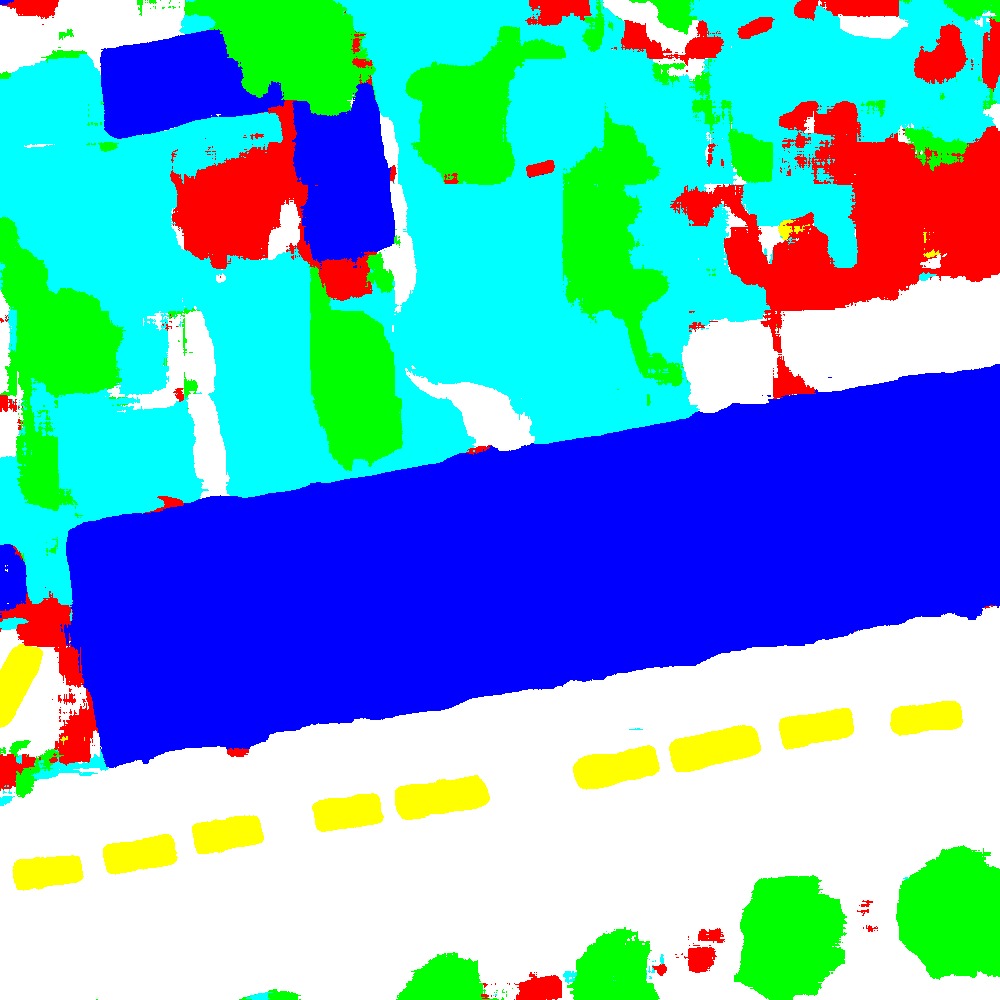} &
				\cincludegraphics[width=\exVaihingen\textwidth]{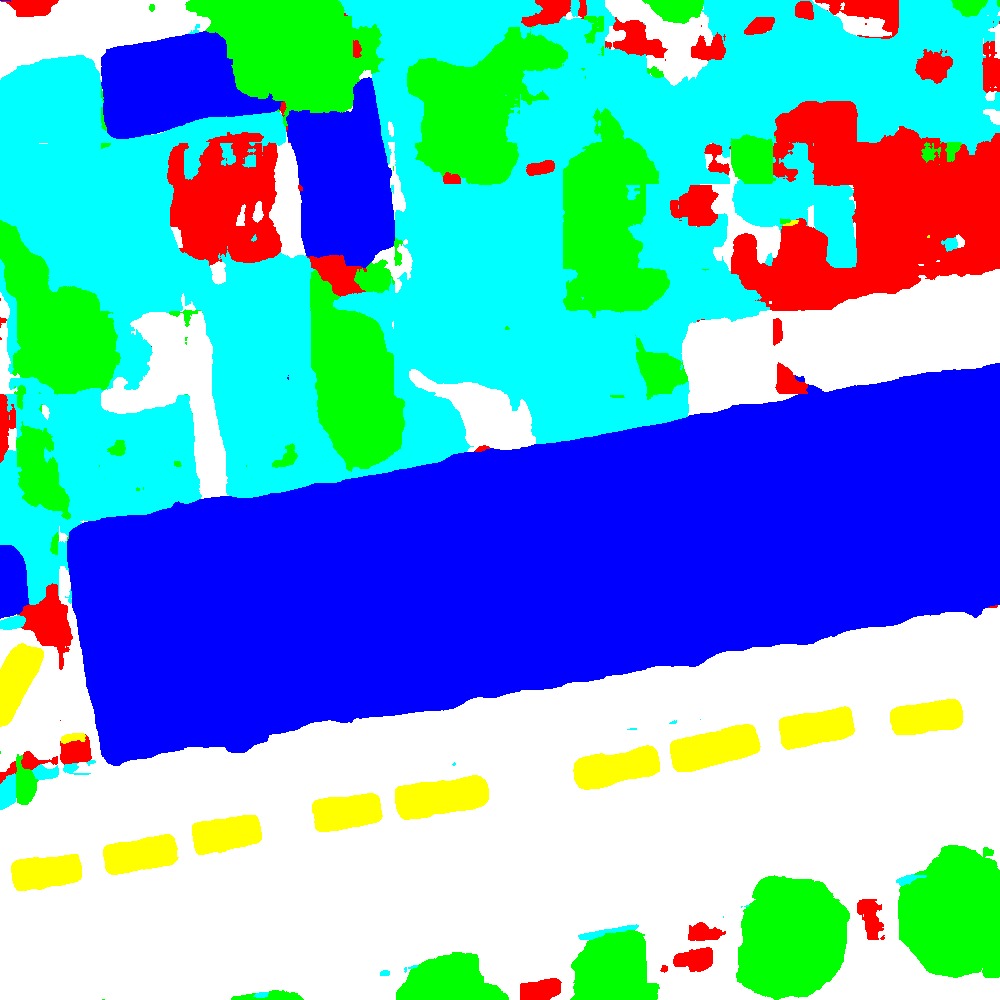} &
				\cincludegraphics[width=\exVaihingen\textwidth]{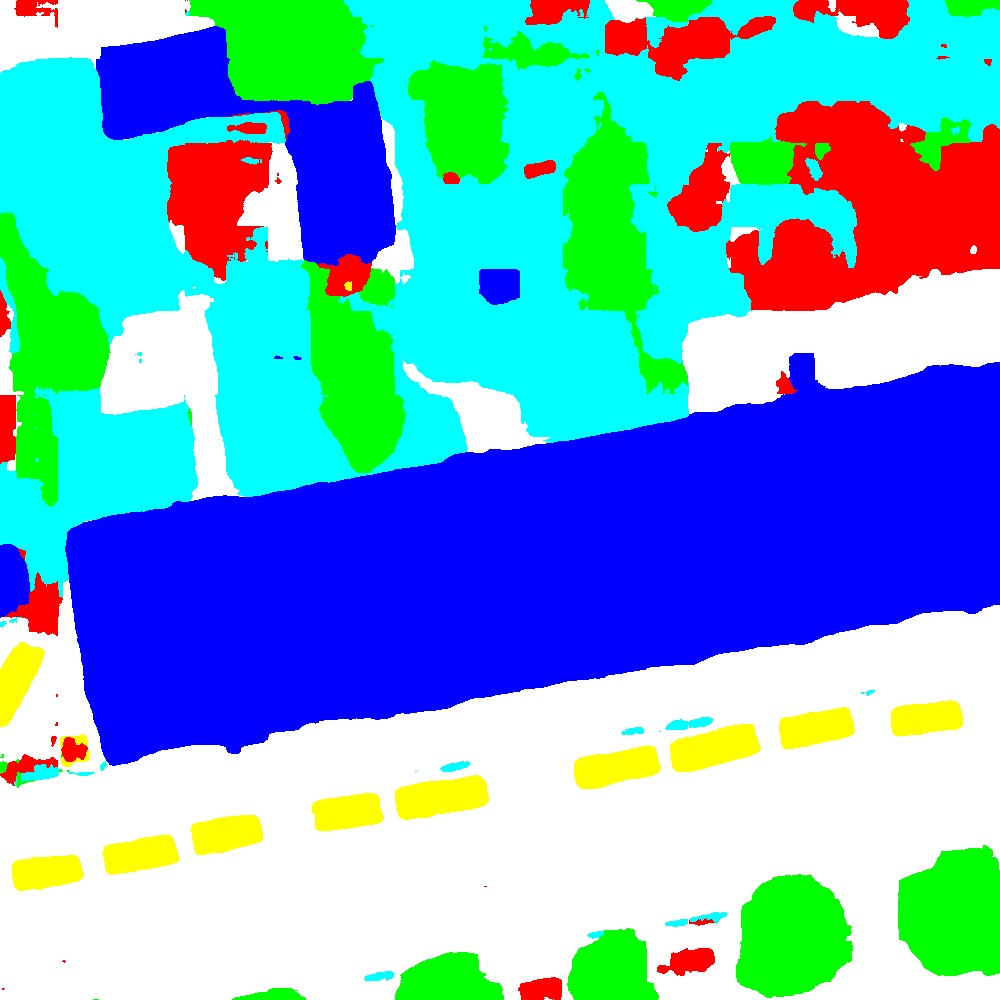} &
				\cincludegraphics[width=\exVaihingen\textwidth]{area5_13_dilated8_1000.jpeg} & \\[1.15cm]
			\end{tabular}
		\end{center}
		\captionof{figure}{Example predictions for the test set of the Potsdam dataset. 
			Legend -- White: impervious surfaces. Blue: buildings. Cyan: low vegetation. Green: trees. Yellow: cars. Red: clutter, background.
		}
		\label{fig:postdam_test_results}
	\end{table*}

	\section{Conclusions} \label{sec:conclusions}
	
	In this paper, we propose a novel approach based on Convolutional Networks to perform semantic segmentation of remote sensing scenes.
	The method exploits networks composed uniquely of dilated convolution layers that do not downsample the input.
	Based on these networks and their no downsampling property, the proposed approach:
	(i) employs, in the training phase, patches of different sizes, allowing the networks to capture multi-context characteristics given the distinct context size, and
	(ii) updates a score for each of these patch sizes in order to select the best one during the testing phase.
	
	We performed experiments on four high-resolution remote sensing datasets with very distinct properties: 
	(i) Coffee dataset~\cite{keiller2016icpr}, composed multispectral high-resolution scenes of coffee crops and non-coffee areas,
	(ii) GRSS Data Fusion dataset~\cite{liao2015processing}, consisting of very high-resolution of visible spectrum images,
	(iii) Vaihingen dataset~\cite{vaihingen}, composed of multispectral high-resolution images and normalized Digital Surface Model, and
	(iv) Potsdam dataset~\cite{potsdam}, also composed of multispectral high-resolution images and normalized Digital Surface Model.
	
	Experimental results have showed that our method is effective and robust.
	It achieved state-of-the-art results in two datasets (Coffee and GRSS Data Fusion datasets) outperforming several techniques (such as Fully Convolutional~\cite{long2015fully} and deconvolutional networks~\cite{badrinarayanan2015segnet}) that also exploit the multi-context paradigm.
	This shows the potential of the proposed method in learning multi-context information using patches of multiple sizes.
	
	For the other datasets (Vaihingen and Potsdam), although the proposed technique did not achieve state-of-the-art, it yielded competitive results.
	In fact, our approach outperformed some relevant baselines that exploit post-processing techniques (although we did not employ any) and other multi-context strategies.
	Among all methods, the proposed one has the least number of parameters and is, therefore, less pruned to overfitting and, consequently, easier to train.
	At the same time, it produces one of the highest accuracies, which shows the effectiveness of the proposed technique in extracting all feasible information from the data using limited (in terms of parameters) architectures.
	Furthermore, the proposed technique achieved one of the best results for the car class, which is one of the most difficult classes of these datasets because of its composition (small objects).
	This demonstrates the benefits of processing the input image without downsampling it, a process that preserves important details for classes that are composed of small objects.
	
	Aside from this, the proposed networks can be fine-tuned for any semantic segmentation application, since they do not depend on the patch size to process the data.
	This allows other applications to benefit from the patterns extracted by our models, a very important process mainly when working with small amounts of labeled data~\cite{nogueira2017towards}.
	
	The presented conclusions open opportunities towards a simplified use of deep learning methods for a better understanding of the Earth's surface, which is still needed for some remote sensing applications, such as agriculture or urban planning.
	In the future, we plan to better analyze the relation between the number of classes in the dataset and the number of parameters in the ConvNet.

	\section*{Acknowledgments}
	
	This work was partially financed by the Pr{\'o}-Reitoria de Pesquisa da Universidade Federal de Minas Gerais, CNPq (grant 312167/2015-6), CAPES (grant 88881.131682/2016-01), and Fapemig (APQ-00449-17).
	The authors gratefully acknowledge the support of NVIDIA Corporation with the donation of the GeForce GTX TITAN X GPU used for this research.
	
	\bibliographystyle{IEEEtran}
	\bibliography{bibliography}
	
\end{document}